\documentclass{article} % For LaTeX2e
\usepackage{iclr2024_conference,times}

\usepackage{amsmath,amsfonts,bm}

\def\eqref#1{equation~\ref{#1}}
\def\1{\bm{1}}

\DeclareMathAlphabet{\mathsfit}{\encodingdefault}{\sfdefault}{m}{sl}
\SetMathAlphabet{\mathsfit}{bold}{\encodingdefault}{\sfdefault}{bx}{n}

\usepackage[colorlinks=true,linkcolor=blue,citecolor=blue]{hyperref}
\usepackage{url}
\usepackage{graphicx}
\usepackage{subcaption}
\usepackage{booktabs}   
\usepackage{todonotes}
\usepackage{algorithm}
\usepackage{algpseudocode}
\usepackage{wrapfig}
\usepackage{amsmath}

\newcommand{\bs}{\mathbf{s}}
\newcommand{\ba}{\mathbf{a}}

\title{Handling Cost and Constraints with Off-Policy Deep Reinforcement Learning}

\author{Jared Markowitz, Jesse Silverberg, Gary Collins \\
Johns Hopkins University Applied Physics Laboratory \\
\texttt{\{Jared.Markowitz, Jesse.Silverberg, Gary.Collins\}@jhuapl.edu}
}

\iclrfinalcopy % Uncomment for camera-ready version, but NOT for submission.
\begin{document}

\maketitle

\begin{abstract}
Methods for off-policy deep reinforcement learning (DRL) offer improved sample efficiency relative to their on-policy counterparts, due to their ability to reuse data throughout the training process. For continuous action spaces, the most popular approaches to off-policy learning include policy improvement steps where a learned state-action ($Q$) value function is maximized over selected batches of data. These updates are often paired with regularization to combat associated overestimation of $Q$ values. With an eye toward safety, we revisit this strategy in environments with ``mixed-sign'' reward functions; that is, with reward functions that include independent positive (incentive) and negative (cost) terms. This setting is common in real-world applications, and may be addressed with or without constraints on the cost terms. We find the combination of function approximation and a term that maximizes $Q$ in the policy update to be problematic in such environments, because systematic errors in value estimation impact the contributions from the competing terms asymmetrically. This results in overemphasis of either incentives or costs and may severely limit learning. We explore two remedies to this issue. First, consistent with prior work, we find that periodic resetting of $Q$ and policy networks can be used to reduce value estimation error and improve learning in this setting. Second, we formulate novel off-policy actor-critic methods for both unconstrained and constrained learning that do not maximize $Q$ in the policy update.  We find that this second approach, when applied to continuous action spaces with mixed-sign rewards, consistently and significantly outperforms state-of-the-art methods augmented by resetting. We further explore the applicability of our approach to more frequently-studied control problems that do not have mixed-sign rewards, finding it to both more reliably produce competent performance and be competitive in terms of overall performance.

\end{abstract}

\section{Introduction}
Model-free deep reinforcement learning (DRL) algorithms have shown significant potential in numerous domains, from robotic manipulation \citep{openai2019solving} to complex games \citep{MnKaSiRuVeBeGrRiFiOsPeBeAnKiKuWiLeHa15} to plasma control \citep{degrave_magnetic_2022}. To become more widely used in real-world applications, however, further improvements in both training efficiency and agent safety are necessary.

Improved sample efficiency of DRL reduces training costs, whether through computation or runtime on physical systems. Off-policy methods for DRL reuse data collected throughout the training process, offering sample efficiency superior to that of on-policy approaches. State-of-the-art off-policy approaches typically leverage a replay buffer of experiences, learn a state-action value ($Q$) function, and find a policy that maximizes the learned $Q$ function. When applied to continuous action spaces, either a stochastic \citep{sac_newer} or deterministic \citep{fujimoto2018td3} policy representation may be used. Various approaches to regularizing learning in this setting have enabled significant increases in efficiency by enabling more network updates to be made per step of training data collected \citep{li2023efficient, d'oro2023sampleefficient}.

Beyond training requirements, agents intended for real-world use must be safe. Environments where safety is critical typically include competing positive (incentive) and negative (cost) terms in their reward functions.  The two may be considered together, or separately through constraints on the cost terms. In this work, we observe deficient behavior of state-of-the-art off-policy approaches for continuous control in such ``mixed-sign'' environments.  This scenario occurs in numerous application spaces, from robotics (navigation in the presence of obstacles, robotic surgery) to resource allocation (when both performance and efficiency must be considered), to financial decision-making (where the risk of losses exists in the pursuit of gains). 

Our contributions in this work are the following:
\begin{itemize}
    \item We diagnose the cause of the aforementioned poor performance, finding that it stems from value estimation error impacting contributions from different reward terms asymmetrically.
    \item To address the issue, we provide a novel algorithm building around the off-policy actor-critic originally proposed by \cite{degris2012off}.  We empirically demonstrate that our method is not prone to harmful levels of value estimation error, and that it provides effective learning in both the unconstrained and constrained settings.
\end{itemize}
We find that in environments with mixed-sign rewards, our method significantly outperforms existing approaches where $Q$ is maximized in the policy update  \citep{sac_newer, fujimoto2018td3}, even when they are augmented by resetting \citep{nikishin2022primacy}.  We further find that our algorithm is competitive with these approaches, as well as more reliable in the sense of producing at least moderate competence, on tasks that do not include mixed-sign rewards.

\begin{figure}
    \centering
    \includegraphics[width=0.325\textwidth]{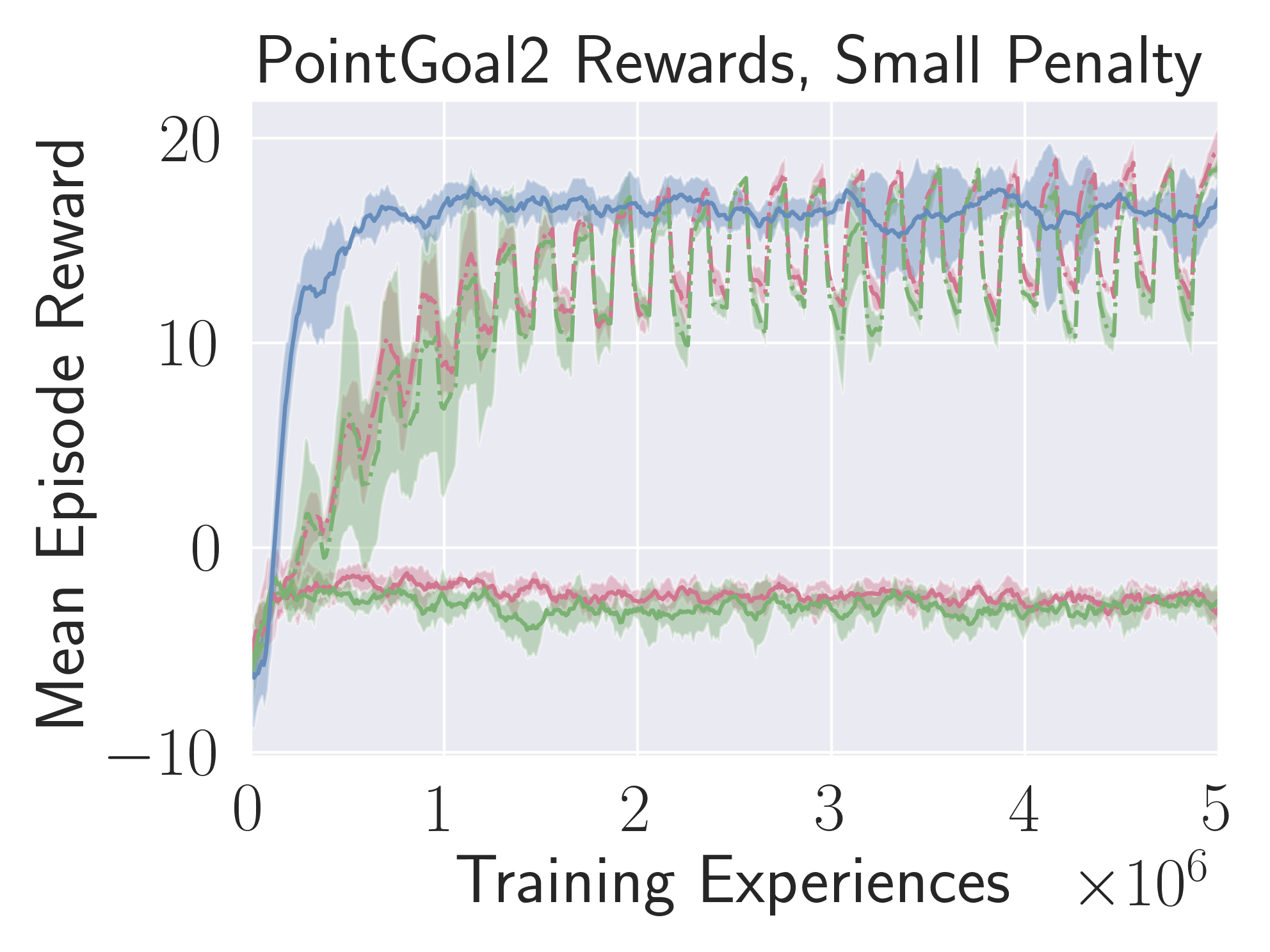}
    \includegraphics[width=0.325\textwidth]{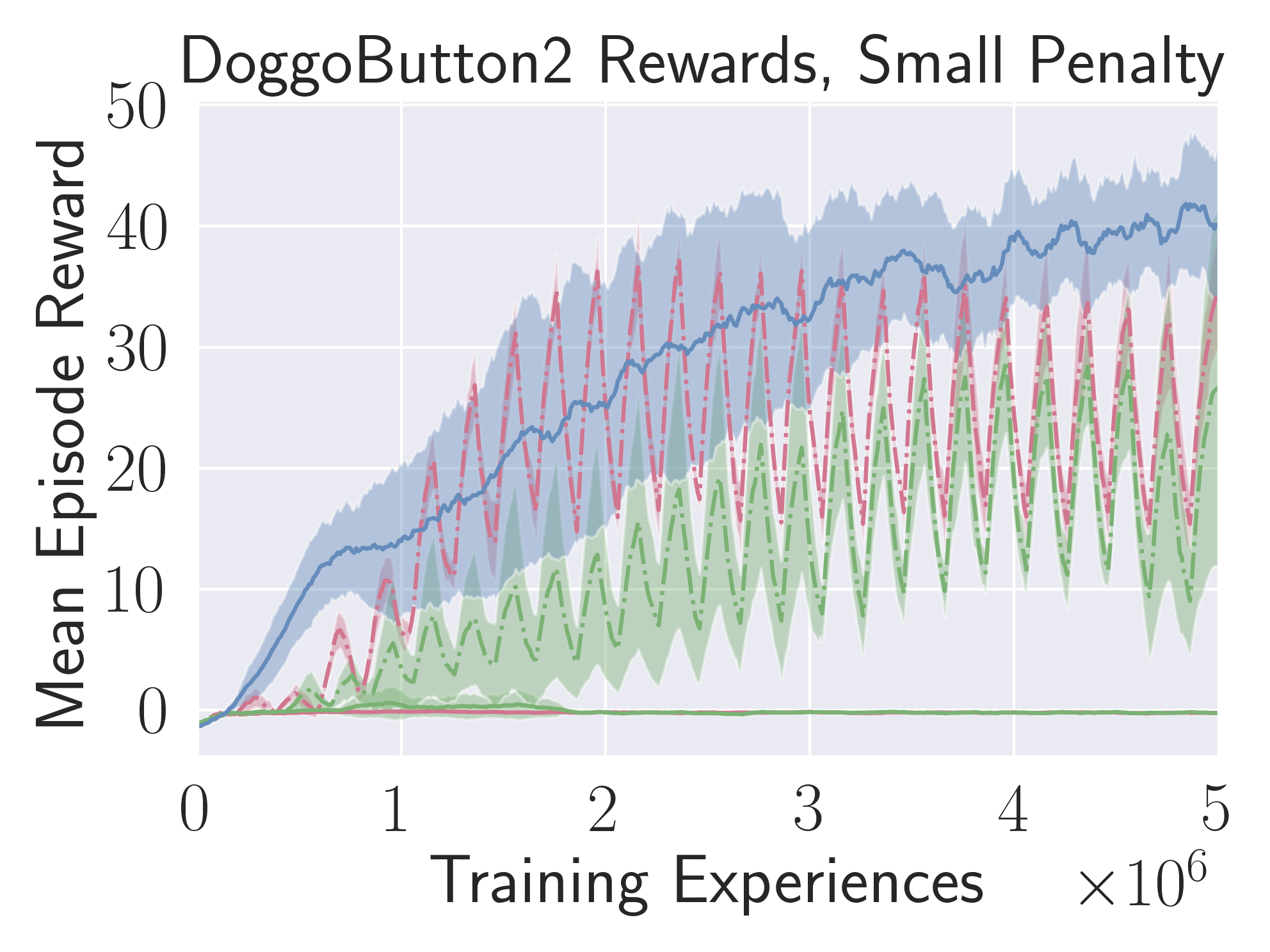}
    \includegraphics[width=0.325\textwidth]{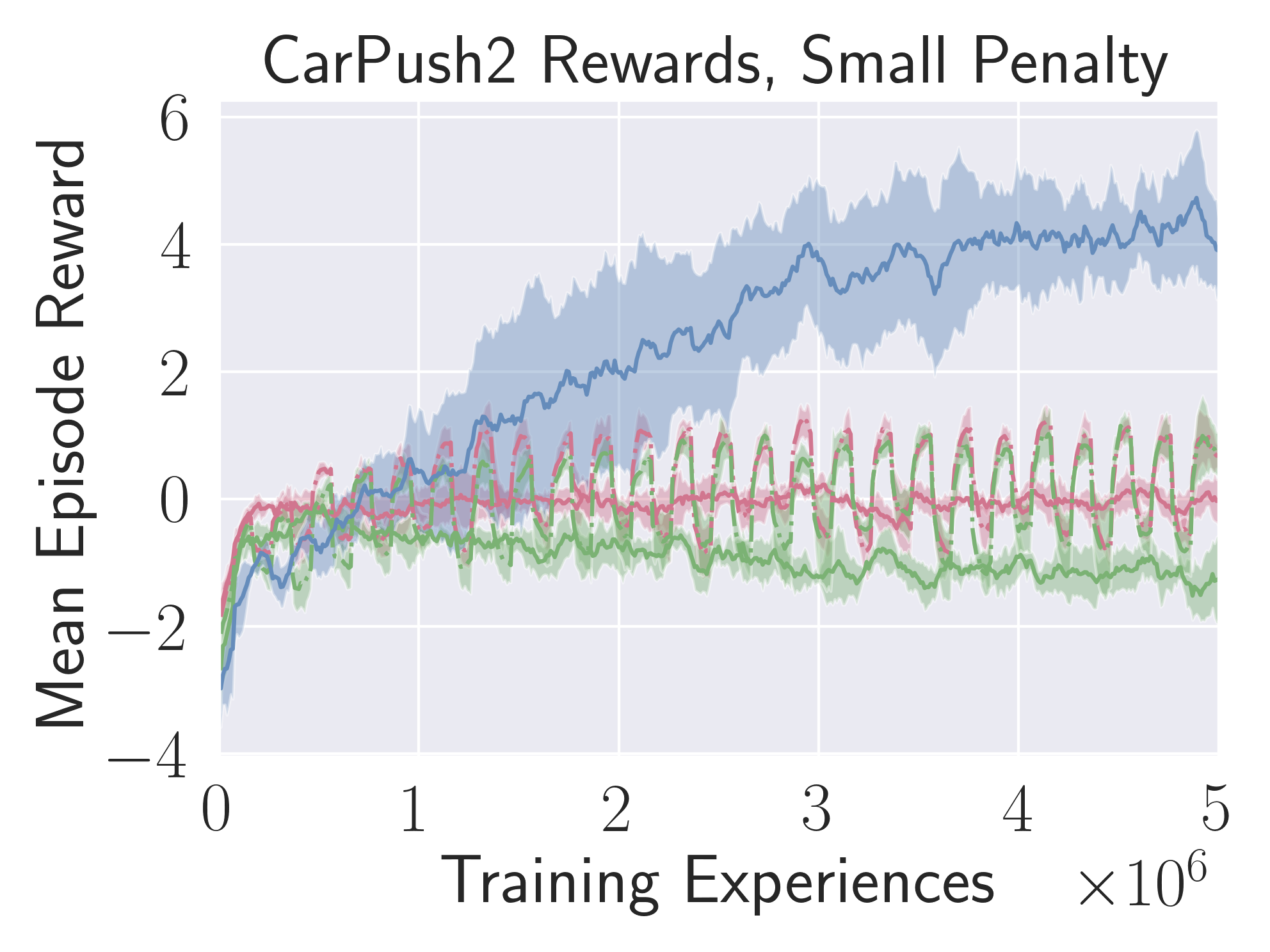}
    \includegraphics[width=0.9\textwidth]{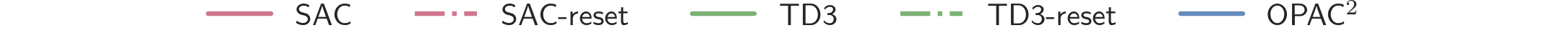}
    \includegraphics[width=0.325\textwidth]{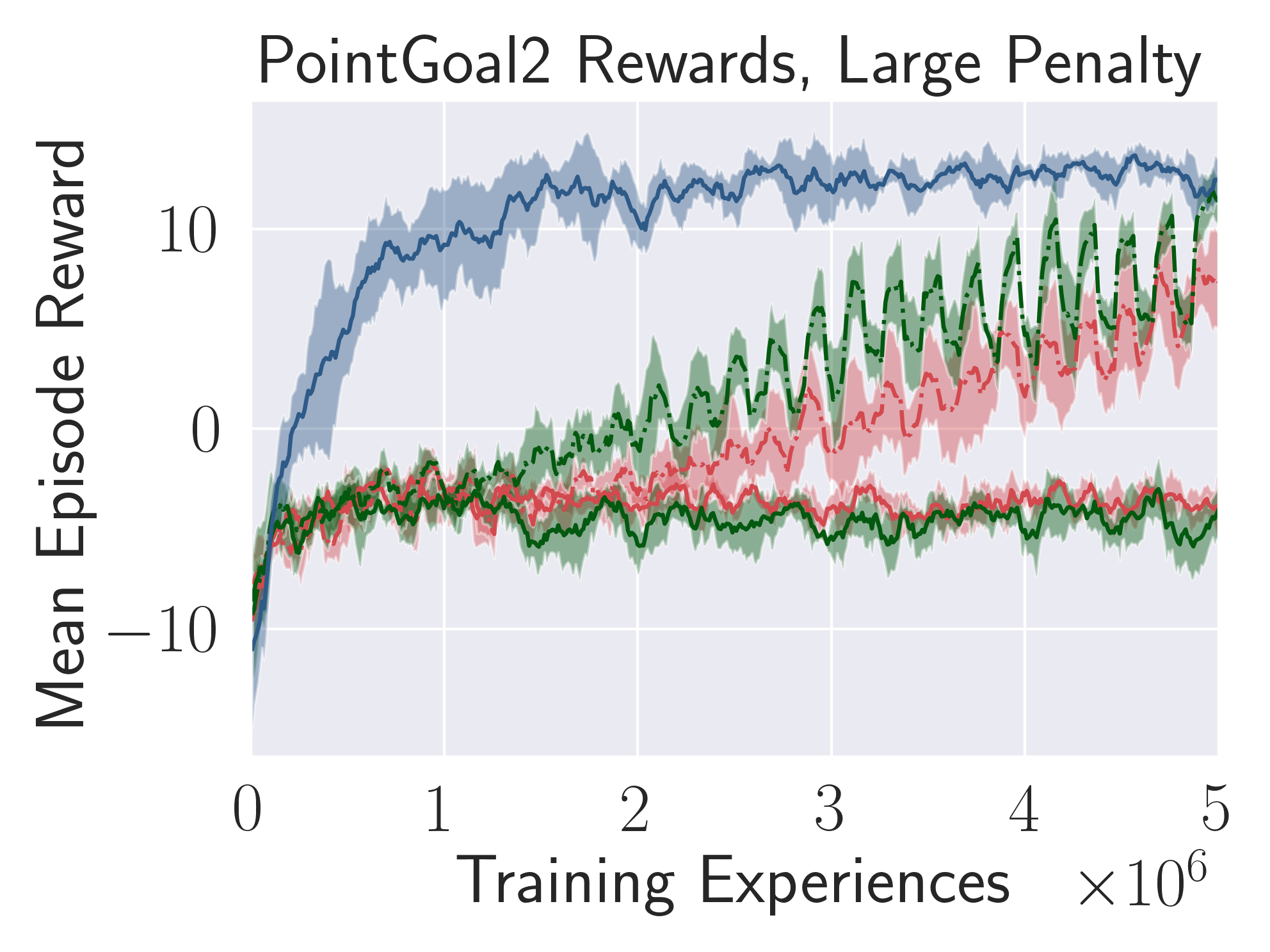}
    \includegraphics[width=0.325\textwidth]{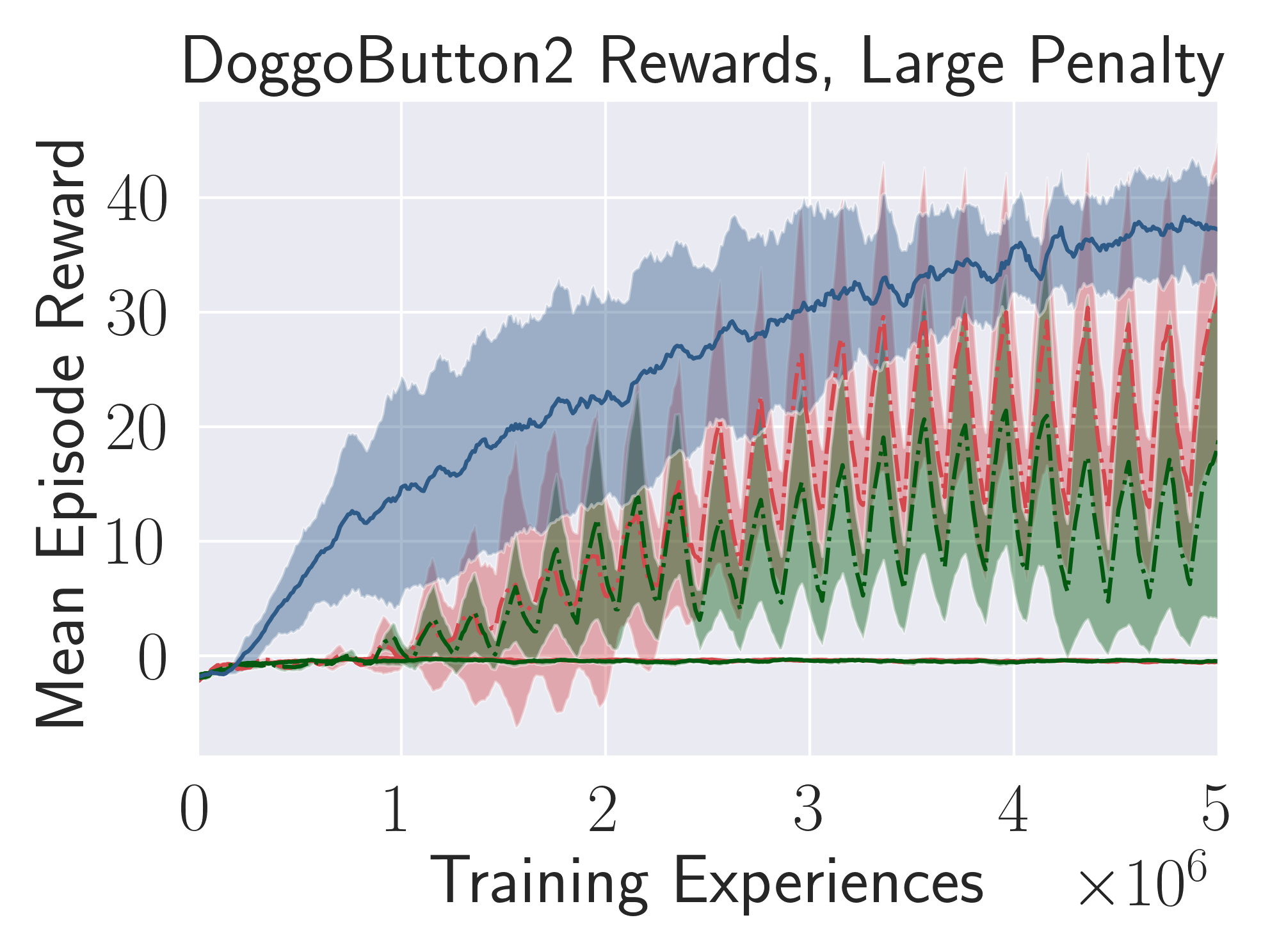}
    \includegraphics[width=0.325\textwidth]{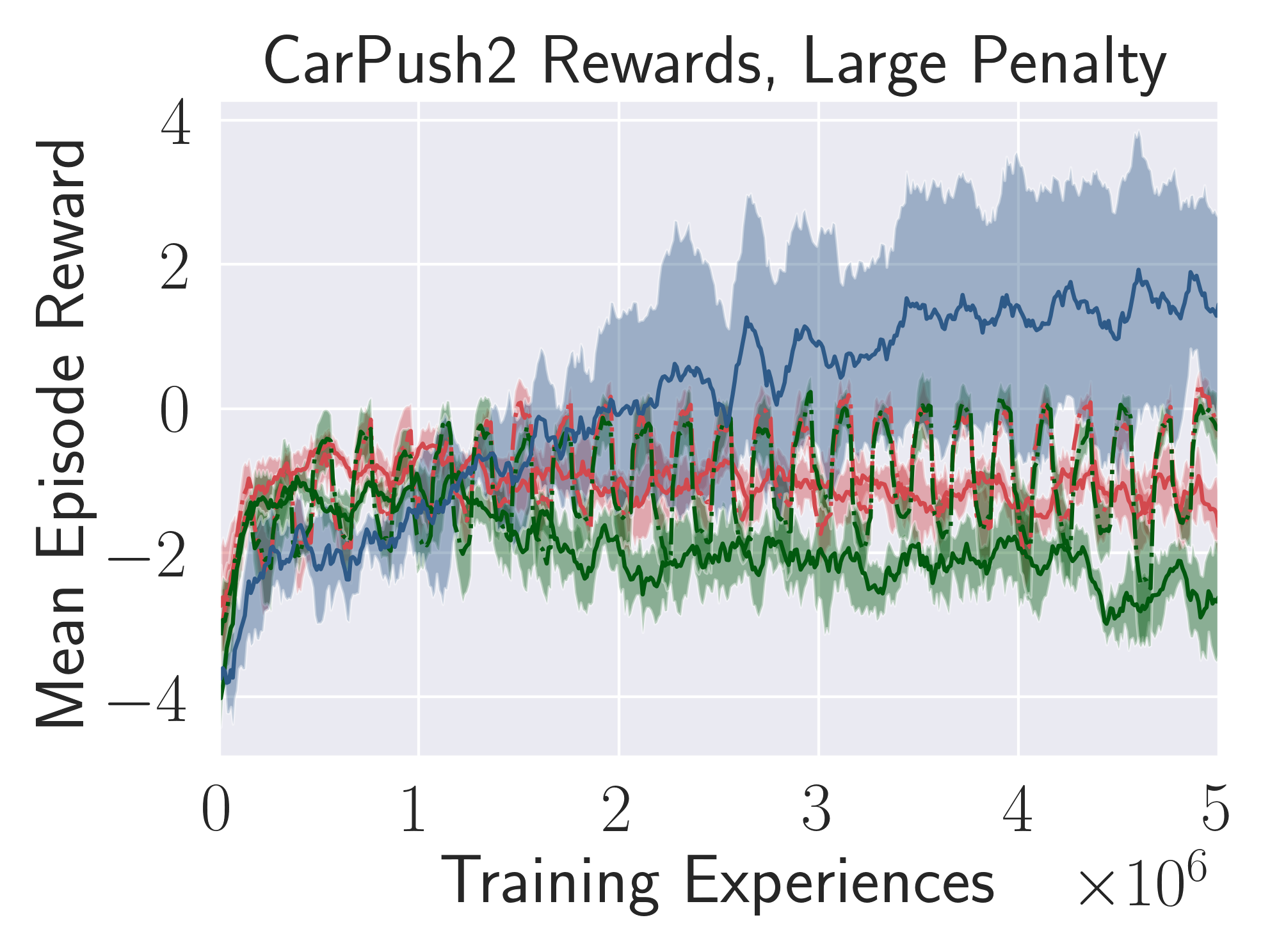}
    \includegraphics[width=0.9\textwidth]{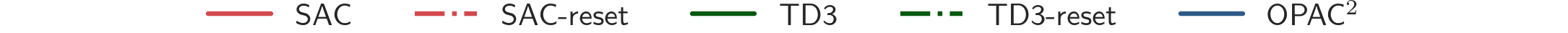}
    \caption{Sample unconstrained Learning on Safety Gym.  \textbf{Top row}: learning with relatively small cost weights.  \textbf{Bottom row}: learning with larger cost weights.  SAC and TD3 struggle to learn without resetting, particularly as the weight on cost terms increases. Even with periodic resets, they do not match the performance of OPAC$^2$.}
    \label{fig:unc_rew}
\end{figure}
\section{Preliminaries: Off-Policy RL for Continuous Action Spaces}
Reinforcement learning (RL) considers the problem of maximizing discounted returns in a Markov Decision Process (MDP) $(\mathcal{S}, \mathcal{A}, \mathcal{P}, r, \gamma)$, where $\mathcal{S}$ is the state space, $\mathcal{A}$ is the action space, $\mathcal{P}$ are the transition dynamics, $r$ is the reward function, and $\gamma\ \in [0,1)$ is the discount factor.  RL maximizes the value function $V(\bs_t) := \sum_{t'=t}^\infty \mathbb{E}_{\pi} [ \gamma^{t'-t} r(\bs_{t'}, \ba_{t'}) | \mathbf{s_t} ]$ at initial ($t=0$) states for trajectories encountered by an agent with policy $\pi(\ba_t|\bs_t)$ in an environment. The state-action value function $Q(\bs_t, \ba_t) :=  \sum_{t'=t}^\infty \mathbb{E}_{\pi} [ \gamma^{t'-t} r(\bs_{t'}, \ba_{t'}) | \mathbf{s_t}, \ba_t ]$ is the expected future discounted reward from a state $\mathbf{s_t}$, given that action $\ba_t$ is taken at time $t$. 

Practical approaches to off-policy deep reinforcement learning maintain a replay buffer $\mathcal{D}$ of transition tuples $(\bs, \ba, r, \bs', d)$, where $d$ indicates episode termination, and interleave data collection with network updates based on samples from the buffer. The network updates include both policy evaluation and policy improvement steps. 

Policy evaluation typically involves estimation of the $Q$ function via a Bellman backup with loss
\begin{equation}
L(\phi) = \mathbb{E}_{(\bs, \ba, r, \bs', d)\sim \mathcal{D}} \left[\left(r + \gamma(1-d)Q^{\pi}_{\mathrm{targ}}(\bs', \ba') -Q^{\pi}(\bs, \ba) \right)^2\right],
\label{eq1}
\end{equation}
where $\ba' \sim \pi_\theta(\ba|\bs')$ is sampled from the current policy evaluated at the next state $\bs'$.   A target network $Q^\pi_{\mathrm{targ}}$, delayed from the current estimate for $Q^{\pi}$, is often used to promote learning stability \citep{MnKaSiRuVeBeGrRiFiOsPeBeAnKiKuWiLeHa15}.

Policy improvement aims to maximize expected reward. The difference between the density of states encountered by the current policy and the totality of the replay buffer is often ignored in this step, introducing a bias in the policy estimate that is often manageable \citep{fu2019}. In practice this amounts to learning a more general policy, applicable to the full range of scenarios encountered by the agent throughout the training process.  The RL objective, for policy parameterized by variables $\theta$, may then be written as
\begin{equation}
J(\theta) \approx E_{\bs\sim\mathcal{D}} V^{\pi_\theta}(\bs) = E_{\bs\sim\mathcal{D}, \ba\sim\pi_\theta(\ba|\bs)} Q^{\pi_\theta}(\bs, \ba)
\end{equation}
The chain rule may be used to compute a policy gradient for this expression, and a more tractable form obtained by ignoring the second term:
\begin{equation}
\begin{split}
\nabla_\theta J(\theta) &\approx E_{\bs\sim\mathcal{D}, \ba\sim\pi_\theta(\ba|\bs)}
\left[\nabla_\theta\log(\pi_\theta(\ba|\bs))Q^{\pi_\theta}(\bs, \ba) + \nabla_\theta Q^{\pi_\theta}(\bs, \ba) \right]\\
&\approx E_{\bs\sim\mathcal{D}, \ba\sim\pi_\theta(\ba|\bs)} \left[\nabla_\theta\log(\pi_\theta(\ba|\bs))Q^{\pi_\theta}(\bs, \ba)\right]
\label{pg}
\end{split}
\end{equation}
Neglect of the second term, which may be difficult to compute off-policy, is shown to preserve the local minima to which the policy converges in the tabular case in \cite{degris2012off}. To reduce variance of the policy gradient estimate, \cite{degris2012off} recommend removing a state dependent baseline, resulting in
\begin{equation}
\nabla_\theta J(\theta) \approx E_{\bs\sim\mathcal{D}, \ba\sim\pi_\theta(\ba|\bs)} \Big[\nabla_\theta \log(\pi_\theta(\ba|\bs)) A^{\pi_\theta}(\bs, \ba)\Big],
\label{opac}
\end{equation}
where $A^{\pi_\theta}(\bs, \ba) = Q^{\pi_\theta}(\bs, \ba) - V^{\pi_\theta}(\bs)$ is the advantage function.   

There are two common alternatives to (Eq. \ref{pg}), both more commonly used today. The first is the deterministic policy gradient (DPG; \cite{dpg}), which represents the policy as a deterministic function $\mu_{\theta}(\bs)$. With that representation, the policy may be updated according to 
\begin{equation}
\nabla_\theta J(\theta) \approx \mathbb{E}_{\bs\sim\mathcal{D}} \left[\nabla_\theta \mu_\theta(\bs) \nabla_{\ba}Q^{\mu_\theta}(\bs, \ba)\vert_{\ba=\mu_\theta(\bs)}\right]
\label{dpg}
\end{equation}
The second is the reparameterization trick, which encodes the stochasticity of the policy in an independent noise variable. The resulting form has provably lower variance than (\ref{pg}) \citep{reparam}, and is used by Soft Actor-Critic (SAC; \cite{sac_original, sac_newer}). SAC also uses a ``squashed'' action representation for enforcing control bounds: $\ba_\theta(\bs, \xi) = \tanh(\mathbf{\mu_\theta}(\bs) + \mathbf{\sigma_\theta}(\bs)\odot\xi)$,  where $\xi \sim \mathcal{N}(\mathbf{0}, I)$ and $\mu_\theta$, $\sigma_\theta$ are the state-dependent mean and variance, respectively. Including a second term to encourage policy entropy, SAC performs soft policy improvement via
\begin{equation}
\nabla_\theta J(\theta) = \mathbb{E}_{\bs\sim\mathcal{D}, \mathbf{\xi} \sim\ \mathcal{N} (\mathbf{0}, \mathbf{1})} \left[ \nabla_\theta Q^{\pi_\theta}(\bs, \ba_\theta(\bs, \mathbf{\xi})) - \alpha \nabla_\theta \log(\pi_\theta(\ba_\theta(\bs, \mathbf{\xi})|\bs) \right].
\label{rp}
\end{equation}
Modern off-policy approaches using the deterministic policy gradient  (e.g., TD3; \cite{fujimoto2018td3}) and reparameterization trick (e.g., SAC) consider multiple $Q$ functions to combat overestimation of $Q$ originating in its maximization in the policy update steps (\ref{dpg}) and (\ref{rp}). This mechanism was addressed for discrete action spaces in \cite{double_q} and explained in detail for continuous action spaces in \cite{fujimoto2018td3}.  It is rooted in overestimation errors in the value update being propagated by the policy update. To combat this effect, both TD3 and SAC employ the ``clipped double $Q$ trick'' in the policy evaluation step
\begin{equation}
L(\phi) = \mathbb{E}_{(\bs, \ba, r, \bs', d)\sim \mathcal{D}} \left[\left(r + \gamma(1-d)\min_{i=1,2} Q^\pi_{\phi_i, \mathrm{targ}}(\bs', \ba') -Q^\pi_\phi(\bs, \ba) \right)^2\right]
\end{equation}
to train two $Q$ functions with different initial parameters $\phi_1, \phi_2$ toward the minimum of their outputs, for each sampled experience.  SAC also uses the minimum of the two $Q$ functions in the policy update. Other methods \citep{redq, droq} go further, training larger ensembles of $Q$ functions. These adjustments  mitigate $Q$ overestimation, but do not fully solve the problem. In some cases, $Q$ may still be overestimated; in others, it may be underestimated. As noted in \cite{fujimoto2018td3}, underestimation is preferred because it is not propagated by the policy update.  

Notably, we find that these corrections are unnecessary with the off-policy actor-critic (Eq. \ref{pg}). While $Q$ may still be slightly overestimated, we find that the ``gentler'' updates provided by the off-policy actor-critic remove the tendency for large inaccuracies.  The addition of the value function in Eq. \ref{opac} both provides variance reduction and weakens the feedback between the policy and Q updates, leading to learning that we find to be both reliable and performant.

\begin{wrapfigure}[30]{r}{0.325\textwidth}
    \includegraphics[width=0.325\textwidth]{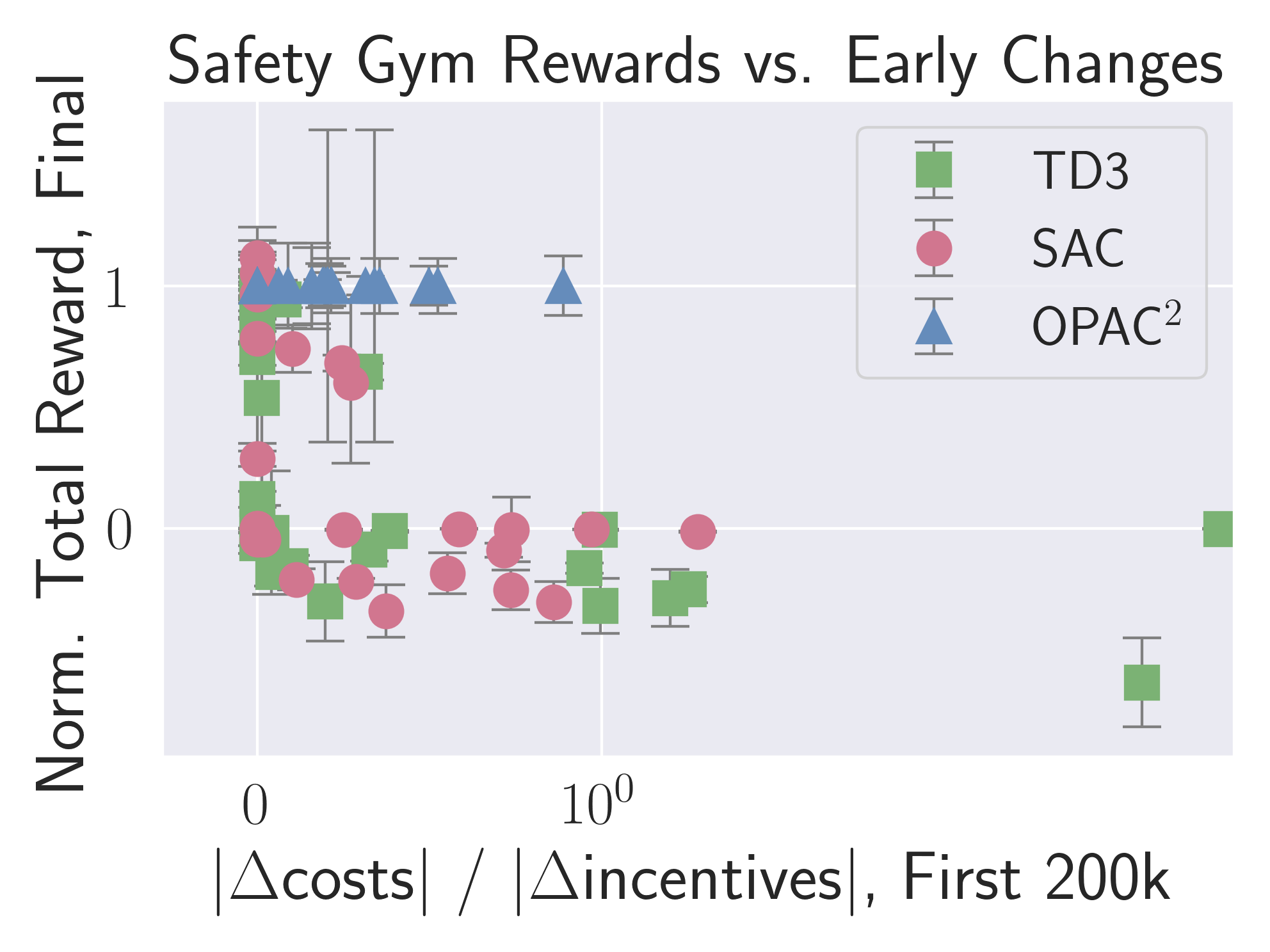}
     \includegraphics[width=0.325\textwidth]{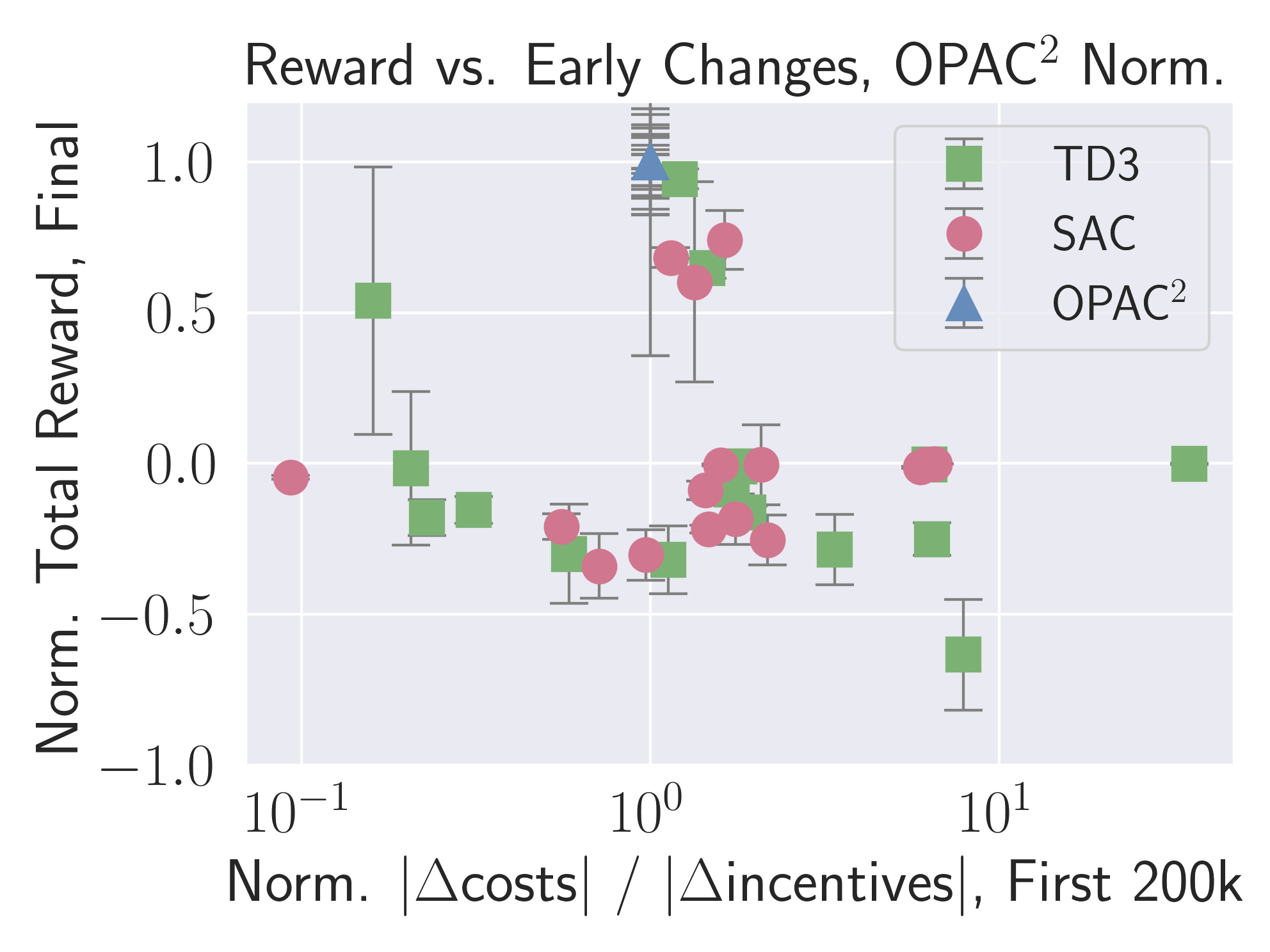}
    \caption{\textbf{Top:} Total (cost and incentive) reward per-episode of fully-trained agents as a function of cost adjustment relative to reward adjustment early in Safety Gym training. \textbf{Bottom:} SAC and TD3 perform best when prioritizing cost similarly to OPAC$^2$.}
    \label{fig:200k}
\end{wrapfigure}
\section{The Impact of Competing Objectives}\label{competing}

As explained in \cite{fujimoto2018td3}, the tendency of $Q$ to be systematically overestimated when maximized in the policy update occurs because of the pairing of that update with the use of $Q$ (albeit delayed) in the target of its own update (Eq. \ref{eq1}). SAC and TD3 combat this upward bias by learning two $Q$ functions, using each as the target for the other's update. To make sure this target is never a worse overestimate than just using the same $Q$ function, the target is set to the minimum of the two $Q$ functions (the ``clipped double $Q$ trick''). SAC also uses this minimum in the policy update. Unfortunately, it is unlikely that any form of regularization will exactly balance the intrinsic tendency of $Q$ to be overestimated when maximized in the policy update.

Now consider a continuous control problem where the reward function is a sum of multiple independent objectives. A special case of this is when rewards are ``mixed-sign;'' that is, contain independent incentive ($>0$) and cost ($<0$) terms. As recently explored \citep{nikishin2022primacy, li2023efficient}, off-policy DRL agents are prone to overemphasizing experiences collected early in training.  In the context of multiple independent objectives, this corresponds to excessive focus on the reward terms most easily accessed by an agent of low capability. Fixation on some terms may lead to others being neglected entirely, as has been observed for entire \emph{tasks} in multi-task learning \citep{grad_surgery, cagrad}. The situation is more likely to become acute with mixed-sign rewards; favoring a given reward term early in training may cause the agent to ignore terms designed to counterbalance it. Even if overemphasis of particular terms does not lead to wholesale neglect of others, it is likely to impact the relative consideration of different terms by the agent and thereby impact behavior.

To explore this issue, we evaluated the performance of different off-policy algorithms on the OpenAI Safety Gym \citep{ray2019safetygym}, a ``mixed-sign'' suite of robotic navigation tasks with obstacles.  In the top panel of Figure \ref{fig:200k}, each marker represents 5 training runs on a given configuration, where a configuration is a combination of robot, obstacle set, task, and penalty assessed each time an obstacle is contacted.  Each algorithm tackled the same set of 24 configurations, covering significant range in the ease of accessing incentive and cost terms early in training.  The $x$-axis reflects how much an agent adjusts its cost levels relative to incentives early on: it is the \emph{change} in average costs accumulated by the agent at the start of training compared to what it accumulates 200k steps in, divided by that same difference for incentives. Points with $x=0$ were configured to have $0$ penalty (ignore costs).  The $y$-axis is the average total reward (including costs and incentives) per-episode of the fully trained agent, normalized by the level achieved by our updated off-policy actor-critic (OPAC$^2$) agent.  We observe that SAC and TD3 are typically able to accumulate incentives well only when cost is close to ignored, and tend to prioritize cost more highly than OPAC$^2$. In the bottom panel, we plot all configurations with nonzero penalty weights.  We normalize the x and y values for each point by those obtained by OPAC$^2$ in the environment configuration corresponding to the point. We observe that SAC and TD3 are able to compete with OPAC$^2$ only when they adjust cost and reward similarly to OPAC$^2$ early in training, suggesting improper balancing in other runs.  We provide more details on these experiments in Appendix \ref{comp_obj_app} and describe OPAC$^2$ further below.

\section{Addressing Erroneous $Q$ Estimates}
We investigate two routes for addressing this problem. Section \ref{sec:resetting} examines the efficacy of a popular regularization strategy in this setting. In Section \ref{sec:algorithm}, we propose a novel algorithm with desirable properties for this problem.

\subsection{Regularization via Resetting} \label{sec:resetting}
One method recently shown to be effective for improving off-policy learning is to periodically reset policy and value networks throughout training, while preserving accumulated experience \citep{nikishin2022primacy}. This method has been shown to enable learning that is extremely sample-efficient, by allowing many network updates to be conducted per collected data point (learning at high ``replay ratios''). \cite{li2023efficient} observed that resetting reduces temporal difference (TD) error on validation (i.e., not seen in training) transitions. We find this observation to be particularly relevant in mixed-sign environments; in many cases, resetting is seen to enable learning with SAC and TD3 when it would not otherwise be possible. However it is not seen to provide optimal performance, and could be inappropriate in scenarios where \emph{cumulative} cost is a consideration\footnote{\cite{d'oro2023sampleefficient} incorporated offline training after resets to mitigate this issue.}.

\subsection{Updating the Off-Policy Actor-Critic} \label{sec:algorithm}
To more directly address erroneous $Q$ estimates in environments with competing objectives, we revisit the off-policy actor-critic (Eq. \ref{opac}). We seek to build a practical algorithm around it, leveraging techniques from more recent methods as well as novel extensions to reduce the variance on the policy gradient estimate while leveraging its potentially less biased value estimation. Differing from \cite{degris2012off}, we neglect importance weights in our gradient estimates,  an approximation that significantly reduces variance and amounts to learning a more general policy over all states in the buffer (rather than one tailored to the current density of states). We adopted several aspects of Soft Actor-Critic, including $Q$ and $V$ updates that match the original version of SAC (though we require only a single $Q$ network) and ``squashing'' of actions with a hyperbolic tangent to respect control bounds.  The latter tactic reduces policy gradient bias relative to methods that allow the environment to clip actions \citep{clipped_action_policy_gradient}. Finally, we normalized advantage estimates prior to use in the policy gradient (Eq. \ref{opac}).

Incentivizing policy entropy has been shown to improve learning in complex problems and with high-dimensional control spaces \citep{sac_original}. One existing approach is the ``maximum-entropy'' framework of \cite{sac_newer}, wherein entropy is bundled with the reward. While this strategy is compatible with our approach, we also considered the use of an entropy bonus to be added directly to the policy loss.  Empirically, we found this entropy bonus to outperform the ``max-entropy'' strategy (Figure \ref{fig:diagnostics}) in environments with mixed-sign rewards. To accommodate squashing, our bonus used an action sampled with the reparameterization trick and was optimized towards a target entropy level, similar to \cite{sac_newer}.  Pseudocode for our unconstrained algorithm (OPAC$^2$) is provided in Appendix \ref{app:unc_alg}. 

\subsubsection{Constrained Approach}
Reinforcement learning with costs may alternatively be formulated with constraints. Differing from the unconstrained setting discussed above, Constrained Markov Decision Processes (CMDPs) have positive rewards $r(\bs, \ba)$ and costs $c(\bs, \ba)$ provided separately for each time step, as well as an overall constraint $C(\tau) = F(c(\bs_1, \ba_1), \ldots, c(\bs_T, \ba_T))$ defined over the whole trajectory $\tau \equiv \bs_1, \ba_1, \ldots,  \bs_T, \ba_T $. The associated learning problem is to maximize the value function $V^{\pi}(\bs_t)$ associated with rewards, such that the expected value of the constraint over trajectories sampled by the agent will not exceed a fixed threshold $M$: $J_C(\theta) = E_{\tau \sim p_\theta(\tau)}C(\tau) < M$. Here $p_\theta(\tau)$ is the probability distribution of trajectories $\tau$ encountered by an agent parameterized by $\theta$. While other functions are possible, here we will be concerned with constraints on total trajectory cost: $C(\tau) = \sum_{t=1}^T c(\bs_t, \ba_t)$.
% Changed d -> M

Constrained RL is often conducted using dual methods \citep{bert, Boyd2004}, and has previously been explored off-policy with SAC \citep{Ha_etal_2020, zhou_c_sac, wcsac}. The goal is to learn a Lagrange multiplier $\beta$ that scales cost relative to reward during policy optimization in order to satisfy the constraint. This circumvents the need for reward shaping, as the weight of the cost terms is learned. As $\beta$ changes throughout training, it is beneficial to train separate $Q$ and $V$ networks for reward ($Q_r$, $V_r$) and cost ($Q_c$, $V_c$). To facilitate cost matching in the states currently frequented by the agent, we update $\beta$ toward the average cost incurred by the agent \emph{over only the last epoch of training}. This update, like those on other quantities, is conducted every time step. Our full constrained approach, Constrained Off-Policy Actor-Critic, SQUAshed and REgularizeD (C-OPAC$^2$), is given in Algorithm \ref{alg:copac2}. This differs from the unconstrained approach in two ways: we now learn two $Q$ and $V$ networks, one each for rewards and costs, and we also learn $\beta$ in order to satisfy the constraint.

\begin{algorithm}[ht]
\caption{Constrained Off-Policy Actor-Critic, SQUAshed and REgularizeD (C-OPAC$^2$) \label{alg:copac2}}
\begin{algorithmic}[1]
\State \textbf{Input:} Initial policy parameters $\theta$; $Q_r$, $Q_c$ parameters $\phi_{r}, \phi_{c}$; $V_r$, $V_c$ parameters $\psi_{r}, \psi_{c}$
\State \textbf{Input:} Initial entropy weight $\alpha$, penalty weight $\beta$
\State Initialize $V_r$, $V_c$ target network parameters: $\psi_{r, \mathrm{targ}} \leftarrow \psi_r; \psi_{c, \mathrm{targ}} \leftarrow \psi_c$
\State Initialize replay buffer $\mathcal{D} = \emptyset$,  a ring buffer of fixed size
\For{iteration $k \in [0, \dots, K-1$}
\For{step $s \in [0, \dots, S-1]$}  \Comment{Typically take just one step ($S=1$)}
\State Sample $\ba \sim \pi_{\theta_k}(\ba|\bs)$; observe $\bs' \sim p(\bs' | \bs, \ba)$  \Comment{One step in CMDP}
\State Store transition tuple $ (\bs,\ba,\bs', r(\bs,\ba), c(\bs, \ba))$ in $\mathcal{D}$ 
\EndFor
\For{gradient step $g \in [0, \dots, G-1]$}
\State Sample batch $B = \{(\bs_i, \ba_i, \bs'_i, r_i, c_i, d_i) \}$ from $\mathcal{D}$ 
\State Update $\beta$: $\beta \leftarrow \beta - \lambda_\beta\nabla_\beta \beta(M-J_C(\theta) )$ \Comment{$J_C(\theta)$ computed over most recent epoch}
\State Compute Q error: $\mathcal{E}_Q(B) = \sum_i \left[Q_r(\bs_i, \ba_i) - (r_i + \gamma(1-d_i) V_{r, \mathrm{targ}}(\bs_i))\right]^2 + $ \\ $\hspace{162pt}\left[Q_c(\bs_i, \ba_i) - (c_i + \gamma(1-d_i) V_{c, \mathrm{targ}}(\bs_i))\right]^2$
\State Update Q: $\phi_r \leftarrow \phi_r - \lambda_\phi \nabla_{\phi_r} \mathcal{E}_Q(B)$; $\phi_c \leftarrow \phi_c - \lambda_\phi \nabla_{\phi_c} \mathcal{E}_Q(B)$
\State Sample $\ba_{i,\pi} \sim \pi(\ba|\bs_i)$, using $\tanh$ squashing
\State Compute $V$ error: $\mathcal{E}_V(B) = \sum_i \left[V_r(\bs_i )- Q_r(\bs_i, \ba_{i, \pi})\right]^2 + \left[V_c(\bs_i) - Q_r(\bs_i, \ba_{i,\pi})\right]^2$
\State Update $V$: $\psi_r \leftarrow \psi_r - \lambda_\psi \nabla_{\psi_r} \mathcal{E}_V(B)$; $\psi_c \leftarrow \psi_c - \lambda_\psi \nabla_{\psi_c} \mathcal{E}_V(B)$
\State Compute $A(\bs_i, \ba_{i,\pi}) = Q_r(\bs_i, \ba_{i,\pi}) - V_r(\bs_i) - \beta [Q_c(\bs_i, \ba_{i,\pi}) - V_c(\bs_i)]$ \Comment{Normalize}
\State Sample $\ba_{\mathrm{rp},\pi} \sim \pi(\ba|\bs_i)$ with reparameterization trick (for $\tanh$ squashing)
\State Compute $\pi$ loss: $\mathcal{E}_\pi(B) = \sum_i\left[ \alpha\log \pi_\theta(\ba_{\mathrm{rp},\pi} | \bs_i) - A(\bs_i, \ba_{i,\pi})\log(\pi_\theta(\ba_{i,\pi} | \bs_i) \right] $
\State Update $\pi$: $\theta \leftarrow \theta - \lambda_\theta \nabla_\theta \mathcal{E}_\pi(B)$
\State Update $\alpha$ : $\alpha \leftarrow \alpha - \lambda_\alpha\nabla_\alpha \left[ -\alpha\log(\pi(\ba_{i,\pi}|\bs_i)) -\alpha\mathcal{H}_{\mathrm{target}}\right]$
\State Update value targets: $\psi_{r, \mathrm{targ}} \leftarrow \rho \psi_{r, \mathrm{targ}} + (1-\rho)\psi_{r}; \psi_{c, \mathrm{targ}} \leftarrow \rho \psi_{c, \mathrm{targ}} + (1-\rho)\psi_{c}$
\EndFor

\EndFor
\end{algorithmic}
\end{algorithm}

\section{Experiments}\label{sec:experiments}
We used the  OpenAI Safety Gym \citep{ray2019safetygym} to explore off-policy learning for environments with continuous action spaces and mixed-sign rewards. Safety Gym is a configurable set of robotic navigation tasks, wherein different robots must navigate through courses containing multiple obstacle types to perform different tasks as many times as possible in a fixed time window. The locations of goals and obstacles are randomized, leading to outcome variability and necessitating the learning of a generalized control strategy. For unconstrained experiments, each cost event incurred a fixed, negative reward. For constrained experiments, the weight coefficient for cost events was learned. We evaluated all robots and tasks for the most obstacle-rich (level 2) publicly available environments. Additional experimental details, including hyperparameters, are given in Appendix \ref{app:exp}.

\subsection{Unconstrained Learning}
To explore the effect of mixed-sign rewards on unconstrained off-policy learning, we considered both small and large static cost weights in our target environments. The larger penalty weights matched those used on-policy in \cite{aaai_paper}, while the smaller weights were reduced by a factor of 2. We compared OPAC$^2$ with SAC and TD3, both with and without resets. A  sample of the resulting learning curves is shown in Figure \ref{fig:unc_rew}.  The remainder, which follow similar qualitative trends, are provided in Appendix \ref{unc_app}. We find SAC and TD3 to be generally unable to reach adequate performance, particularly as the penalty weight increases. The situation is improved when resetting is applied; however, it often does not result in performance that reaches the level of OPAC$^2$. Attesting to the sample efficiency of off-policy methods, we find that OPAC$^2$ is typically able to reach performance comparable to that of Proximal Policy Optimization (PPO; \cite{ScWoDhRaKl17}) and Trust Region Policy Optimization (TRPO; \cite{schulman2017trust}) using 20--50 times fewer samples (based on results reported by \cite{aaai_paper}). On the Doggo environments, OPAC$^2$ exceeds the positive reward accumulation of PPO and TRPO trained \emph{without cost terms} and with twice as many samples \citep{ray2019safetygym}.

\begin{figure}
\includegraphics[width=0.325\textwidth]{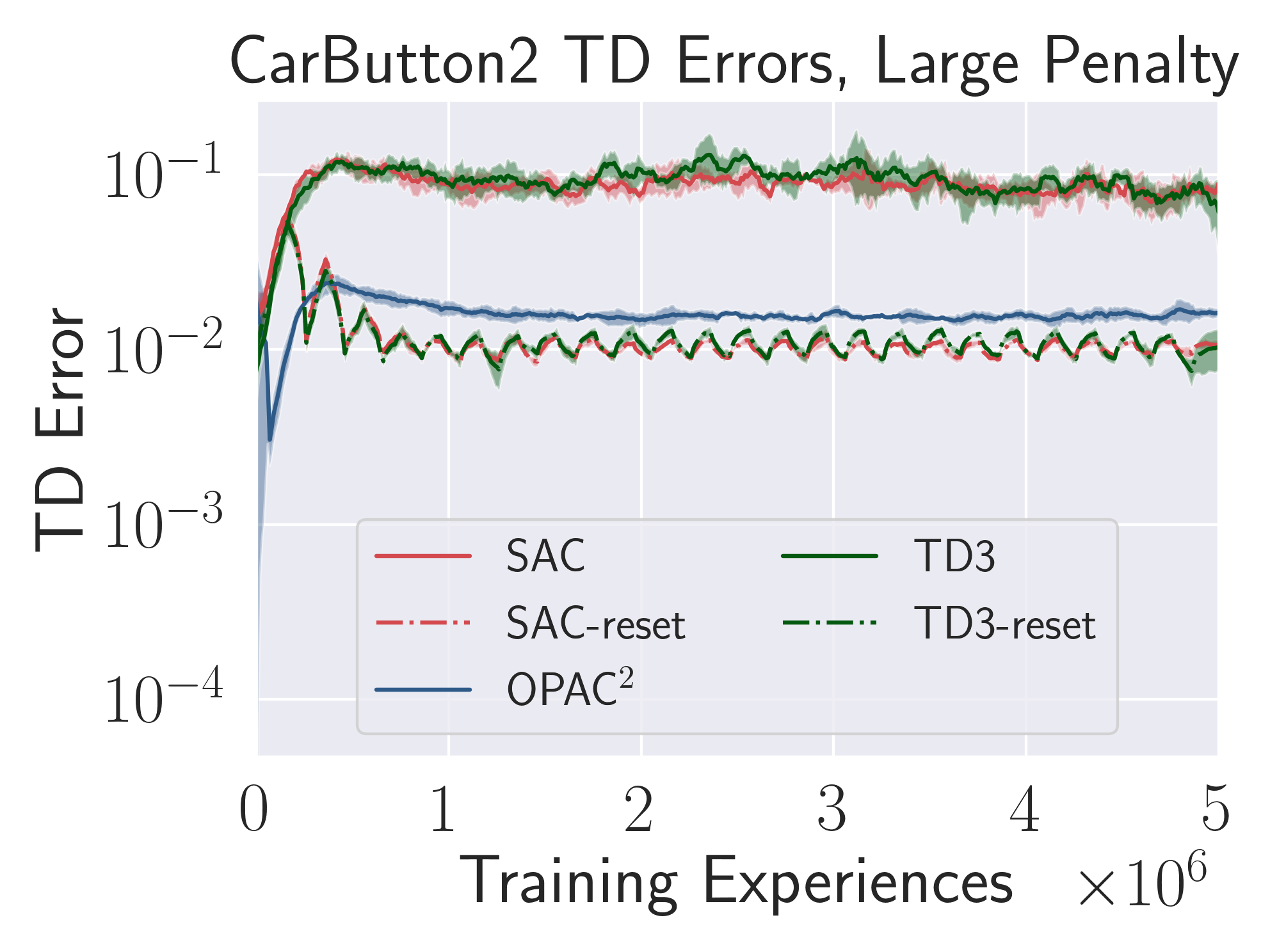}
\includegraphics[width=0.325\textwidth]{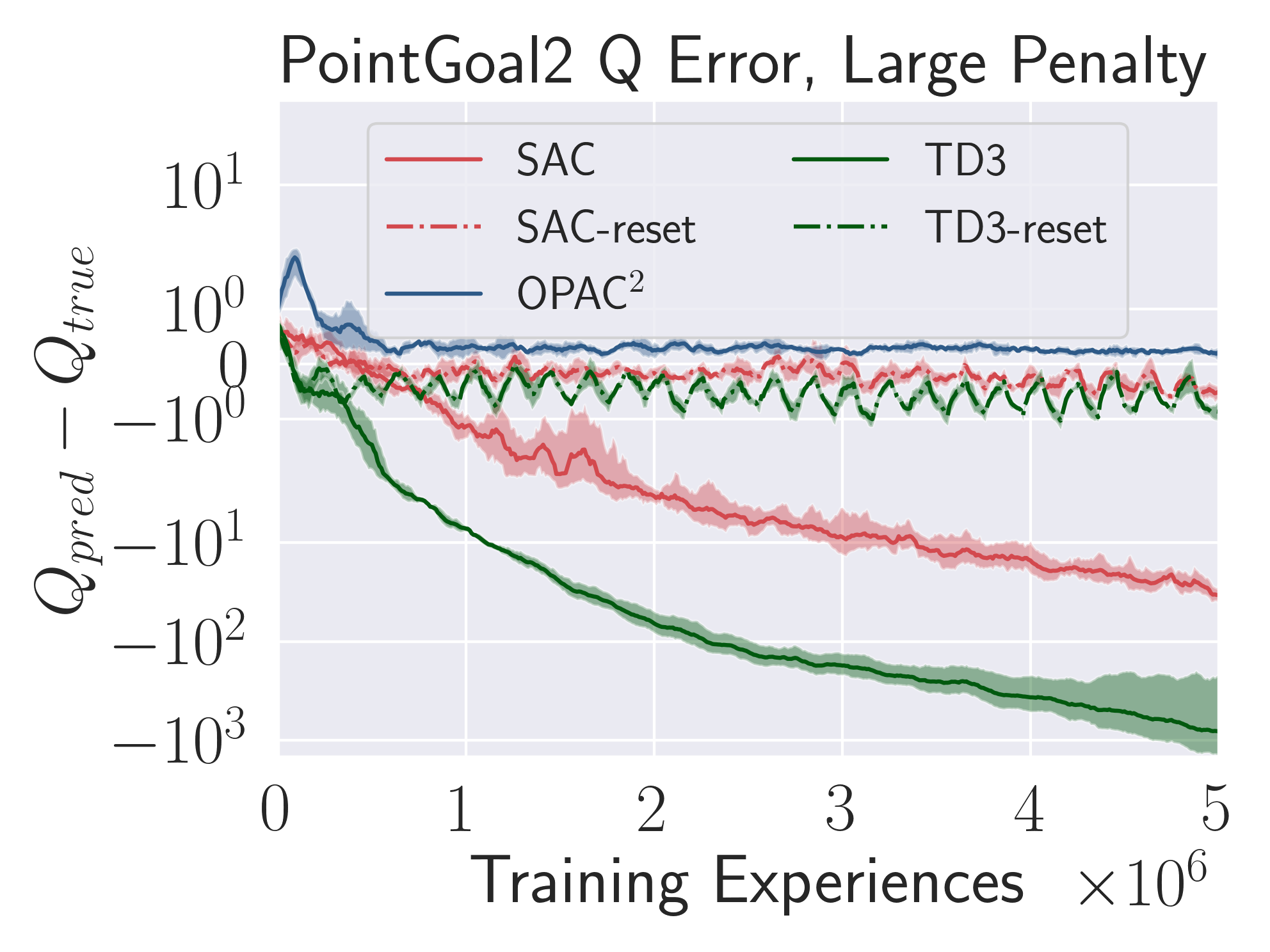}
\includegraphics[width=0.325\textwidth]{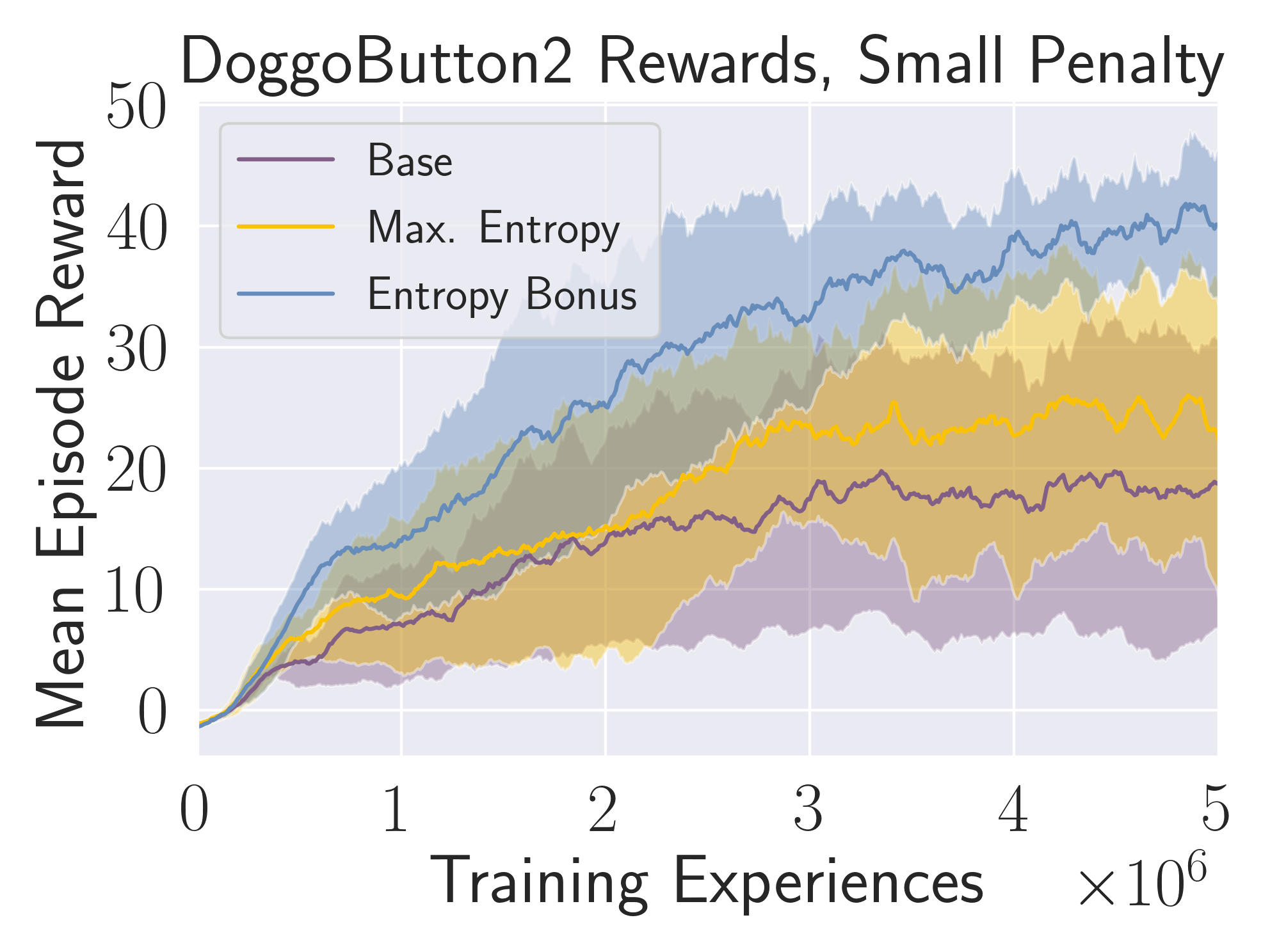}
\caption{Unconstrained learning with mixed-sign rewards. \textbf{Left}: TD error in validation is generally higher with SAC and TD3 than OPAC$^2$, but may be reduced with resetting. \textbf{Middle}: $Q$ is underestimated by SAC and TD3 in all ``Car'' and ``Point'' environments, but may be improved by resetting. \textbf{Right}: Our entropy bonus (blue) outperforms max-entropy and no regularization when used with OPAC$^2$ on the higher-dimensional Doggo control problems.}
\label{fig:diagnostics}
\end{figure}

To explore the issues affecting SAC and TD3, we examined the validation TD error as well as the $Q$ function error. As in \cite{li2023efficient}, ``validation'' refers to a held-out set of transitions not in the replay buffer. Figure \ref{fig:diagnostics} and Appendix \ref{unc_app} demonstrate that SAC and TD3 are prone to large TD error, and that periodic resets mitigate this. This phenomenon was observed by \cite{li2023efficient} in the context of learning at high replay ratios.  OPAC$^2$ is not nearly as prone to high TD error, and accordingly does not benefit from resets (Appendix \ref{reset_app}). 
% Resetting OPAC$^2$ actually results in slightly degraded performance in Safety Gym, presumably because the disruption in training caused by resetting outweighs the minimal gains to be had in more accurate value estimation (Appendix \ref{reset_app}). 
We also compared $Q$ function estimates to discounted Monte Carlo returns to measure $Q$ function accuracy.  We found that SAC and TD3 \emph{underestimate} the true $Q$ value in all ``Car'' and ``Point'' environments, while the trend was more variable in the ``Doggo'' environments. 
% This may be a result of the relative frequency of cost events early in ``Car'' and ``Point'' training compared to ``Doggo'' (where the robot must also learn to walk and where the entropy term in SAC may play a larger role).  

In the rightmost panel of Figure \ref{fig:diagnostics} and Appendix \ref{ent_app}, we display the efficacy of our entropy bonus strategy on ``Doggo'' environments. ``Doggo'' is a higher-dimensional control problem than ``Car'' and ``Point'' (12-dimensional vs. 2-dimensional), potentially leading it to require more exploration.  In all environments with mixed-sign rewards tested, we observed better performance with our entropy bonus than with a  maximum entropy approach. 
% We conjecture that this may be attributed to bias introduced into $Q$ and $V$ estimates by the inclusion of an entropy term also present in the policy update. 

\subsection{Constrained Learning}\label{constrained_results}
We further compared all methods in the constrained setting, again on Safety Gym. In all environments, we chose a target cost level equal to half the cost accumulated by a fully-trained TRPO agent unaware of cost (as reported by \cite{ray2019safetygym}). We chose this target to force the agent to strongly consider safely, but not be constrained to the point of being unable to complete the task. All agents learned separate value networks for reward and cost, to accommodate the variability of the penalty weight throughout training. We offer comparison with \textit{constrained versions} of SAC and TD3. 

\begin{figure}
    \centering
    \includegraphics[width=0.325\textwidth]{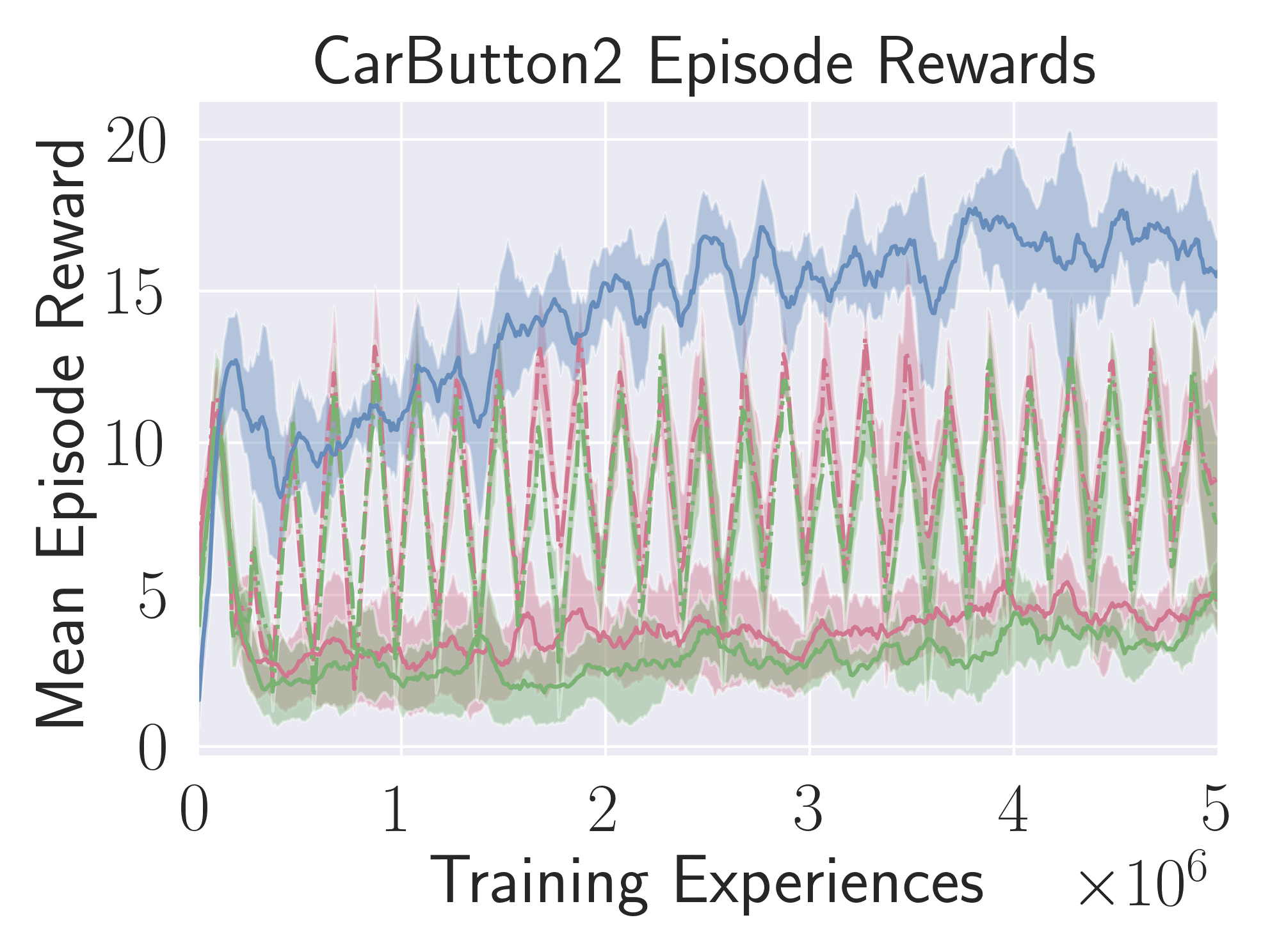}
    \includegraphics[width=0.325\textwidth]{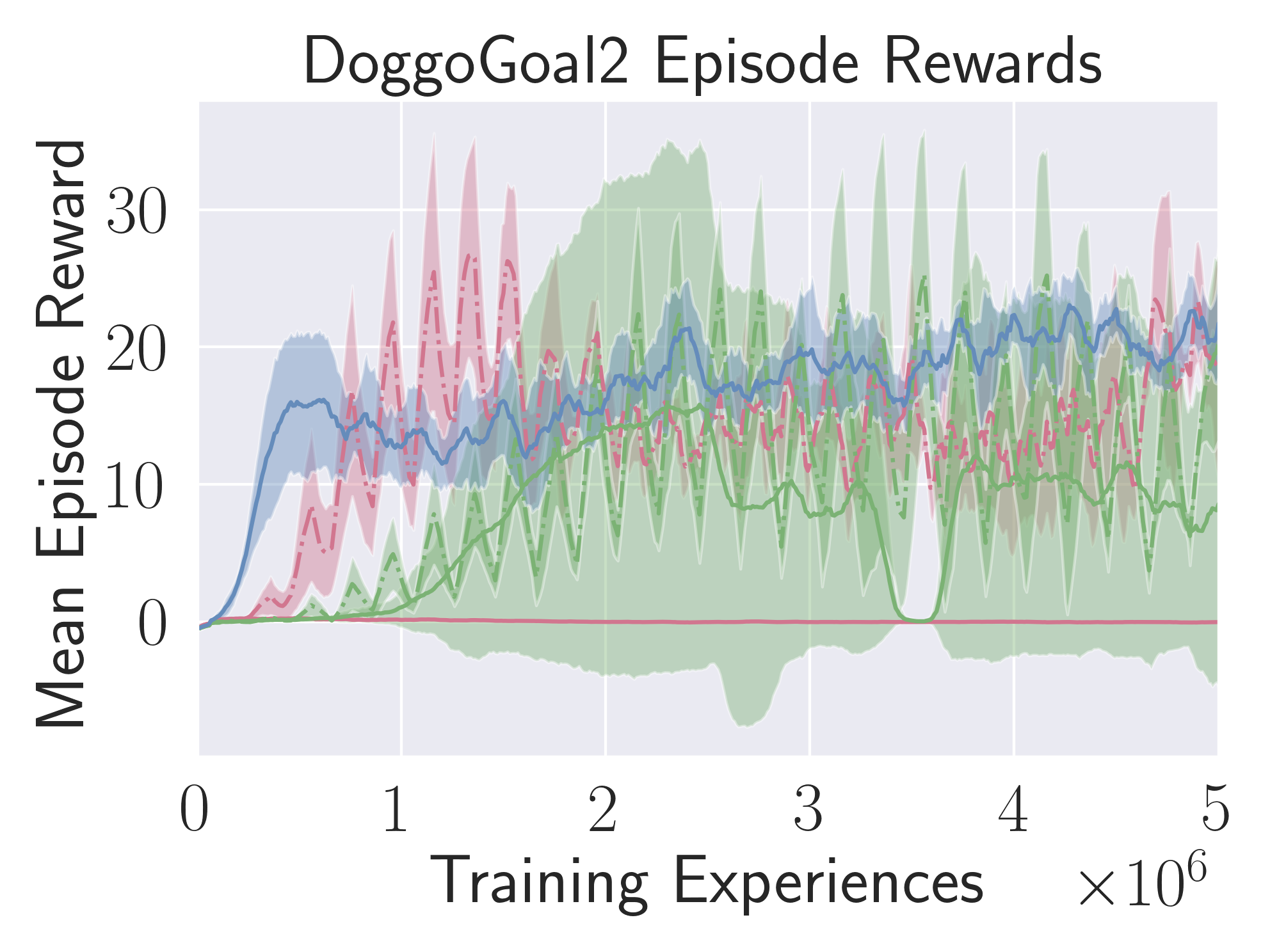}
    \includegraphics[width=0.325\textwidth]{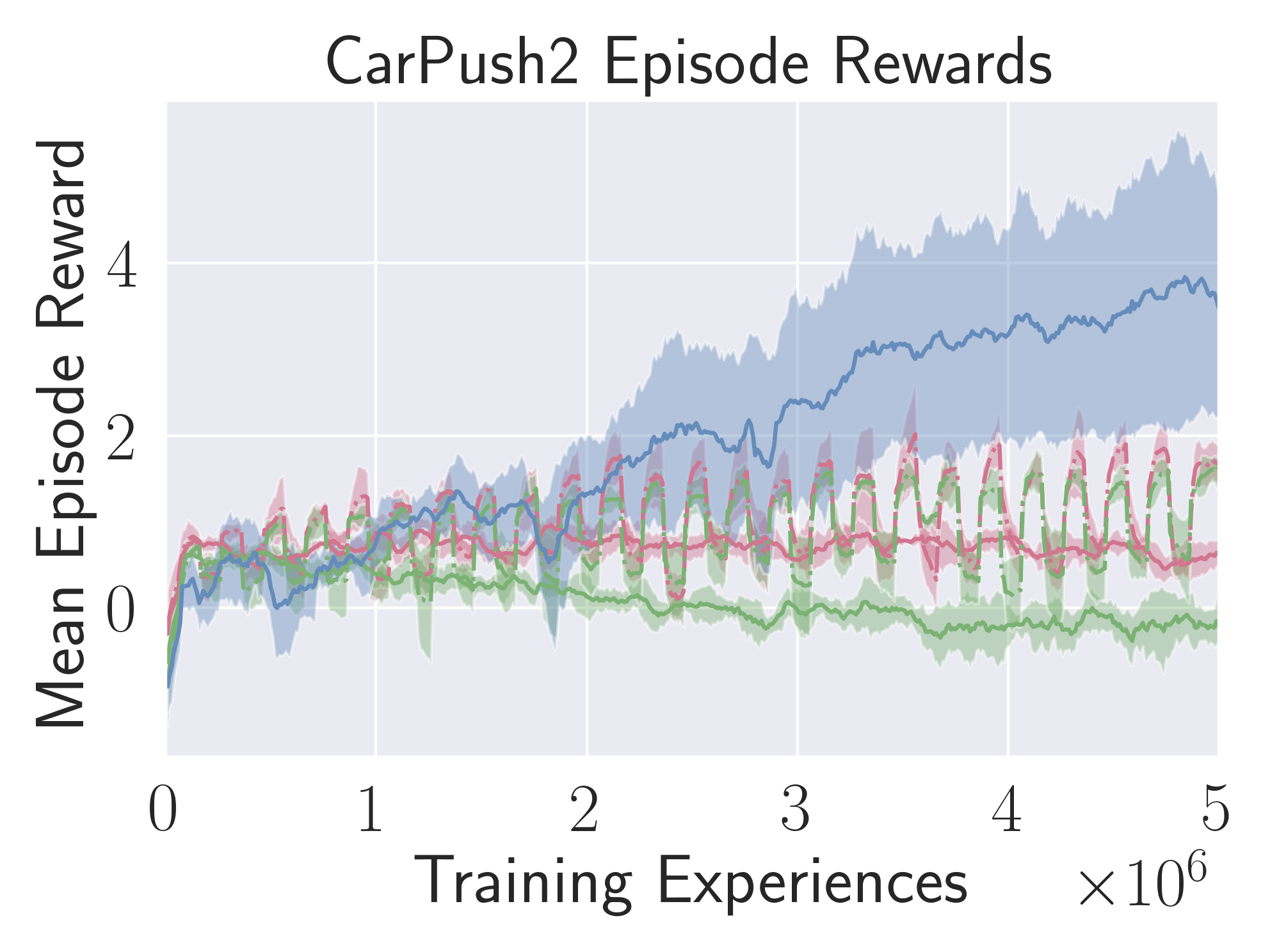}
    \includegraphics[width=0.325\textwidth]{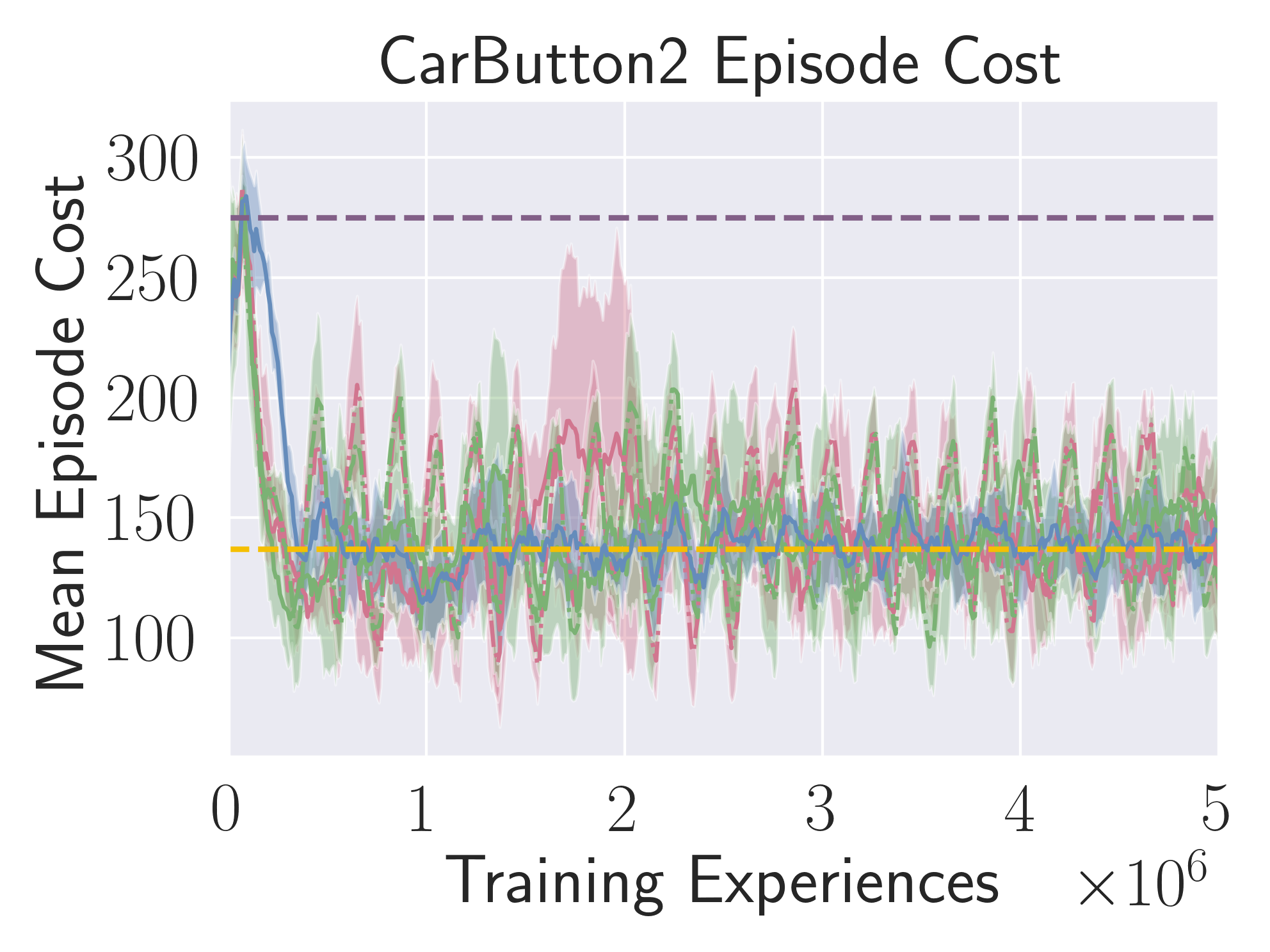}
    \includegraphics[width=0.325\textwidth]{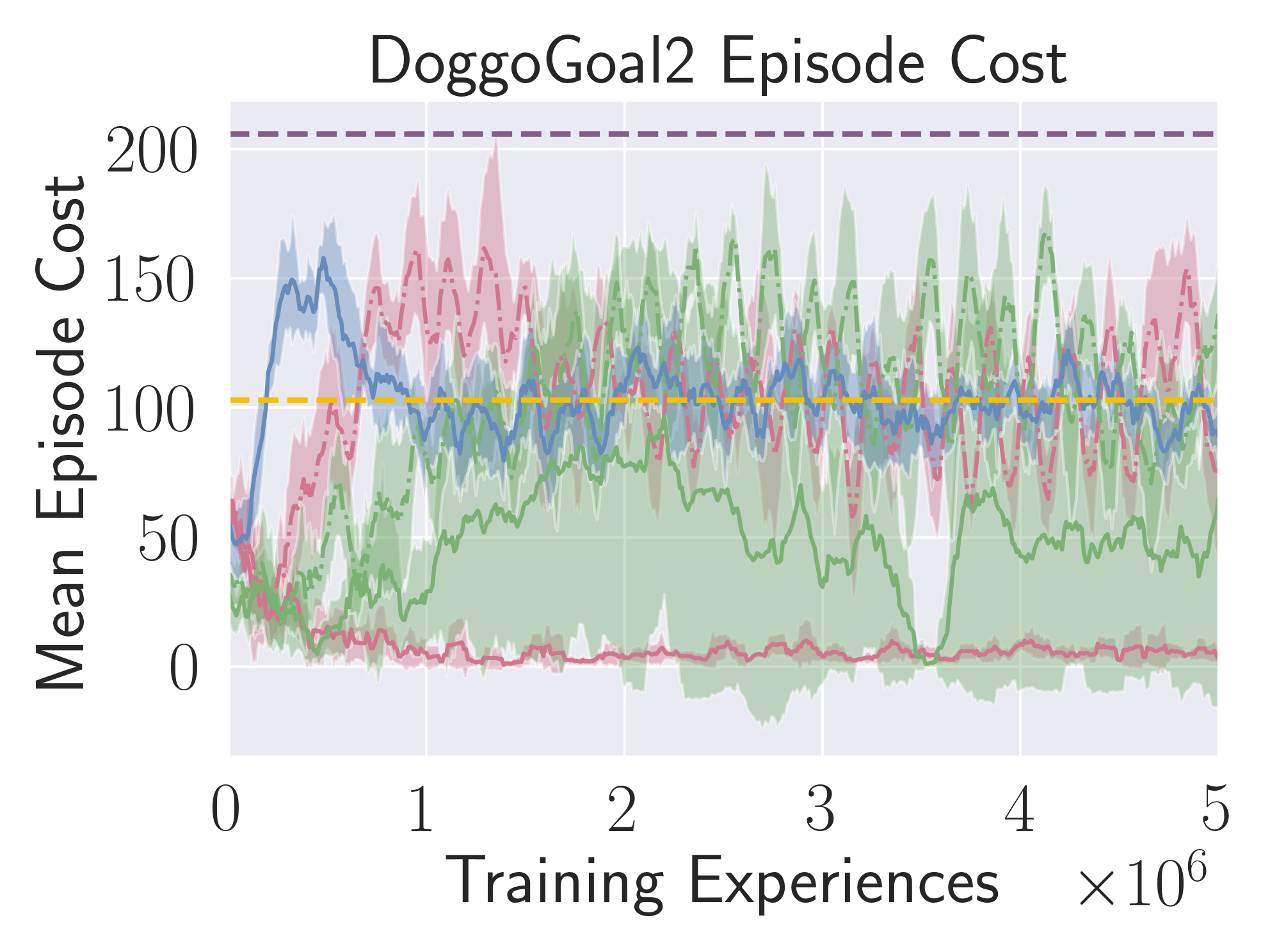}
    \includegraphics[width=0.325\textwidth]{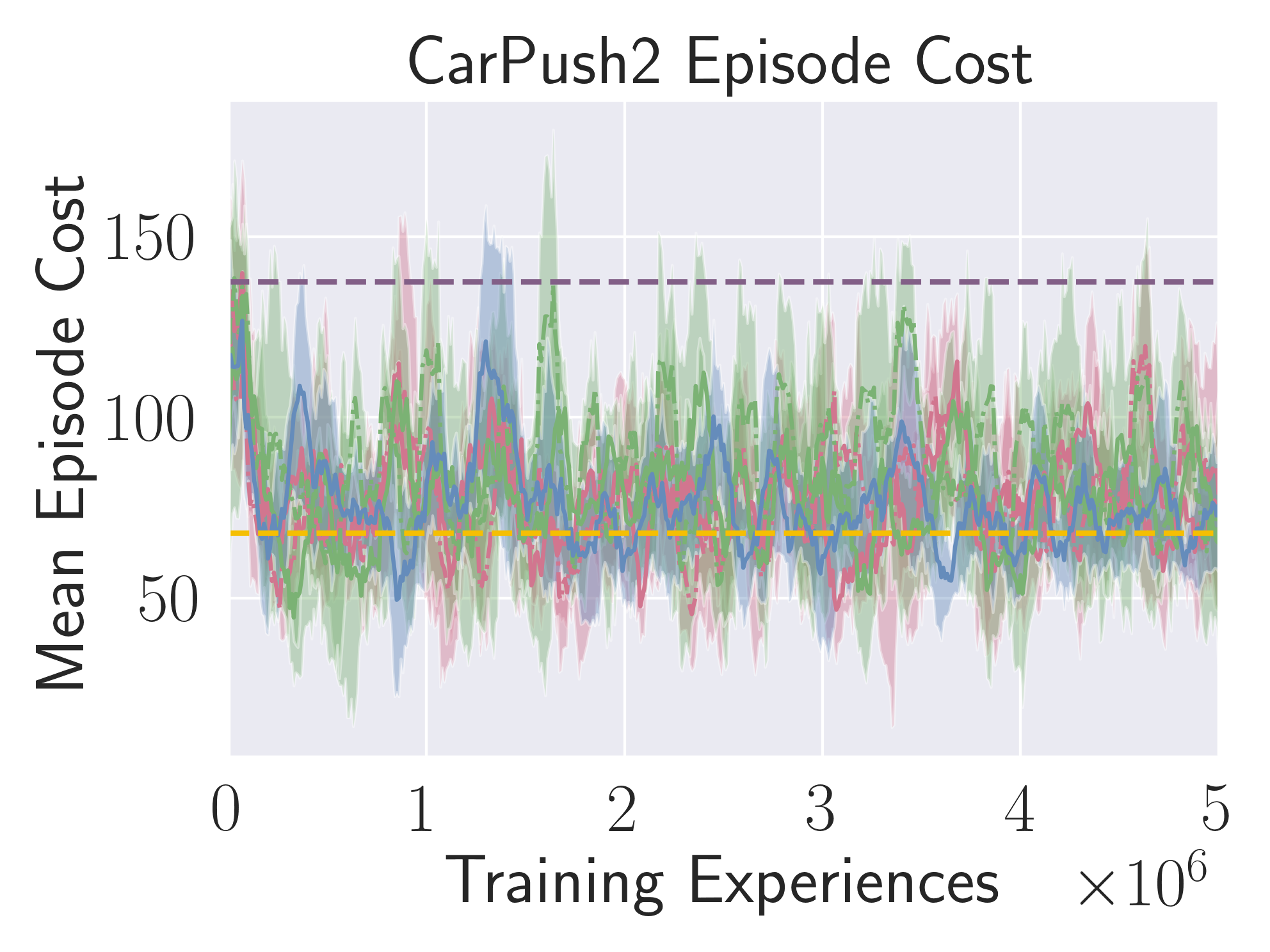}
    \includegraphics[width=.9\textwidth]{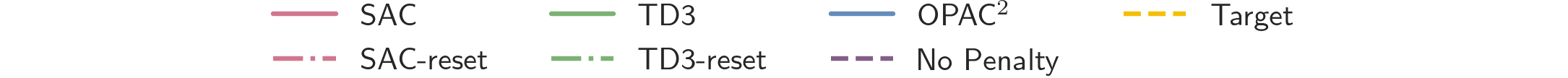}
    \caption{Representative sample of constrained learning on Safety Gym. \textbf{Top Row}: Positive reward accumulated by agents. Note that reward accumulation initially increases sharply while $\beta$, the penalty weight, is low. It dips and then rises again as $\beta$ increases and stabilizes, allowing performance to be optimized at the appropriate cost level. \textbf{Bottom Row}: Cost converges to the target level (yellow).}
    \label{fig:con_rew}
\end{figure}

A representative sample of constrained learning performance is given in Figure \ref{fig:con_rew}, with full results being provided in Appendix \ref{con_app}. We found all methods capable of reaching the target cost level, with OPAC$^2$ achieving the highest positive rewards in all environments tested.  Sample efficiency was again greatly improved relative to on-policy approaches, with performance matching that of state-of-the-art on-policy methods using roughly 50 times less data \citep{aaai_paper}. TD3 and SAC were both seen to benefit from resetting, allowing them to reduce the error on their cost and reward value estimates.  However, this improvement is not always enough to match the performance of OPAC$^2$. Finally, we note that this approach is highly configurable; the initial value and learning rate of the cost weight $\beta$ may be adjusted upward to provide faster convergence to the prescribed  cost 
 \begin{wrapfigure}[21]{r}{0.325\textwidth}
    \includegraphics[width=0.325\textwidth]{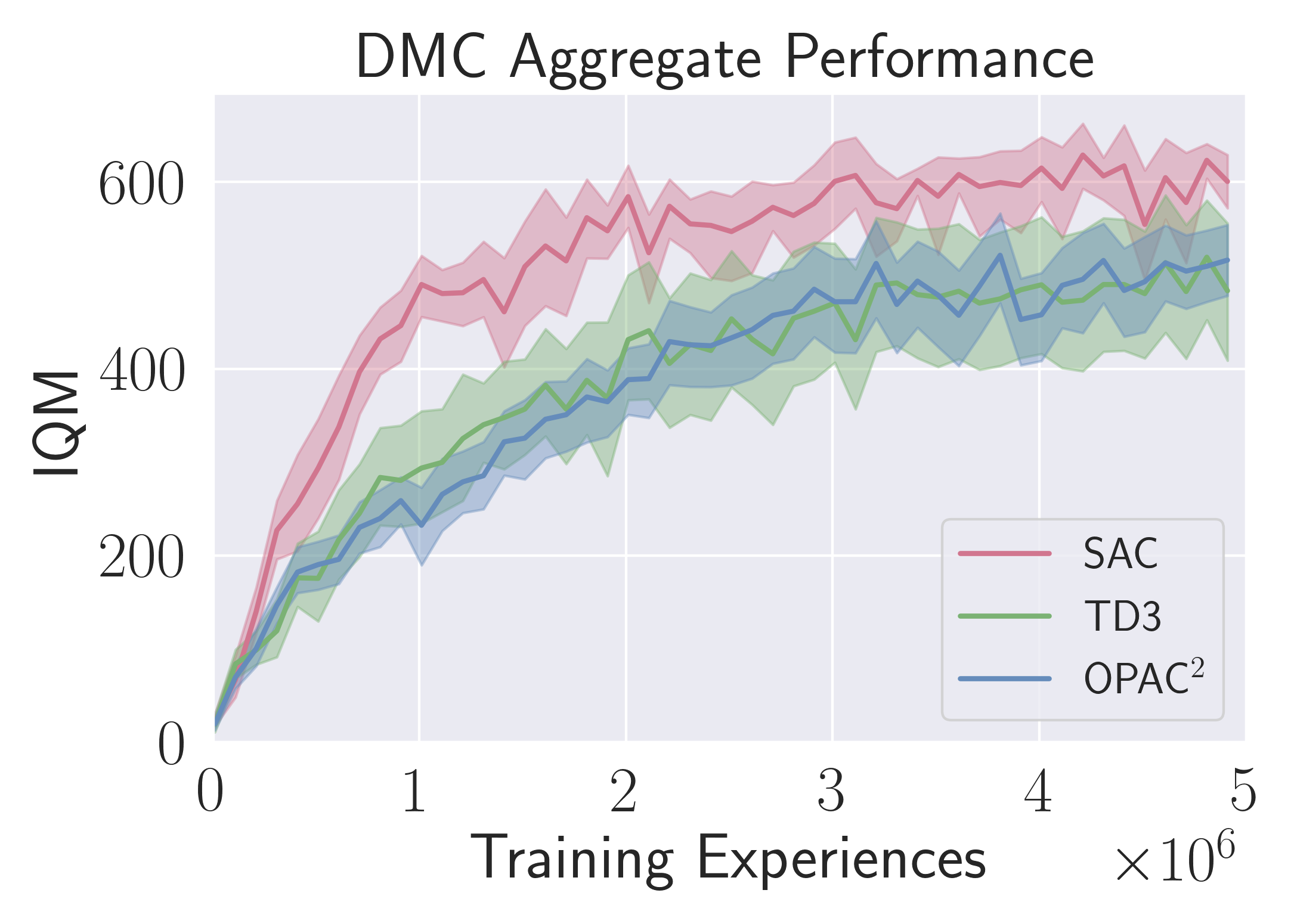}
     \includegraphics[width=0.325\textwidth]{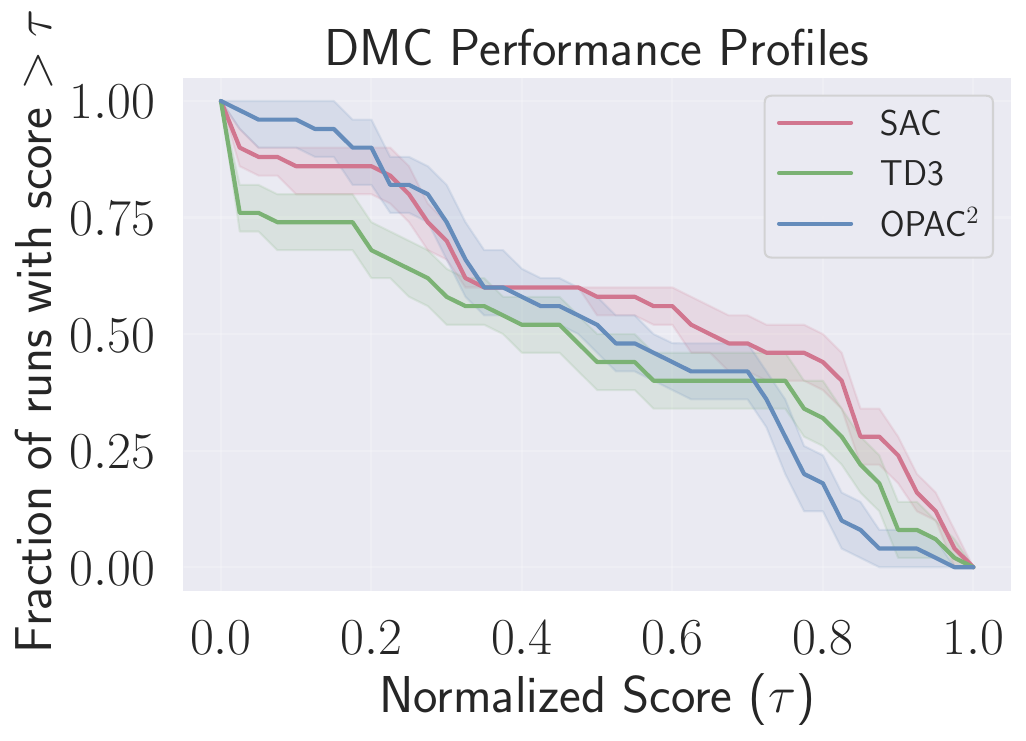}
    \caption{\textbf{Top}: Rewards aggregated over 10 tasks from the DeepMind Control Suite.  \textbf{Bottom}: Frequency of reward accumulation matching given performance levels.}
    \label{fig:dmc_agg}
\end{wrapfigure}
target or downward to enable more early exploration and potentially higher final positive rewards. 
\subsection{Evaluation on DeepMind Control}

Given the consistently strong performance of OPAC$^2$ in Safety Gym, we chose to additionally evaluate it on 10 tasks without mixed-sign rewards from the DeepMind Control Suite \citep{tunyasuvunakool2020dmc}. We used the same 10 tasks as the authors of \cite{nikishin2022primacy}, motivated by the fact that they provide some challenge for SAC.  Following the practice of \cite{rliable}, we aggregate performance across tasks using the interquartile mean (IQM), calculated as the mean score of the middle 50\% of runs.  We also computed performance profiles for each method, tabulating the fraction of experiments that exceeded each possible performance level.

As shown in Figure \ref{fig:dmc_agg}, the aggregated task performance of OPAC$^2$ is worse than SAC but on par with or slightly better than TD3 (full results in Appendix \ref{app:dmc_full}). This is to be expected, given the likely lower variance of the policy gradient estimates of SAC. However, OPAC$^2$ was found to be more reliable than the other methods at producing at least some learning in each environment, as evidenced by the left side of the performance profile plot.

\section{Related Work}
Our methods build on the off-policy actor-critic introduced in \cite{degris2012off}, applied in differing forms over the past decade \citep{wang2017sample, espeholt2018impala, gu2017qprop}, and summarized in \cite{levine_offline}. We compare it with more popular methods for off-policy DRL with continuous action spaces \citep{fujimoto2018td3, sac_newer}.  The resetting strategy that we employ was explored by \cite{nikishin2022primacy} and subsequently further evaluated and extended \citep{d'oro2023sampleefficient, schwarzer23aBBF}. Its ability to curtail TD error, as well as the role of TD error in limiting off-policy DRL more generally, was discussed in \cite{li2023efficient}.  To our knowledge, its impact on environments with mixed-sign rewards had not previously been quantified.

Constrained DRL has previously been explored on-policy \citep{BHATNAGAR2010760, achiam2017cpo, ray2019safetygym, Chow2019, TeMaMa19, gap, zhang20, aaai_paper, Moskovitz_etal_2023}, typically leveraging dual gradient descent \citep{bert, Boyd2004}. Constraints have recently been applied off-policy \citep{Ha_etal_2020, zhou_c_sac} for specific applications and with a distributional critic \citep{wcsac}. They have also been applied to probabilistically ensure safety \emph{during training} \citep{bharadhwaj2021conservative}, a useful requirement that we do not consider here. To our knowledge, this work is the first to apply constraints with the off-policy actor-critic, the first to explore the use of resetting in this context, and the first to specify constraints via consideration of only the most recent epoch in the replay buffer.

\section{Discussion}

The success of OPAC$^2$ relative to SAC and TD3 in environments with mixed-sign rewards underscores the necessity of managing function approximation error in off-policy deep reinforcement learning.  Off-policy DRL agents are known to be sensitive to experience gained early in training \citep{nikishin2022primacy}, tending to exploit any initial success they have. When multiple independent reward terms are present, this translates to some terms being favored over others.  In environments with mixed-sign rewards, the explicit competition between incentives and costs exacerbates this tendency. This effect is fundamentally a result of overfitting in $Q$ estimation.  While OPAC$^2$ does not fully eliminate $Q$ estimation errors, we empirically observe it to significantly dampen them.  Further, the likely higher variance of the policy gradient estimate in OPAC$^2$ may actually be beneficial in this setting: while not being large enough to preclude learning, it may help to address competition between reward terms by preventing the policy from converging too quickly to one that ignores some terms. This combination gives OPAC$^2$ a significant advantage over methods that update the policy to maximize the $Q$ function estimate when competing reward terms are present, even when the latter are periodically ``course corrected'' by resetting.

Several additional observations may be made based on these findings.  First, they argue for the use of a policy-gradient-based approach whenever mixed-sign rewards are present (including constrained learning).  This is particularly true when resetting is impractical, for instance when cumulative cost is a consideration or for offline learning.  When rewards do not include terms of mixed signs, we may expect OPAC$^2$ to perform competently and reliably, but often not as well as SAC. It is interesting to note that OPAC$^2$ showed strong learning on the \texttt{acrobot-swingup} task, the only DeepMind Control environment we tested that SAC and TD3 both failed to solve. Finally, we note that a policy-gradient-based approach may prove beneficial for multi-task or meta-learning, where competing reward terms from multiple tasks must be considered.  

\section{Conclusions}

In this work, we examine the tendency of state-of-the-art approaches to off-policy DRL for continuous action spaces to struggle when applied to environments with mixed-sign rewards. We elucidate the role of function approximation error in the process; in particular, the sensitivity of off-policy methods to experience gathered early in training may lead to the neglect of some terms of the reward function. To remedy the situation, we consider both a popular regularization strategy (periodic resetting) and a novel approach that produces more reliable value estimates (OPAC$^2$). Empirically we find the latter, which represents an update to the off-policy actor critic \citep{degris2012off} and related methods, to produce more performant learning in the presence of costs.

\section*{Acknowledgements}
The authors thank Cash Costello, Nathan Drenkow, and Ted Staley for useful conversations on related topics.

\section*{Ethics Statement}
In the interest of transparency and reproducibility, we provide full details necessary for reproducing experiments in the Appendix.  We will also post the source code publicly.
While our work is immediately targeted at making safer robotic agents, it is true that others could repurpose our code for malicious applications. 
We encourage the authors of any subsequent work to consider the societal impacts of future results.

\section*{Reproducibility Statement}

We provide detailed pseudocode of our constrained algorithm in Algorithm \ref{alg:copac2}, and for our unconstrained algorithm in Algorithm \ref{alg:opac2}.
In Appendix \ref{app:exp}, we provide hyperparameters, evaluation procedures, and implementation details necessary for replicating all experiments. 
Additionally, we will make the source code available publicly.

\bibliography{iclr2024_conference}

\begin{thebibliography}{44}
\providecommand{\natexlab}[1]{#1}
\providecommand{\url}[1]{\texttt{#1}}
\expandafter\ifx\csname urlstyle\endcsname\relax
  \providecommand{\doi}[1]{doi: #1}\else
  \providecommand{\doi}{doi: \begingroup \urlstyle{rm}\Url}\fi

\bibitem[Achiam et~al.(2017)Achiam, Held, Tamar, and Abbeel]{achiam2017cpo}
Joshua Achiam, David Held, Aviv Tamar, and Pieter Abbeel.
\newblock Constrained policy optimization.
\newblock In Doina Precup and Yee~Whye Teh (eds.), \emph{Proceedings of the
  34th International Conference on Machine Learning}, volume~70 of
  \emph{Proceedings of Machine Learning Research}, pp.\  22--31. PMLR, 06--11
  Aug 2017.
\newblock URL \url{https://proceedings.mlr.press/v70/achiam17a.html}.

\bibitem[Agarwal et~al.(2021)Agarwal, Schwarzer, Castro, Courville, and
  Bellemare]{rliable}
Rishabh Agarwal, Max Schwarzer, Pablo~Samuel Castro, Aaron Courville, and
  Marc~G Bellemare.
\newblock Deep reinforcement learning at the edge of the statistical precipice.
\newblock \emph{Advances in Neural Information Processing Systems}, 2021.

\bibitem[Akkaya et~al.(2019)Akkaya, Andrychowicz, Chociej, Litwin, McGrew,
  Petron, Paino, Plappert, Powell, Ribas, Schneider, Tezak, Tworek, Welinder,
  Weng, Yuan, Zaremba, and Zhang]{openai2019solving}
Ilge Akkaya, Marcin Andrychowicz, Maciek Chociej, Mateusz Litwin, Bob McGrew,
  Arthur Petron, Alex Paino, Matthias Plappert, Glenn Powell, Raphael Ribas,
  Jonas Schneider, Nikolas Tezak, Jerry Tworek, Peter Welinder, Lilian Weng,
  Qiming Yuan, Wojciech Zaremba, and Lei Zhang.
\newblock Solving rubik's cube with a robot hand, 2019.

\bibitem[Bertsekas(1996)]{bert}
Dimitri~P. Bertsekas.
\newblock \emph{Constrained Optimization and Lagrange Multiplier Methods
  (Optimization and Neural Computation Series)}.
\newblock Athena Scientific, 1 edition, 1996.
\newblock ISBN 1886529043.

\bibitem[Bharadhwaj et~al.(2021)Bharadhwaj, Kumar, Rhinehart, Levine, Shkurti,
  and Garg]{bharadhwaj2021conservative}
Homanga Bharadhwaj, Aviral Kumar, Nicholas Rhinehart, Sergey Levine, Florian
  Shkurti, and Animesh Garg.
\newblock Conservative safety critics for exploration.
\newblock In \emph{International Conference on Learning Representations}, 2021.
\newblock URL \url{https://openreview.net/forum?id=iaO86DUuKi}.

\bibitem[Bhatnagar(2010)]{BHATNAGAR2010760}
Shalabh Bhatnagar.
\newblock An actor–critic algorithm with function approximation for
  discounted cost constrained markov decision processes.
\newblock \emph{Systems \& Control Letters}, 59\penalty0 (12):\penalty0
  760--766, 2010.
\newblock ISSN 0167-6911.
\newblock \doi{https://doi.org/10.1016/j.sysconle.2010.08.013}.
\newblock URL
  \url{https://www.sciencedirect.com/science/article/pii/S0167691110001246}.

\bibitem[Boyd \& Vandenberghe(2004)Boyd and Vandenberghe]{Boyd2004}
Stephen~P Boyd and Lieven Vandenberghe.
\newblock \emph{Convex optimization}.
\newblock Cambridge university press, 2004.

\bibitem[Chen et~al.(2021)Chen, Wang, Zhou, and Ross]{redq}
Xinyue Chen, Che Wang, Zijian Zhou, and Keith~W. Ross.
\newblock Randomized ensembled double q-learning: Learning fast without a
  model.
\newblock In \emph{International Conference on Learning Representations}, 2021.
\newblock URL \url{https://openreview.net/forum?id=AY8zfZm0tDd}.

\bibitem[Chow et~al.(2019)Chow, Nachum, Faust, Ghavamzadeh, and
  Du{\'{e}}{\~{n}}ez{-}Guzm{\'{a}}n]{Chow2019}
Yinlam Chow, Ofir Nachum, Aleksandra Faust, Mohammad Ghavamzadeh, and Edgar~A.
  Du{\'{e}}{\~{n}}ez{-}Guzm{\'{a}}n.
\newblock Lyapunov-based safe policy optimization for continuous control.
\newblock \emph{CoRR}, abs/1901.10031, 2019.
\newblock URL \url{http://arxiv.org/abs/1901.10031}.

\bibitem[Degrave et~al.(2022)Degrave, Felici, Buchli, Neunert, Tracey,
  Carpanese, Ewalds, Hafner, Abdolmaleki, de~las Casas, Donner, Fritz,
  Galperti, Huber, Keeling, Tsimpoukelli, Kay, Merle, Moret, Noury, Pesamosca,
  Pfau, Sauter, Sommariva, Coda, Duval, Fasoli, Kohli, Kavukcuoglu, Hassabis,
  and Riedmiller]{degrave_magnetic_2022}
Jonas Degrave, Federico Felici, Jonas Buchli, Michael Neunert, Brendan Tracey,
  Francesco Carpanese, Timo Ewalds, Roland Hafner, Abbas Abdolmaleki, Diego
  de~las Casas, Craig Donner, Leslie Fritz, Cristian Galperti, Andrea Huber,
  James Keeling, Maria Tsimpoukelli, Jackie Kay, Antoine Merle, Jean-Marc
  Moret, Seb Noury, Federico Pesamosca, David Pfau, Olivier Sauter, Cristian
  Sommariva, Stefano Coda, Basil Duval, Ambrogio Fasoli, Pushmeet Kohli, Koray
  Kavukcuoglu, Demis Hassabis, and Martin Riedmiller.
\newblock Magnetic control of tokamak plasmas through deep reinforcement
  learning.
\newblock \emph{Nature}, 602\penalty0 (7897):\penalty0 414--419, February 2022.
\newblock ISSN 1476-4687.
\newblock \doi{10.1038/s41586-021-04301-9}.
\newblock URL \url{https://doi.org/10.1038/s41586-021-04301-9}.

\bibitem[Degris et~al.(2012)Degris, White, and Sutton]{degris2012off}
Thomas Degris, Martha White, and Richard~S Sutton.
\newblock Off-policy actor-critic.
\newblock In \emph{International Conference on Machine Learning}, 2012.

\bibitem[D'Oro et~al.(2023)D'Oro, Schwarzer, Nikishin, Bacon, Bellemare, and
  Courville]{d'oro2023sampleefficient}
Pierluca D'Oro, Max Schwarzer, Evgenii Nikishin, Pierre-Luc Bacon, Marc~G
  Bellemare, and Aaron Courville.
\newblock Sample-efficient reinforcement learning by breaking the replay ratio
  barrier.
\newblock In \emph{The Eleventh International Conference on Learning
  Representations}, 2023.
\newblock URL \url{https://openreview.net/forum?id=OpC-9aBBVJe}.

\bibitem[Espeholt et~al.(2018)Espeholt, Soyer, Munos, Simonyan, Mnih, Ward,
  Doron, Firoiu, Harley, Dunning, Legg, and Kavukcuoglu]{espeholt2018impala}
Lasse Espeholt, Hubert Soyer, Remi Munos, Karen Simonyan, Volodymir Mnih, Tom
  Ward, Yotam Doron, Vlad Firoiu, Tim Harley, Iain Dunning, Shane Legg, and
  Koray Kavukcuoglu.
\newblock Impala: Scalable distributed deep-rl with importance weighted
  actor-learner architectures, 2018.

\bibitem[Fu et~al.(2019)Fu, Kumar, Soh, and Levine]{fu2019}
Justin Fu, Aviral Kumar, Matthew Soh, and Sergey Levine.
\newblock Diagnosing bottlenecks in deep q-learning algorithms.
\newblock In Kamalika Chaudhuri and Ruslan Salakhutdinov (eds.),
  \emph{Proceedings of the 36th International Conference on Machine Learning},
  volume~97 of \emph{Proceedings of Machine Learning Research}, pp.\
  2021--2030. PMLR, 09--15 Jun 2019.
\newblock URL \url{https://proceedings.mlr.press/v97/fu19a.html}.

\bibitem[Fujimoto et~al.(2018)Fujimoto, van Hoof, and Meger]{fujimoto2018td3}
Scott Fujimoto, Herke van Hoof, and David Meger.
\newblock Addressing function approximation error in actor-critic methods.
\newblock In Jennifer Dy and Andreas Krause (eds.), \emph{Proceedings of the
  35th International Conference on Machine Learning}, volume~80 of
  \emph{Proceedings of Machine Learning Research}, pp.\  1587--1596. PMLR,
  10--15 Jul 2018.
\newblock URL \url{https://proceedings.mlr.press/v80/fujimoto18a.html}.

\bibitem[Fujita \& Maeda(2018)Fujita and Maeda]{clipped_action_policy_gradient}
Yasuhiro Fujita and Shin-ichi Maeda.
\newblock Clipped action policy gradient.
\newblock In Jennifer Dy and Andreas Krause (eds.), \emph{Proceedings of the
  35th International Conference on Machine Learning}, volume~80 of
  \emph{Proceedings of Machine Learning Research}, pp.\  1597--1606. PMLR,
  10--15 Jul 2018.
\newblock URL \url{https://proceedings.mlr.press/v80/fujita18a.html}.

\bibitem[Gu et~al.(2017)Gu, Lillicrap, Ghahramani, Turner, and
  Levine]{gu2017qprop}
Shixiang Gu, Timothy Lillicrap, Zoubin Ghahramani, Richard~E. Turner, and
  Sergey Levine.
\newblock Q-prop: Sample-efficient policy gradient with an off-policy critic.
\newblock In \emph{International Conference on Learning Representations}, 2017.
\newblock URL \url{https://openreview.net/forum?id=SJ3rcZcxl}.

\bibitem[Ha et~al.(2020)Ha, Xu, Tan, Levine, and Tan]{Ha_etal_2020}
Sehoon Ha, Peng Xu, Zhenyu Tan, Sergey Levine, and Jie Tan.
\newblock Learning to walk in the real world with minimal human effort.
\newblock \emph{arXiv preprint arXiv:2002.08550}, 2020.

\bibitem[Haarnoja et~al.(2018)Haarnoja, Zhou, Abbeel, and Levine]{sac_original}
Tuomas Haarnoja, Aurick Zhou, Pieter Abbeel, and Sergey Levine.
\newblock Soft actor-critic: Off-policy maximum entropy deep reinforcement
  learning with a stochastic actor.
\newblock In Jennifer~G. Dy and Andreas Krause (eds.), \emph{Proceedings of the
  35th International Conference on Machine Learning, {ICML} 2018,
  Stockholmsm{\"{a}}ssan, Stockholm, Sweden, July 10-15, 2018}, volume~80 of
  \emph{Proceedings of Machine Learning Research}, pp.\  1856--1865. {PMLR},
  2018.
\newblock URL \url{http://proceedings.mlr.press/v80/haarnoja18b.html}.

\bibitem[Haarnoja et~al.(2019)Haarnoja, Zhou, Hartikainen, Tucker, Ha, Tan,
  Kumar, Zhu, Gupta, Abbeel, and Levine]{sac_newer}
Tuomas Haarnoja, Aurick Zhou, Kristian Hartikainen, George Tucker, Sehoon Ha,
  Jie Tan, Vikash Kumar, Henry Zhu, Abhishek Gupta, Pieter Abbeel, and Sergey
  Levine.
\newblock Soft actor-critic algorithms and applications, 2019.

\bibitem[Hasselt(2010)]{double_q}
Hado Hasselt.
\newblock Double q-learning.
\newblock In J.~Lafferty, C.~Williams, J.~Shawe-Taylor, R.~Zemel, and
  A.~Culotta (eds.), \emph{Advances in Neural Information Processing Systems},
  volume~23. Curran Associates, Inc., 2010.
\newblock URL
  \url{https://proceedings.neurips.cc/paper_files/paper/2010/file/091d584fced301b442654dd8c23b3fc9-Paper.pdf}.

\bibitem[Hiraoka et~al.(2022)Hiraoka, Imagawa, Hashimoto, Onishi, and
  Tsuruoka]{droq}
Takuya Hiraoka, Takahisa Imagawa, Taisei Hashimoto, Takashi Onishi, and
  Yoshimasa Tsuruoka.
\newblock Dropout q-functions for doubly efficient reinforcement learning.
\newblock In \emph{International Conference on Learning Representations}, 2022.
\newblock URL \url{https://openreview.net/forum?id=xCVJMsPv3RT}.

\bibitem[Kingma \& Ba(2015)Kingma and Ba]{KiBa17}
Diederik~P. Kingma and Jimmy Ba.
\newblock Adam: {A} method for stochastic optimization.
\newblock In \emph{3rd International Conference on Learning Representations},
  2015.

\bibitem[Kingma et~al.(2015)Kingma, Salimans, and Welling]{reparam}
Diederik~P. Kingma, Tim Salimans, and Max Welling.
\newblock Variational dropout and the local reparameterization trick, 2015.

\bibitem[Levine et~al.(2020)Levine, Kumar, Tucker, and Fu]{levine_offline}
Sergey Levine, Aviral Kumar, George Tucker, and Justin Fu.
\newblock Offline reinforcement learning: Tutorial, review, and perspectives on
  open problems, 2020.

\bibitem[Li et~al.(2023)Li, Kumar, Kostrikov, and Levine]{li2023efficient}
Qiyang Li, Aviral Kumar, Ilya Kostrikov, and Sergey Levine.
\newblock Efficient deep reinforcement learning requires regulating
  overfitting.
\newblock In \emph{The Eleventh International Conference on Learning
  Representations (ICLR)}, 2023.
\newblock URL \url{https://openreview.net/forum?id=14-kr46GvP-}.

\bibitem[Liu et~al.(2021)Liu, Liu, Jin, Stone, and Liu]{cagrad}
Bo~Liu, Xingchao Liu, Xiaojie Jin, Peter Stone, and Qiang Liu.
\newblock Conflict-averse gradient descent for multi-task learning.
\newblock In M.~Ranzato, A.~Beygelzimer, Y.~Dauphin, P.S. Liang, and J.~Wortman
  Vaughan (eds.), \emph{Advances in Neural Information Processing Systems},
  volume~34, pp.\  18878--18890. Curran Associates, Inc., 2021.
\newblock URL
  \url{https://proceedings.neurips.cc/paper_files/paper/2021/file/9d27fdf2477ffbff837d73ef7ae23db9-Paper.pdf}.

\bibitem[Markowitz et~al.(2023)Markowitz, Gardner, Llorens, Arora, and
  Wang]{aaai_paper}
Jared Markowitz, Ryan~W. Gardner, Ashley Llorens, Raman Arora, and I-Jeng Wang.
\newblock A risk-sensitive approach to policy optimization.
\newblock \emph{Proceedings of the AAAI Conference on Artificial Intelligence},
  37\penalty0 (12):\penalty0 15019--15027, Jun. 2023.
\newblock \doi{10.1609/aaai.v37i12.26753}.
\newblock URL \url{https://ojs.aaai.org/index.php/AAAI/article/view/26753}.

\bibitem[Mnih et~al.(2015)Mnih, Kavukcuoglu, Silver, Rusu, Veness, Bellemare,
  Graves, Riedmiller, Fidjeland, Ostrovski, Petersen, Beattie, Sadik,
  Antonoglou, King, Kumaran, Wierstra, Legg, and
  Hassabis]{MnKaSiRuVeBeGrRiFiOsPeBeAnKiKuWiLeHa15}
Volodymyr Mnih, Koray Kavukcuoglu, David Silver, Andrei~A. Rusu, Joel Veness,
  Marc~G. Bellemare, Alex Graves, Martin Riedmiller, Andreas~K. Fidjeland,
  Georg Ostrovski, Stig Petersen, Charles Beattie, Amir Sadik, Ioannis
  Antonoglou, Helen King, Dharshan Kumaran, Daan Wierstra, Shane Legg, and
  Demis Hassabis.
\newblock Human-level control through deep reinforcement learning.
\newblock \emph{Nature}, 518\penalty0 (7540):\penalty0 529--533, 2015.

\bibitem[Moskovitz et~al.(2023)Moskovitz, O’Donoghue, Veeriah, Flennerhag,
  Singh, and Zahavy]{Moskovitz_etal_2023}
Ted Moskovitz, Brendan O’Donoghue, Vivek Veeriah, Sebastian Flennerhag,
  Satinder Singh, and Tom Zahavy.
\newblock Reload: Reinforcement learning with optimistic ascent-descent for
  last-iterate convergence in constrained mdps.
\newblock In \emph{International Conference on Machine Learning}, pp.\
  25303--25336. PMLR, 2023.

\bibitem[Nikishin et~al.(2022)Nikishin, Schwarzer, D'Oro, Bacon, and
  Courville]{nikishin2022primacy}
Evgenii Nikishin, Max Schwarzer, Pierluca D'Oro, Pierre-Luc Bacon, and Aaron
  Courville.
\newblock The primacy bias in deep reinforcement learning.
\newblock In Kamalika Chaudhuri, Stefanie Jegelka, Le~Song, Csaba Szepesvari,
  Gang Niu, and Sivan Sabato (eds.), \emph{Proceedings of the 39th
  International Conference on Machine Learning}, volume 162 of
  \emph{Proceedings of Machine Learning Research}, pp.\  16828--16847. PMLR,
  17--23 Jul 2022.
\newblock URL \url{https://proceedings.mlr.press/v162/nikishin22a.html}.

\bibitem[Paternain et~al.(2019)Paternain, Chamon, Calvo-Fullana, and
  Ribeiro]{gap}
Santiago Paternain, Luiz Chamon, Miguel Calvo-Fullana, and Alejandro Ribeiro.
\newblock Constrained reinforcement learning has zero duality gap.
\newblock In \emph{Advances in Neural Information Processing Systems},
  volume~32. Curran Associates, Inc., 2019.
\newblock URL
  \url{https://proceedings.neurips.cc/paper_files/paper/2019/file/c1aeb6517a1c7f33514f7ff69047e74e-Paper.pdf}.

\bibitem[Ray et~al.(2019)Ray, Achiam, and Amodei]{ray2019safetygym}
Alex Ray, Joshua Achiam, and Dario Amodei.
\newblock {Benchmarking Safe Exploration in Deep Reinforcement Learning}.
\newblock 2019.

\bibitem[Schulman et~al.(2017{\natexlab{a}})Schulman, Levine, Moritz, Jordan,
  and Abbeel]{schulman2017trust}
John Schulman, Sergey Levine, Philipp Moritz, Michael~I. Jordan, and Pieter
  Abbeel.
\newblock Trust region policy optimization.
\newblock In \emph{Proceedings of the 32nd International Conference on Machine
  Learning (ICML)}, 2017{\natexlab{a}}.

\bibitem[Schulman et~al.(2017{\natexlab{b}})Schulman, Wolski, Dhariwal,
  Radford, and Klimov]{ScWoDhRaKl17}
John Schulman, Filip Wolski, Prafulla Dhariwal, Alec Radford, and Oleg Klimov.
\newblock Proximal policy optimization algorithms.
\newblock arXiv:1707.06347, 2017{\natexlab{b}}.

\bibitem[Schwarzer et~al.(2023)Schwarzer, Obando~Ceron, Courville, Bellemare,
  Agarwal, and Castro]{schwarzer23aBBF}
Max Schwarzer, Johan~Samir Obando~Ceron, Aaron Courville, Marc~G Bellemare,
  Rishabh Agarwal, and Pablo~Samuel Castro.
\newblock Bigger, better, faster: Human-level {A}tari with human-level
  efficiency.
\newblock In Andreas Krause, Emma Brunskill, Kyunghyun Cho, Barbara Engelhardt,
  Sivan Sabato, and Jonathan Scarlett (eds.), \emph{Proceedings of the 40th
  International Conference on Machine Learning}, volume 202 of
  \emph{Proceedings of Machine Learning Research}, pp.\  30365--30380. PMLR,
  23--29 Jul 2023.
\newblock URL \url{https://proceedings.mlr.press/v202/schwarzer23a.html}.

\bibitem[Silver et~al.(2014)Silver, Lever, Heess, Degris, Wierstra, and
  Riedmiller]{dpg}
David Silver, Guy Lever, Nicolas Heess, Thomas Degris, Daan Wierstra, and
  Martin~A. Riedmiller.
\newblock Deterministic policy gradient algorithms.
\newblock In \emph{ICML}, volume~32 of \emph{JMLR Workshop and Conference
  Proceedings}, pp.\  387--395. JMLR.org, 2014.
\newblock URL
  \url{http://dblp.uni-trier.de/db/conf/icml/icml2014.html#SilverLHDWR14}.

\bibitem[Tessler et~al.(2019)Tessler, Mankowitz, and Mannor]{TeMaMa19}
Chen Tessler, Daniel~J. Mankowitz, and Shie Mannor.
\newblock Reward constrained policy optimization.
\newblock In \emph{Proceedings of the International Conference on Learning
  Representations (ICLR)}, 2019.

\bibitem[Tunyasuvunakool et~al.(2020)Tunyasuvunakool, Muldal, Doron, Liu,
  Bohez, Merel, Erez, Lillicrap, Heess, and Tassa]{tunyasuvunakool2020dmc}
Saran Tunyasuvunakool, Alistair Muldal, Yotam Doron, Siqi Liu, Steven Bohez,
  Josh Merel, Tom Erez, Timothy Lillicrap, Nicolas Heess, and Yuval Tassa.
\newblock dm\_control: Software and tasks for continuous control.
\newblock \emph{Software Impacts}, 6:\penalty0 100022, 2020.
\newblock ISSN 2665-9638.
\newblock \doi{https://doi.org/10.1016/j.simpa.2020.100022}.
\newblock URL
  \url{https://www.sciencedirect.com/science/article/pii/S2665963820300099}.

\bibitem[Wang et~al.(2017)Wang, Bapst, Heess, Mnih, Munos, Kavukcuoglu, and
  de~Freitas]{wang2017sample}
Ziyu Wang, Victor Bapst, Nicolas Heess, Volodymyr Mnih, Remi Munos, Koray
  Kavukcuoglu, and Nando de~Freitas.
\newblock Sample efficient actor-critic with experience replay.
\newblock In \emph{International Conference on Learning Representations}, 2017.
\newblock URL \url{https://openreview.net/forum?id=HyM25Mqel}.

\bibitem[Yang et~al.(2021)Yang, Simão, Tindemans, and Spaan]{wcsac}
Qisong Yang, Thiago~D. Simão, Simon~H Tindemans, and Matthijs T.~J. Spaan.
\newblock Wcsac: Worst-case soft actor critic for safety-constrained
  reinforcement learning.
\newblock \emph{Proceedings of the AAAI Conference on Artificial Intelligence},
  35\penalty0 (12):\penalty0 10639--10646, May 2021.
\newblock \doi{10.1609/aaai.v35i12.17272}.
\newblock URL \url{https://ojs.aaai.org/index.php/AAAI/article/view/17272}.

\bibitem[Yu et~al.(2020)Yu, Kumar, Gupta, Levine, Hausman, and
  Finn]{grad_surgery}
Tianhe Yu, Saurabh Kumar, Abhishek Gupta, Sergey Levine, Karol Hausman, and
  Chelsea Finn.
\newblock Gradient surgery for multi-task learning.
\newblock In H.~Larochelle, M.~Ranzato, R.~Hadsell, M.F. Balcan, and H.~Lin
  (eds.), \emph{Advances in Neural Information Processing Systems}, volume~33,
  pp.\  5824--5836. Curran Associates, Inc., 2020.
\newblock URL
  \url{https://proceedings.neurips.cc/paper_files/paper/2020/file/3fe78a8acf5fda99de95303940a2420c-Paper.pdf}.

\bibitem[Zhang et~al.(2020)Zhang, Vuong, and Ross]{zhang20}
Yiming Zhang, Quan Vuong, and Keith Ross.
\newblock First order constrained optimization in policy space.
\newblock In H.~Larochelle, M.~Ranzato, R.~Hadsell, M.F. Balcan, and H.~Lin
  (eds.), \emph{Advances in Neural Information Processing Systems}, volume~33,
  pp.\  15338--15349. Curran Associates, Inc., 2020.
\newblock URL
  \url{https://proceedings.neurips.cc/paper/2020/file/af5d5ef24881f3c3049a7b9bfe74d58b-Paper.pdf}.

\bibitem[Zhou et~al.(2022)Zhou, Zhang, Zhao, Xiong, and Wei]{zhou_c_sac}
Xuanhan Zhou, Xiaochen Zhang, Haitao Zhao, Jun Xiong, and Jibo Wei.
\newblock Constrained soft actor-critic for energy-aware trajectory design in
  uav-aided iot networks.
\newblock \emph{IEEE Wireless Communications Letters}, 11\penalty0
  (7):\penalty0 1414--1418, 2022.
\newblock \doi{10.1109/LWC.2022.3172336}.

\end{thebibliography}
\bibliographystyle{iclr2024_conference}
\clearpage
\appendix

\section{Additional Empirical Results on the Impact of Competing Objectives}\label{comp_obj_app}

Here we supplement the empirical findings of Section \ref{competing}. First, we note that the duration of ``early'' training reflected by the x-axis was chosen to match the time before the first network reset in experiments that featured network resetting.

\begin{figure}[H]
\centering
\includegraphics[width=0.325\textwidth]{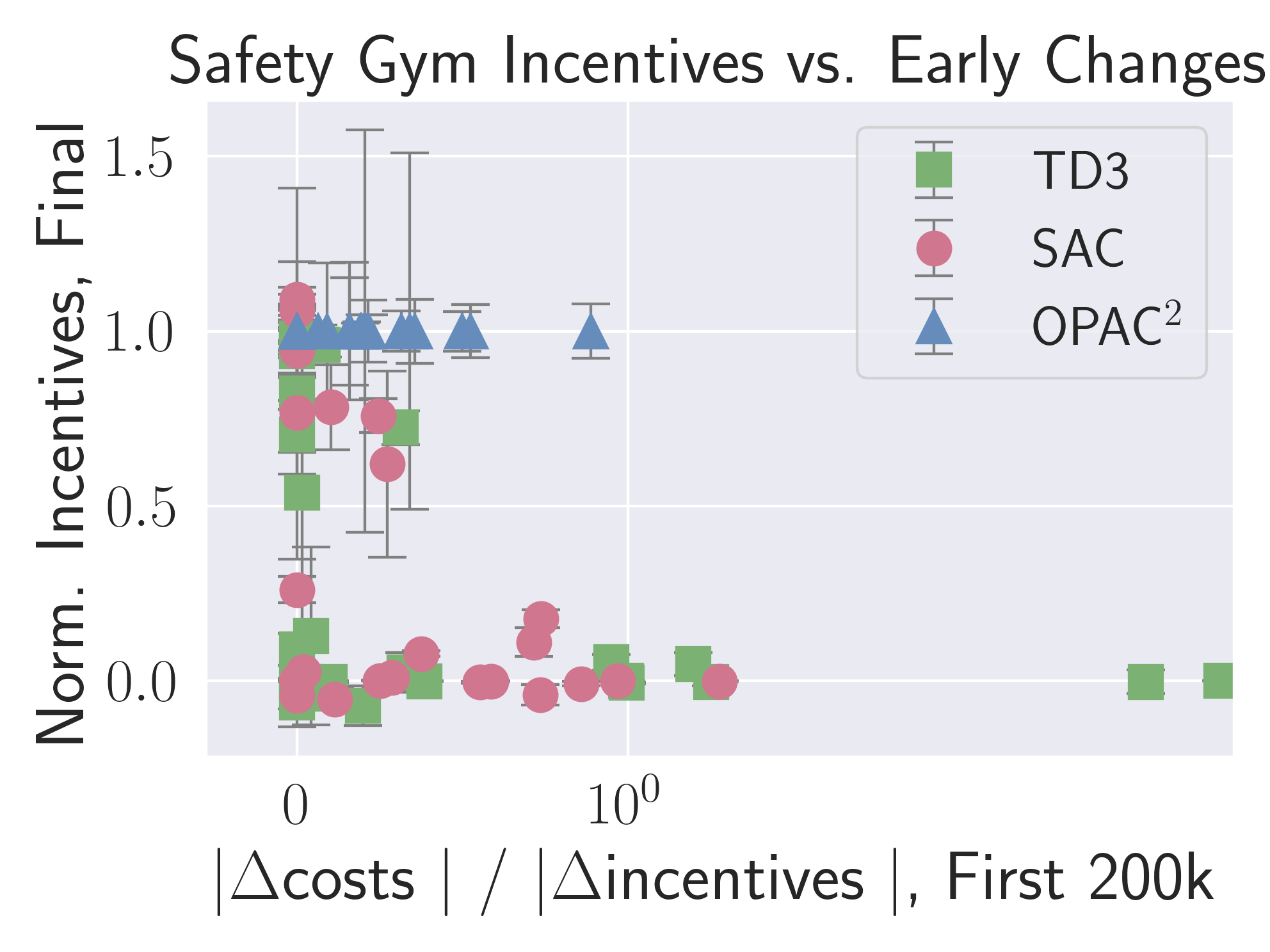}
\includegraphics[width=0.325\textwidth]{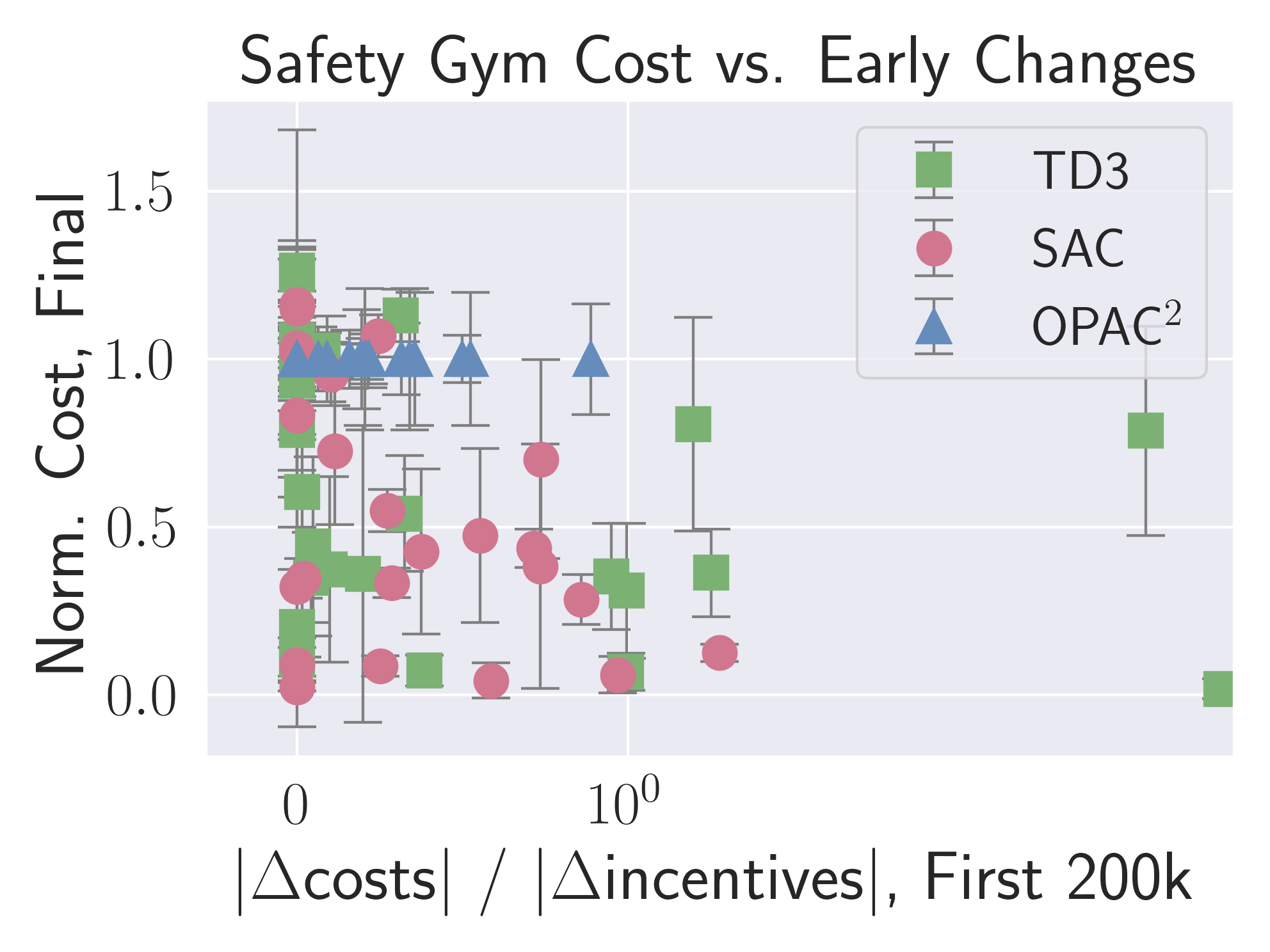} 
\caption{\textbf{Left panel}: Average positive reward (incentive) per episode at the end of training, as a function of initial prioritization of cost.  \textbf{Right panel}: Average cost per episode at the end of training, as a function of initial prioritization of cost.  In both cases, quantities are normalized by the OPAC$^2$ result for the given environmental configuration.}
 \label{fig:a11}
\end{figure}

In Figure \ref{fig:a11}, we show the breakdown of incentive and cost in the experiments that constitute Figure \ref{fig:200k}, again normalizing by the performance of OPAC$^2$ on each environment.  We find that, particularly with nonzero cost weights, OPAC$^2$ achieves far higher levels of incentives than TD3 or SAC (left panel).  OPAC$^2$ does accumulate higher costs than the others (right panel), but by not nearly as large a factor.  These trends correspond to OPAC$^2$ accumulating much higher \emph{total} reward (Figure \ref{fig:200k}), as is the goal of the optimization.  When cost is considered in these environments, the SAC and TD3 agents largely fail to properly balance the competing objectives.

\begin{figure}[H]
\centering
\includegraphics[width=0.325\textwidth]{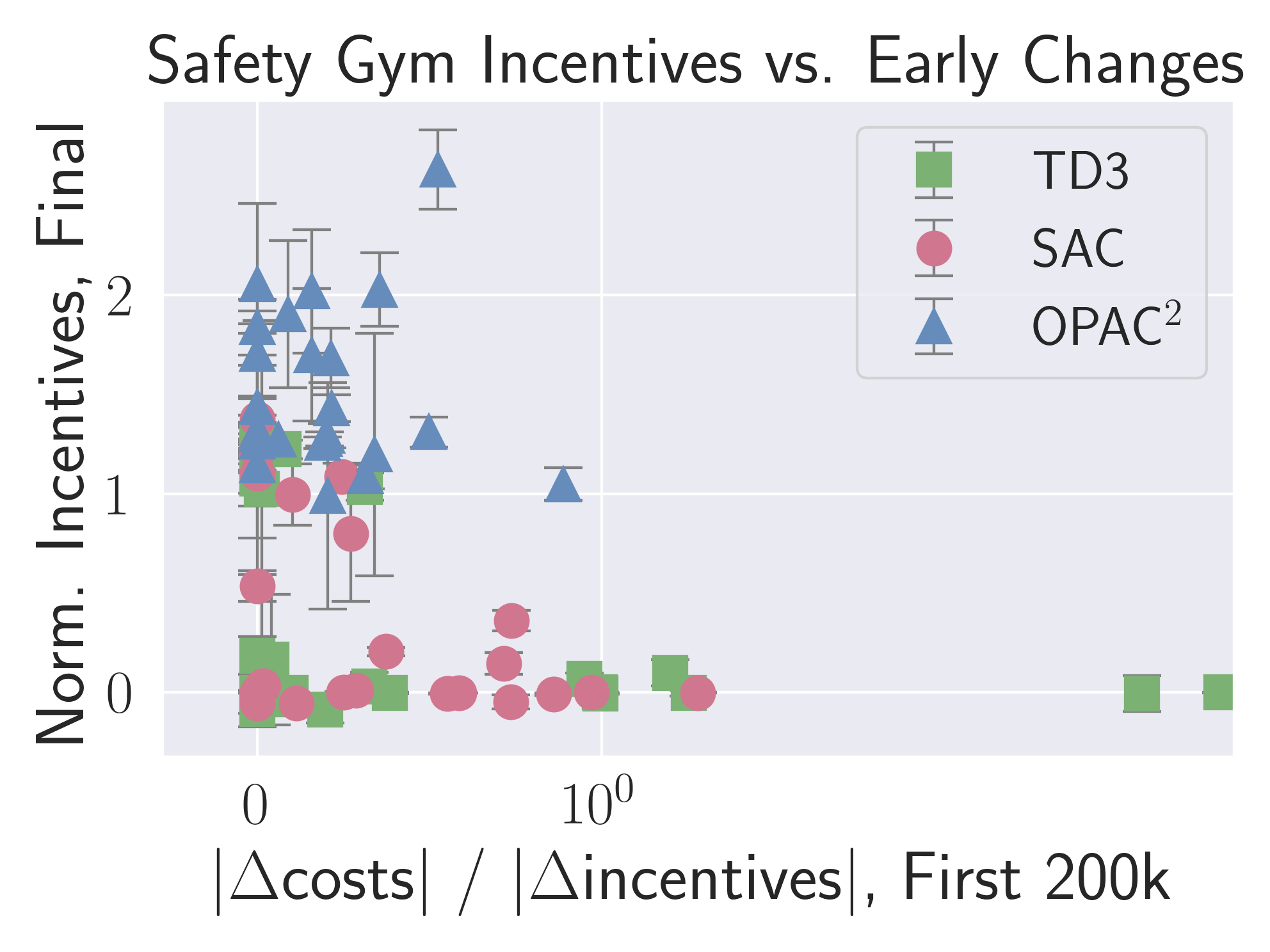}
\includegraphics[width=0.325\textwidth]{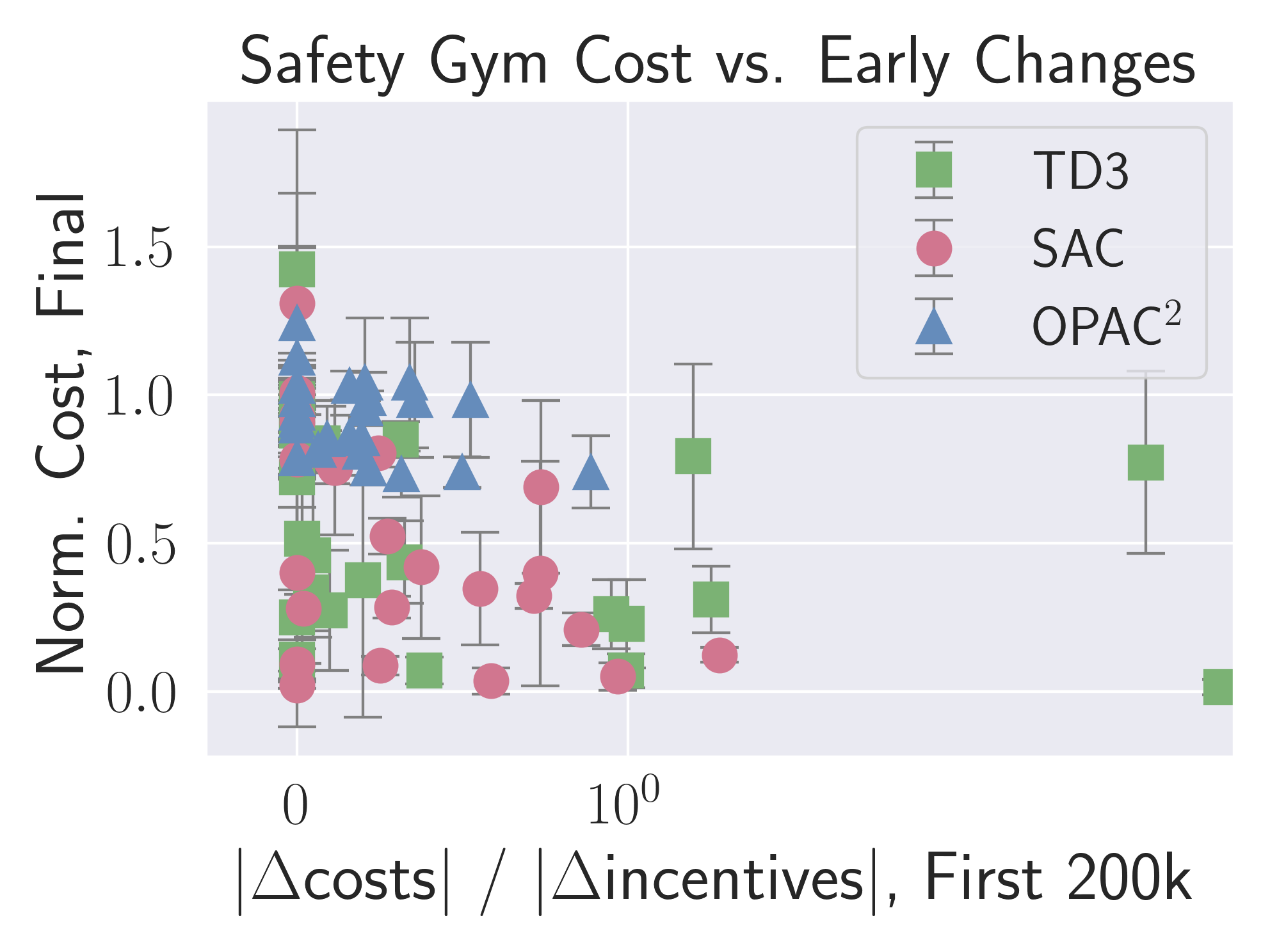}
\caption{\textbf{Left panel}: Average positive reward (incentive) per episode at the end of training, as a function of initial prioritization of cost.  \textbf{Right panel}: Average cost per episode at the end of training, as a function of initial prioritization of cost.  In both cases, quantities are normalized by the best on-policy result from \cite{ray2019safetygym} for the given environmental configuration.}
 \label{fig:a12}
\end{figure}

In Figure \ref{fig:a12}, we again show the incentives (left panel) and costs (right panel) at the end of training plotted against the ratio of early change in cost accumulation to early change in incentives gained, but this time normalized by the best on-policy result for an agent unaware of cost \cite{ray2019safetygym}.  We see that OPAC$^2$ vastly outperforms the on-policy methods in terms of positive reward accumulation, despite using half as much training data.  As OPAC$^2$ is increasingly configured to consider cost, its accumulation of both cost and reward decreases.

\section{Unconstrained Off-Policy Actor-Critic, SQUashed and REgularizeD} \label{app:unc_alg}
Below is pseudocode for the unconstrained version of our method, OPAC$^2$.  It represents a simplification of the constrained approach provided in the main text.

\begin{algorithm}[H]
\caption{Off-Policy Actor-Critic, SQUAshed and REgularizeD (OPAC$^2$) \label{alg:opac2}}
\begin{algorithmic}[1]
\State \textbf{Input:} Initial policy parameters $\theta$; $Q$ parameters $\phi$; $V$ parameters $\psi$
\State \textbf{Input:} Initial entropy weight $\alpha$
\State Set $V$ target equal to main parameters: $\psi_{\mathrm{targ}} \leftarrow \psi$
\State Initialize optimizers with learning rates $\lambda_\pi=\lambda_V=\lambda_Q; \lambda_\alpha$
\State Initialize replay buffer $\mathcal{D} = \emptyset$,  a ring buffer of fixed size
\State Initialize environment : $\bs \sim p(\bs_0)$ 
\For{iteration $k \in [0, \dots, K-1]$}
\For{step $s \in [0, \dots, S-1]$}  \Comment{Typically take just one step ($S=1$)}
\State Sample $\ba \sim \pi_{\theta_k}(\ba|\bs)$; observe $\bs' \sim p(\bs' | \bs, \ba)$  \Comment{One step in MDP}
\State $\mathcal{D} \leftarrow \mathcal{D} \cup \{(\bs,\ba,\bs', r(\bs,\ba))\}$  \Comment{Update buffer}
\EndFor
\For{gradient step $g \in [0, \dots, G-1]$}
\State Sample batch $B = \{(\bs_i, \ba_i, \bs'_i, r_i, d_i) \}$ from $\mathcal{D}$ \Comment{or, a batch for each parameter set}
\State Compute $Q$ error: $\mathcal{E}_Q(B) = \sum_i \left[Q(\bs_i, \ba_i) - (r_i + \gamma(1-d_i) V_{\mathrm{targ}}(\bs_i))\right]^2$
\State Update $Q$: $\phi \leftarrow \phi - \lambda_\phi \nabla_{\phi} \mathcal{E}_Q(B)$
\State Sample $\ba_{i,\pi} \sim \pi(\ba|\bs_i)$, using $\tanh$ squashing
\State Compute $V$ error: $\mathcal{E}_V(B) = \sum_i \left[(V(\bs_i )- Q(\bs_i, \ba_{i, \pi})\right]^2$
\State Update $V$: $\psi \leftarrow \psi - \lambda_\psi \nabla_{\psi} \mathcal{E}_V(B)$
\State Compute $A(\bs_i, \ba_{i,\pi}) = Q(\bs_i, \ba_{i,\pi}) - V(\bs_i)$ \Comment{Normalize}
\State Sample $\ba_{\mathrm{rp},\pi} \sim \pi(\ba|\bs_i)$ with reparameterization trick, for $\tanh$ squashing
\State Compute $\pi$ loss: $\mathcal{E}_\pi(B) = \sum_i\left[\alpha\log \pi_\theta(\ba_{\mathrm{rp},\pi} | \bs_i) - A(\bs_i, \ba_{i,\pi})\log(\pi_\theta(\ba_{i,\pi} | \bs_i) \right] $
\State Update $\pi$: $\theta \leftarrow \theta - \lambda_\theta \nabla_\theta \mathcal{E}_\pi(B)$
\State Compute $\alpha$ loss: $\mathcal{E}_\alpha(B)=-\alpha\left[\log(\pi(\ba_{i,\pi}|\bs_i)) +\mathcal{H}_{\mathrm{target}}\right]$
\State Update $\alpha$ : $\alpha \leftarrow \alpha - \lambda_\alpha\nabla_\alpha \mathcal{E}_\alpha(B)$
\State Update value target: $\psi_{\mathrm{targ}} \leftarrow \rho \psi_{\mathrm{targ}} + (1-\rho)\psi$
\EndFor

\EndFor
\end{algorithmic}
\end{algorithm}

\section{Experimental Details} \label{app:exp}

Full source code is available in the supplementary material, and will be released publicly pending review. Hyperparameters are listed in Table \ref{tab:hyper_shared}. For all Safety Gym experiments, we used a learning rate of $10^{-4}$ for all algorithms (matching the setting in the OpenAI \texttt{safety-starter-agents} accompanying the suite). For all DeepMind Control experiments, we used a learning rate of $3\times 10^{-4}$ for SAC and TD3, following standard practice.  We retained the learning rate of $10^{-4}$ for OPAC$^2$ on DeepMind Control.  All methods shared the same learning rate for the logarithm of the entropy weight $\alpha$ ($5\times 10^-4$) and all constrained experiments used a learning rate of $5\times 10^-6$ was used for the Lagrangian $\beta$. For all DeepMind Control experiments, we configured OPAC$^2$ to use max-entropy regularization, finding that it performed better than the entropy bonus we used for experiments with mixed-sign rewards. 

Throughout, all neural networks considered were multilayer perceptrons, with two hidden layers of 256 units each.  As is standard, ReLU activations were used for SAC and TD3.  We chose $\tanh$ activations for OPAC$^2$ in order to match on-policy methods with a similar policy update.  We evaluated $\tanh$ activations on SAC and TD3 as well, but found them to make little difference.
For experiments involving resetting, we followed the practice of \cite{nikishin2022primacy} of resetting all networks and optimizers (except for those corresponding to the learned temperature) every 200k environment steps. All traces shown reflect five random seeds. 

The policy networks output the mean values of a multivariate normal distribution with diagonal covariance. For OPAC$^2$, control variances were optimized.  Variances were independent of state for all constrained experiments with the Car and Point robots, as well as for unconstrained experiments for Car and Point on Goal and Push.  They varied with state for all experiments with Doggo robot, as well as for the Car and Point robots on the unconstrained Button task. For SAC, control variances always varied with state.

\begin{table}[H]
    \centering
 \begin{tabular}{lc}
        \toprule
        Parameter & Value \\
        \midrule
    Discount  & 0.99  \\
    Replay Buffer Size  & $10^6$  \\
    Optimizer & Adam \citep{KiBa17} \\
    Network Layers & 2\\
    Network Hidden Units (per layer) & 256 \\
    Batch Size & 256 \\
    Target Network Update Interval & 1\\
    $\tau$ (target network averaging)  & 0.995 \\
    Initial Exploration  & 10000  \\
    \bottomrule
    \end{tabular}
    \caption{Shared hyperparameters for SAC, TD3, and OPAC$^2$}
    \label{tab:hyper_shared}
\end{table}

As mentioned in Section \ref{sec:experiments} of the main text, we chose to evaluate our approach using the OpenAI Safety Gym \citep{ray2019safetygym}. This choice was governed by our desire to test in conditions with clear cost-incentive trade-offs, significant stochasticity, adequate complexity, and available benchmarks. 

The environments chosen were the most obstacle-rich of the publicly available environments; we considered all robots and all tasks.  The Point robot is constrained to the 2D plane and has two control dimensions: one for moving forward/backward and one for turning. The Car robot also has two control dimensions, corresponding to independently actuated parallel wheels. It has a freely rotating wheel and, while it is not constrained to the 2D plane, typically remains in it.  The Doggo robot is a quadrupedal robot with bilateral symmetry and 12 control dimensions. Several types of obstacles and tasks were present in the environments we evaluated.  In all cases, the robot is given a fixed amount of time (1000 steps) to complete the prescribed task as many times as possible and is motivated by both sparse and dense reward contributions.  In the ``Goal'' environments, the robot must navigate to a series of randomly-assigned goal positions, with a new target being assigned as soon as a goal is reached.  In the ``Button'' environments, the robot must reach and press a sequence of goal buttons while avoiding other buttons.  In the ``Push'' task, the robot must push a box to a series of goal positions.  The set of obstacles are different for each task; among the three environments there are a total of five different constraint elements (hazards, vases, incorrect buttons, pillars, and gremlins), each with different dynamics.  See \cite{ray2019safetygym} for further details.

All of our experiments used a single indicator for overall cost at each time step (the OpenAI default).  In the unconstrained experiments, each cost event (robot contacting obstacle) was assigned a fixed (negative) weight in the reward function.  The Car and Point robots used large penalty weights that matched \cite{aaai_paper}; the small ones were reduced by a factor of 2.  The DoggoButton2 environment used factors of 0.0125 and 0.00625 for large and small penalty weights, respectively, while DoggoGoal2 used 0.025 and 0.0125.

\section{Additional Unconstrained Results: Safety Gym}\label{unc_app}
Below we supplement the unconstrained results for Safety Gym provided in the main text.  We include all learning curves, as well as representative plots of TD error and Q approximation error, both on held-out validation transitions.  Five episodes (5000 transitions) of evaluation data were collected every 10000 training steps throughout the learning process.

\begin{figure}[H]
    \centering
    \includegraphics[width=0.325\textwidth]{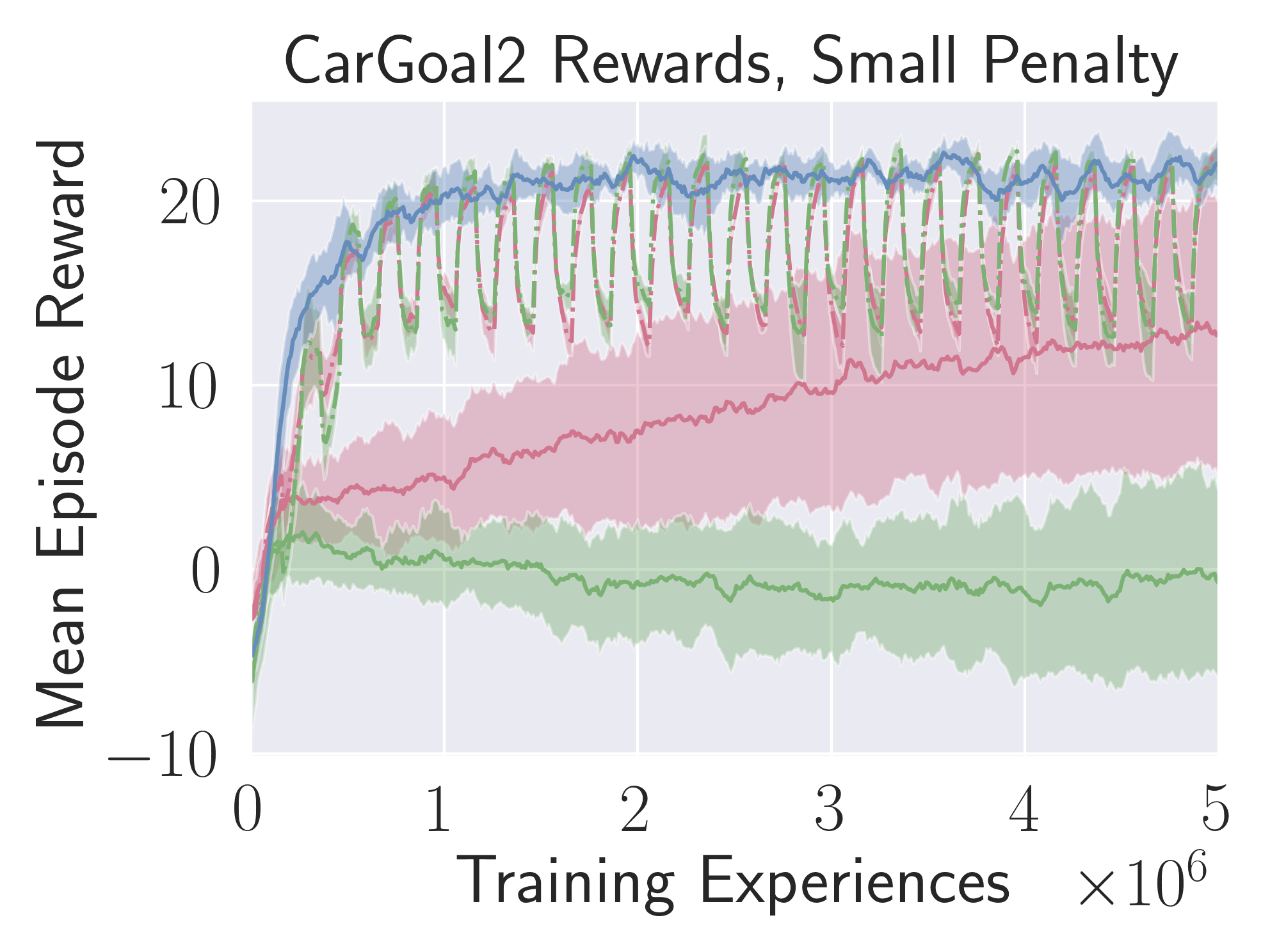}
    \includegraphics[width=0.325\textwidth]{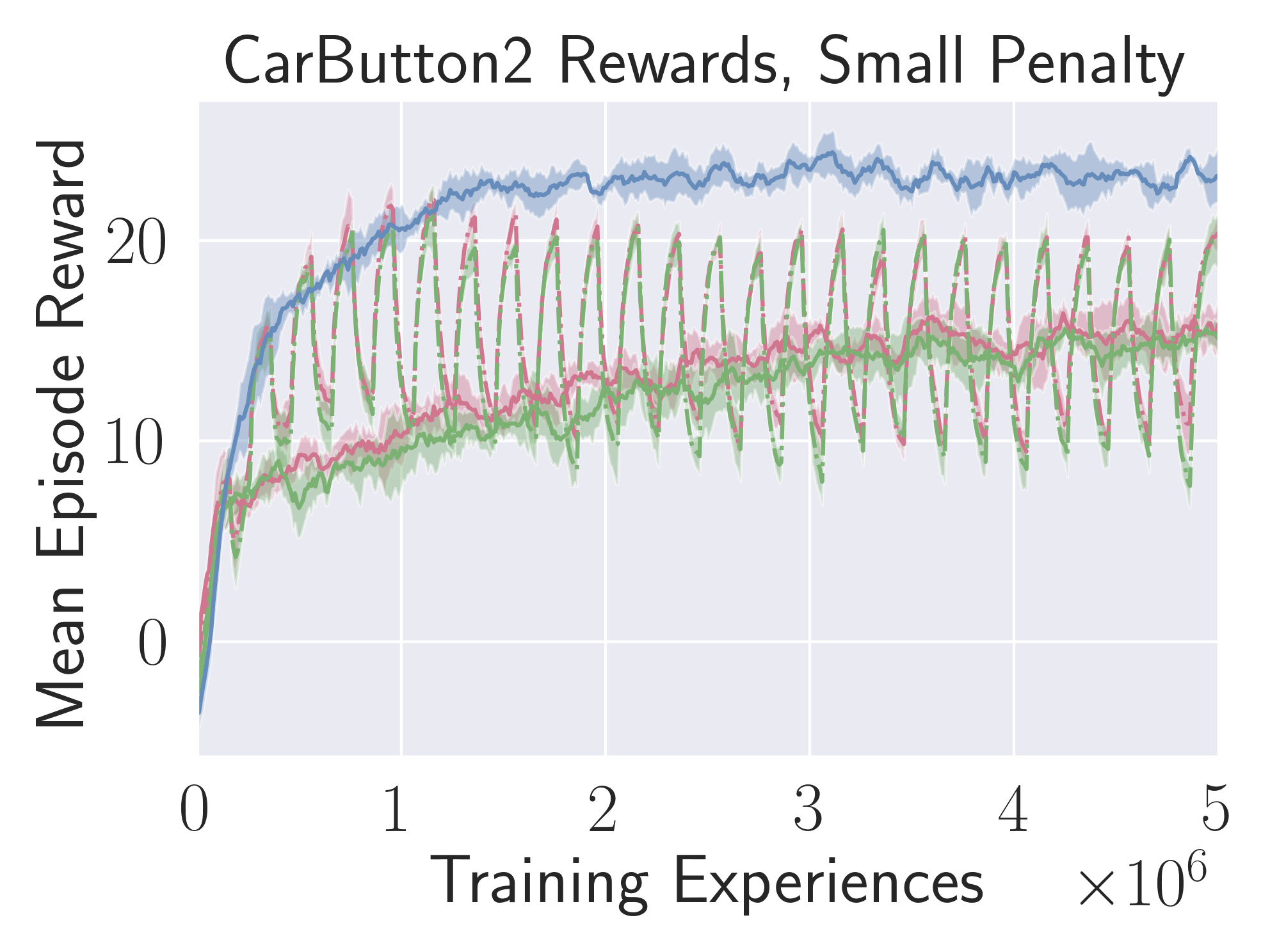}
    \includegraphics[width=0.325\textwidth]{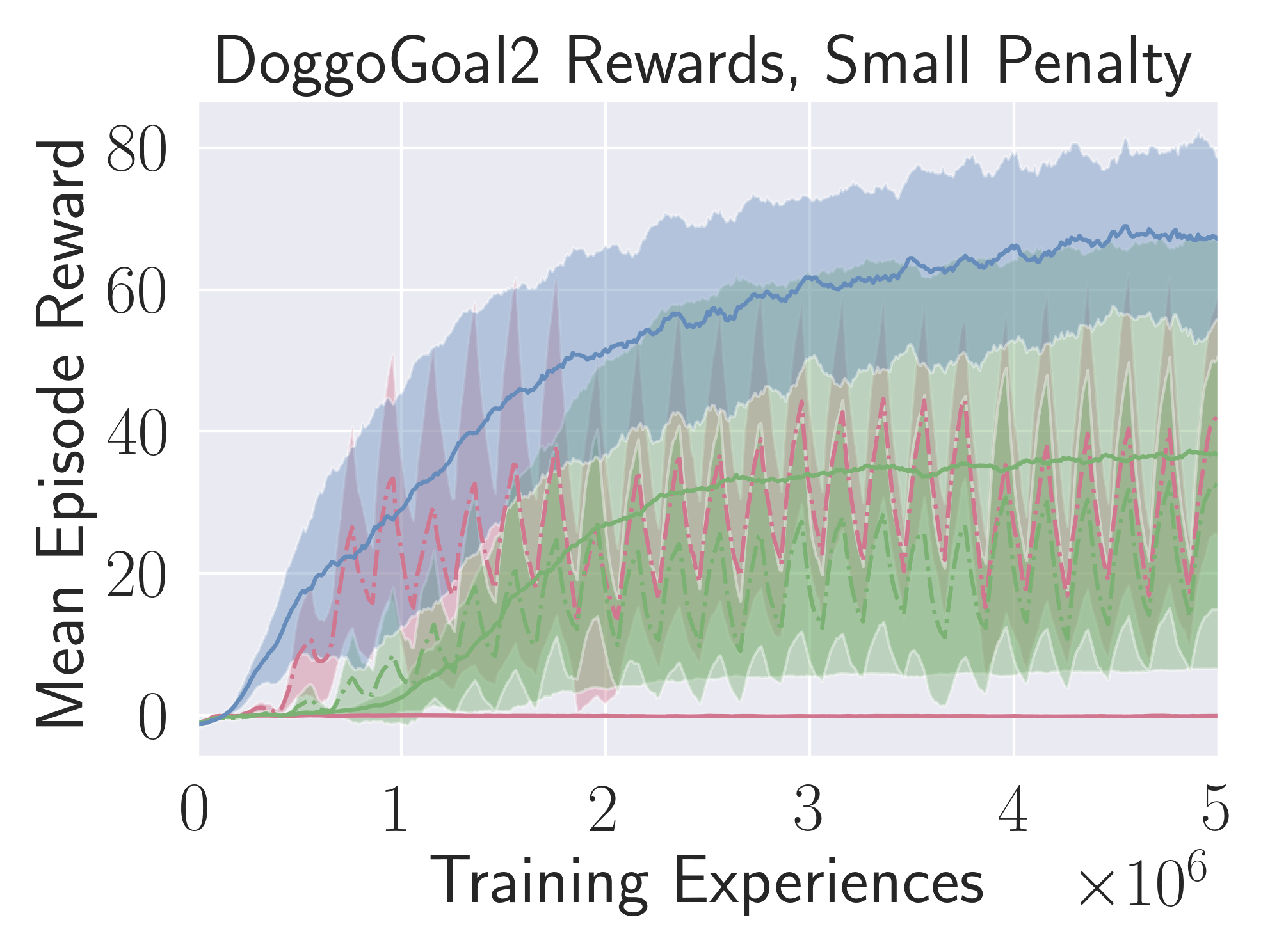}
    \includegraphics[width=0.325\textwidth]{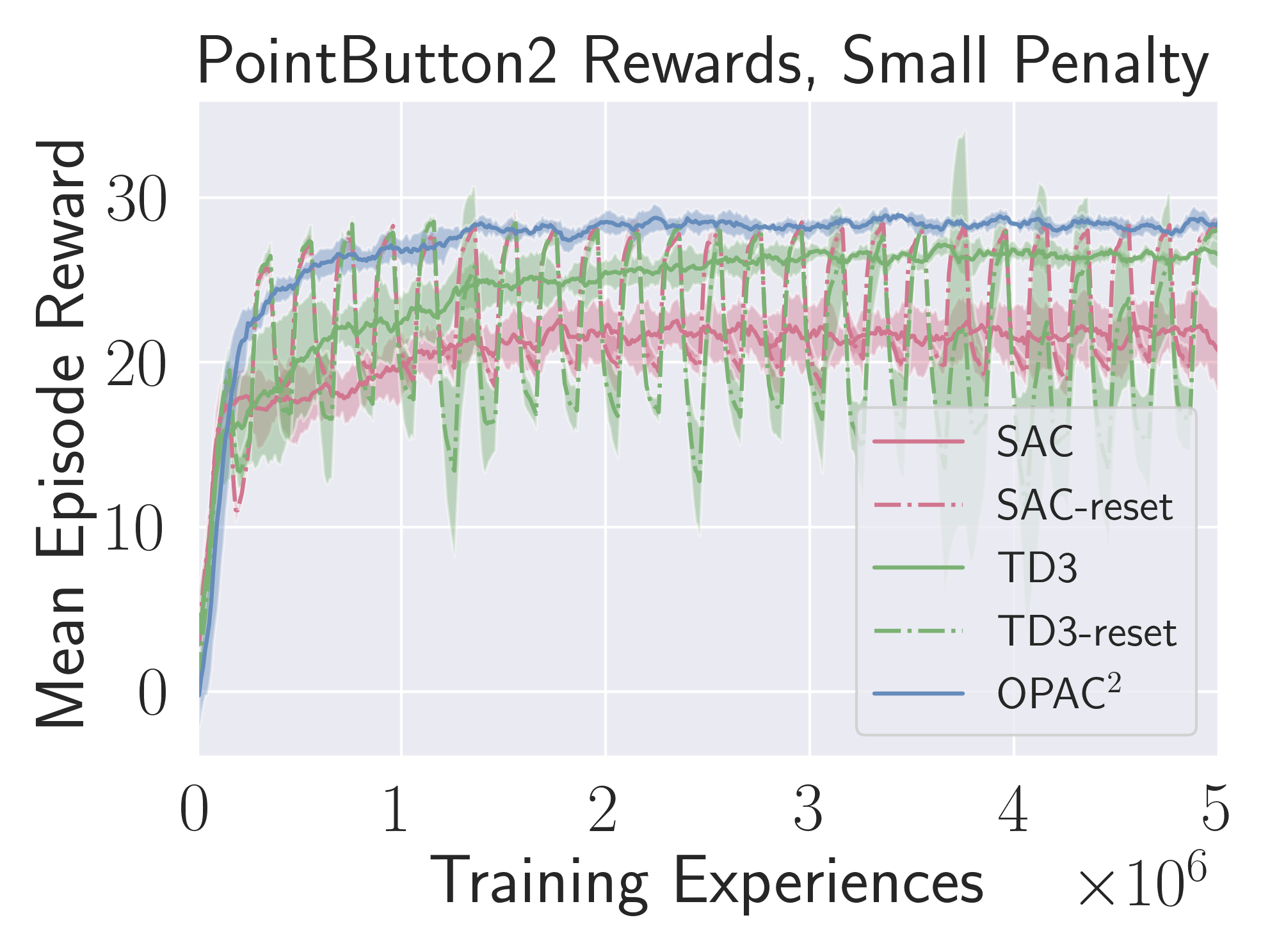}
    \includegraphics[width=0.325\textwidth]{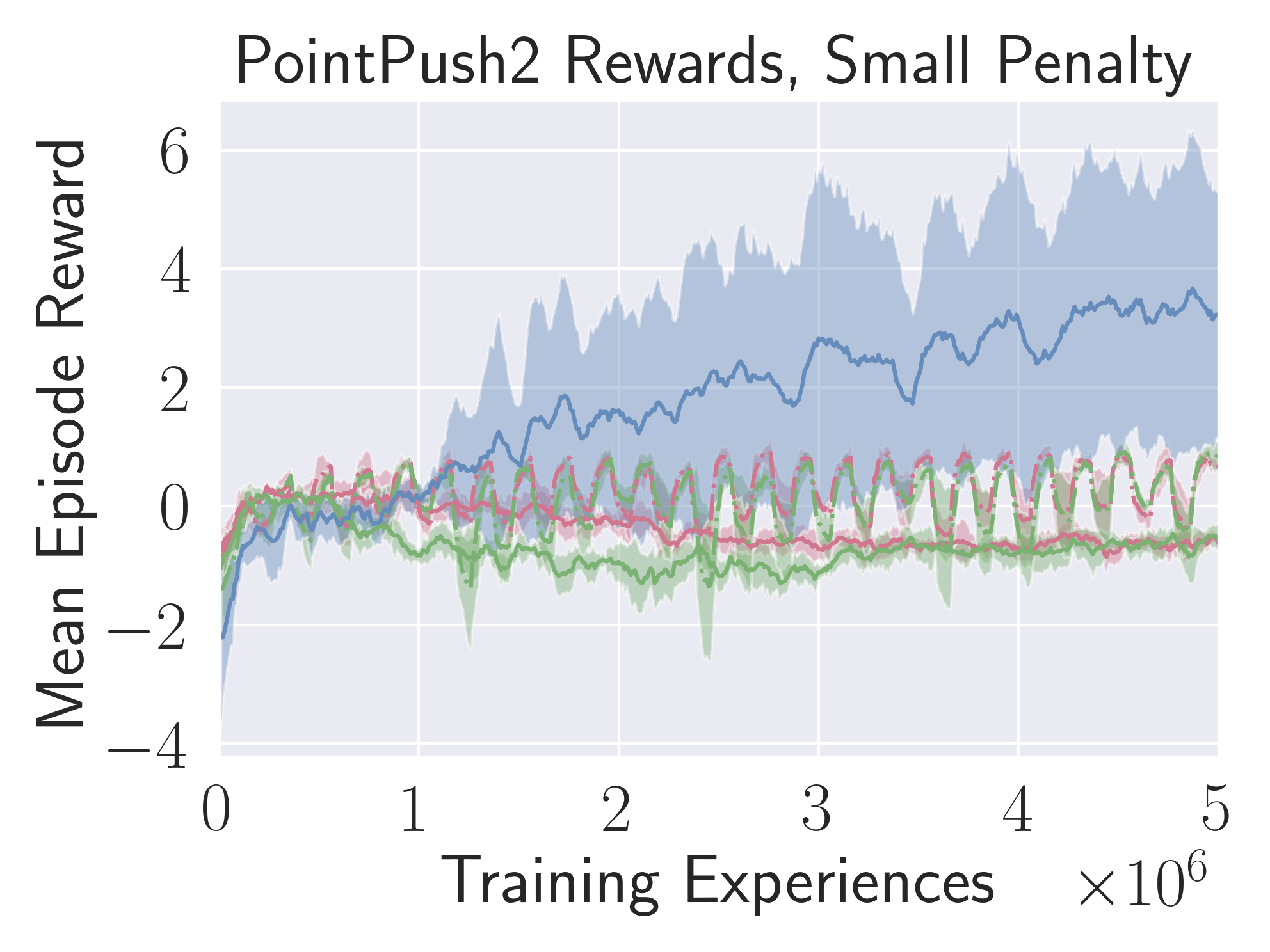}\\
    \includegraphics[width=0.9\textwidth]{figures/legend_1.png}
    \includegraphics[width=0.325\textwidth]{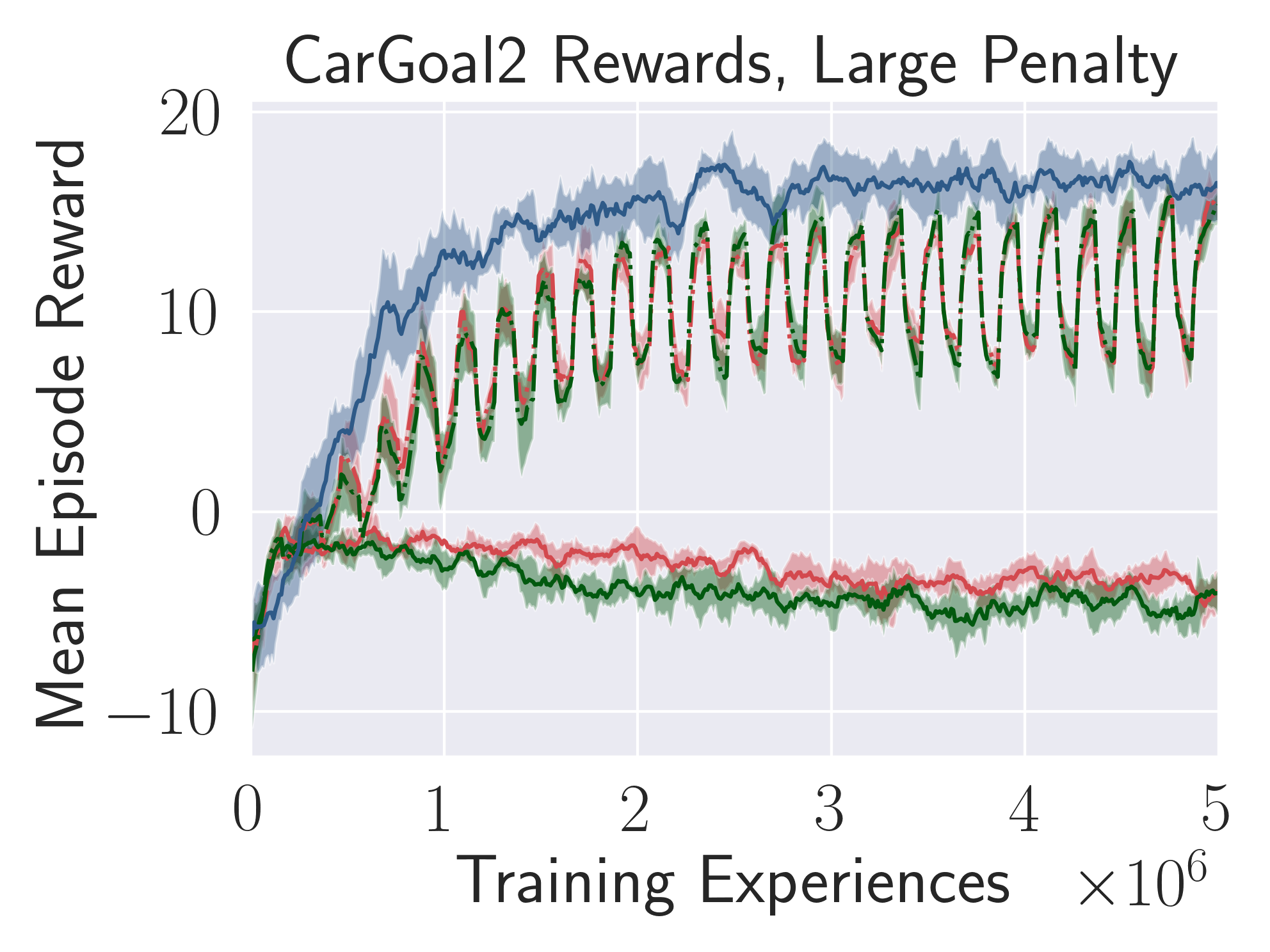}
    \includegraphics[width=0.325\textwidth]{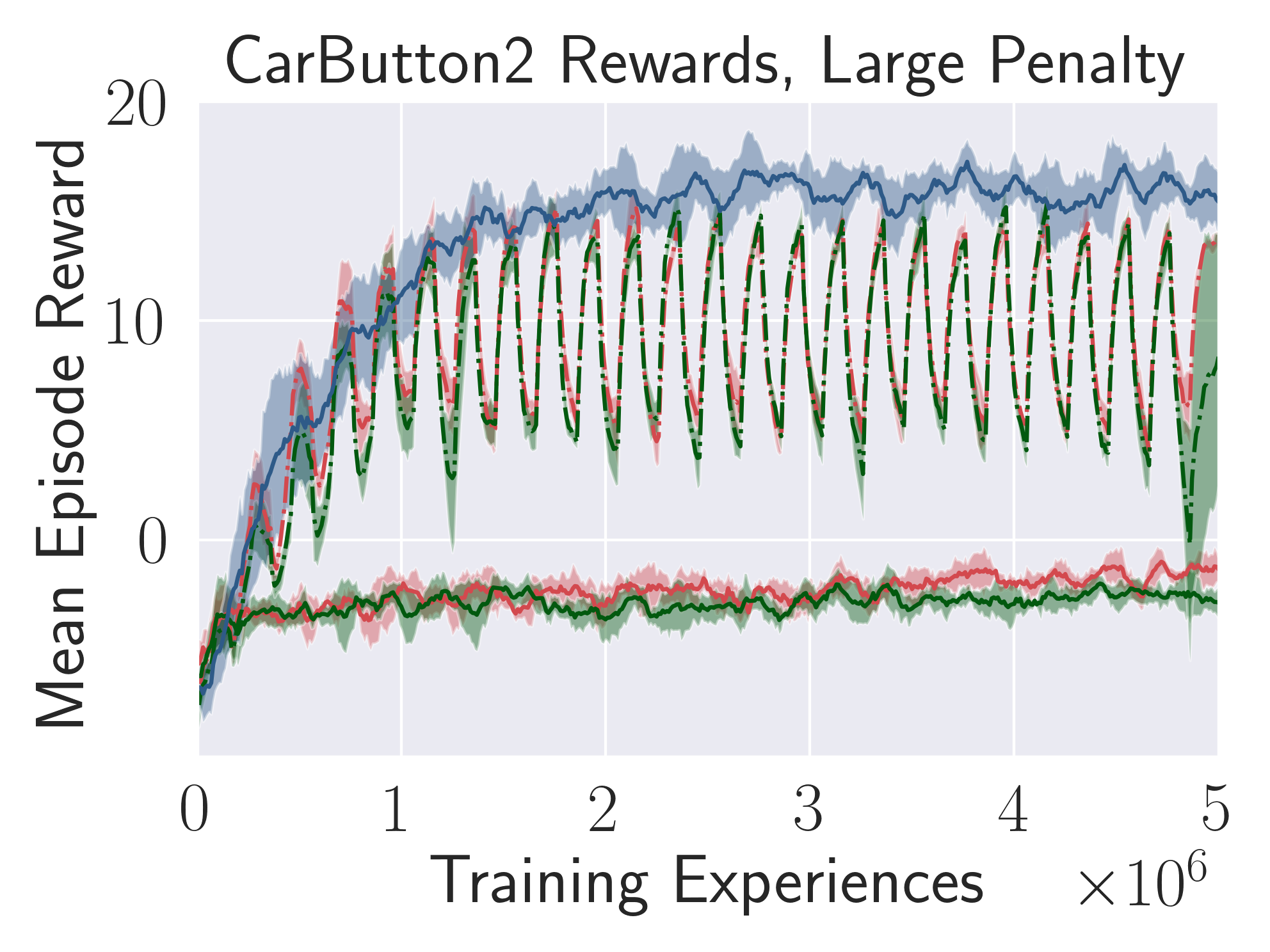}
    \includegraphics[width=0.325\textwidth]{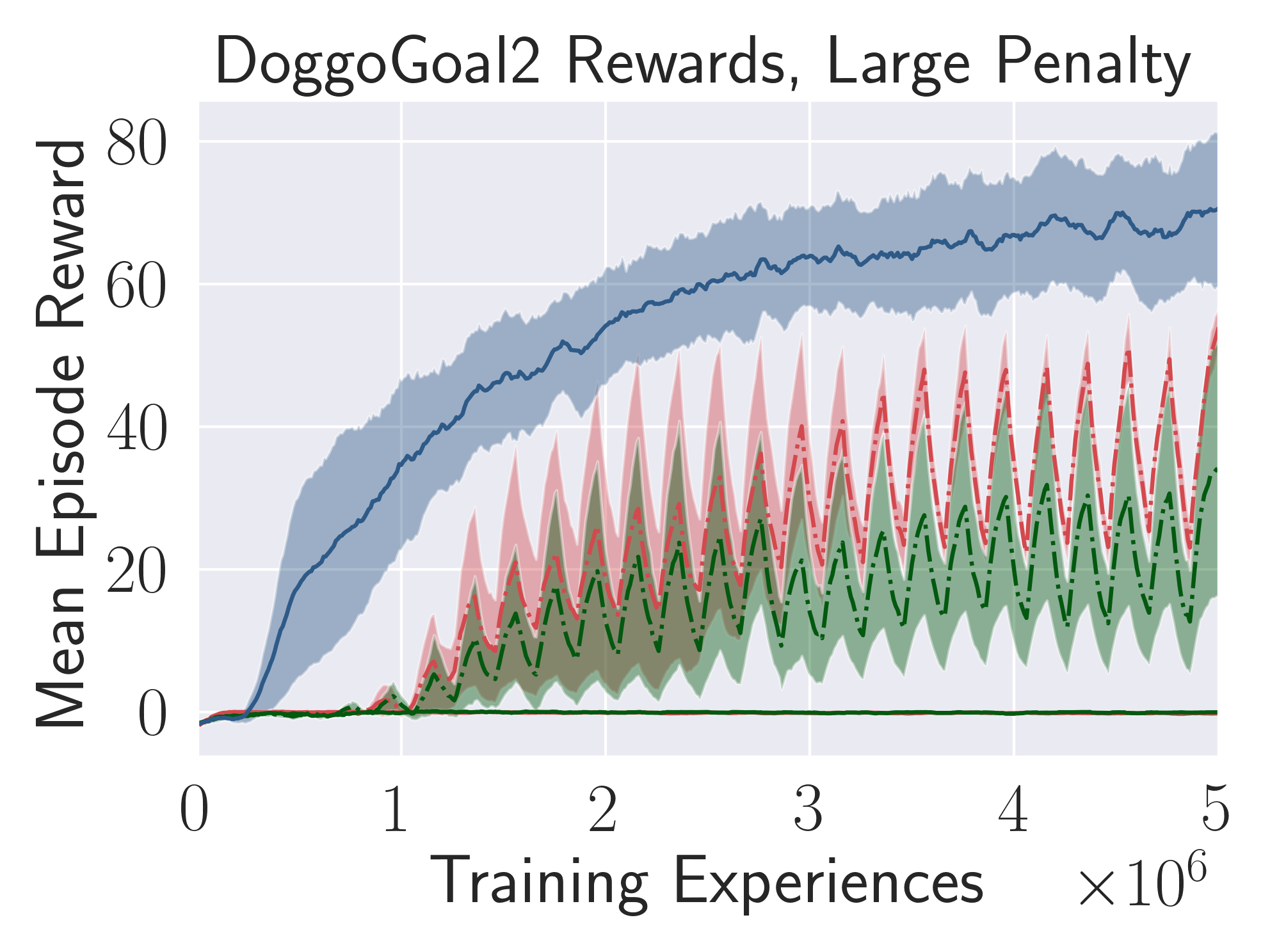}
    \includegraphics[width=0.325\textwidth]{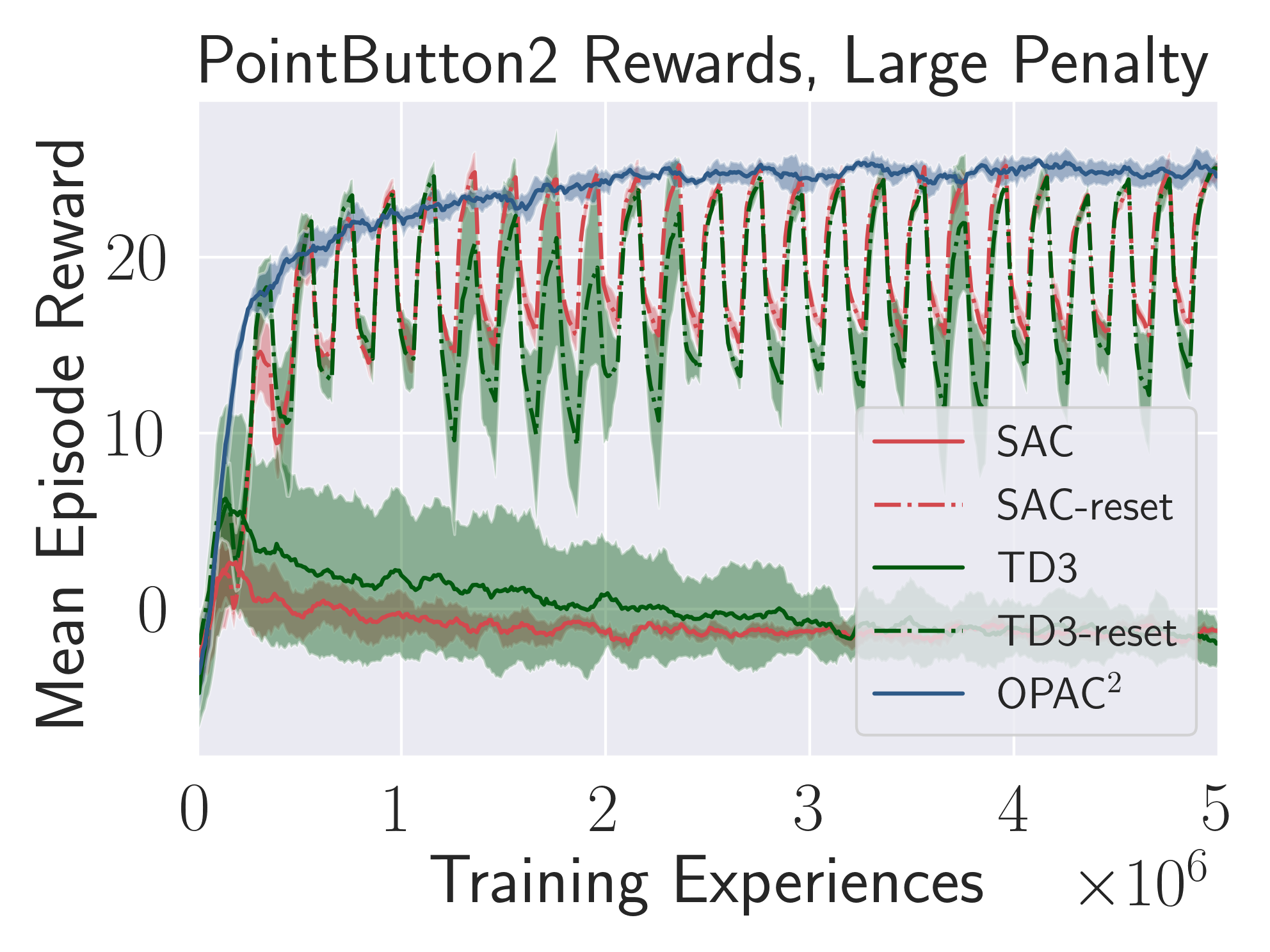}
    \includegraphics[width=0.325\textwidth]{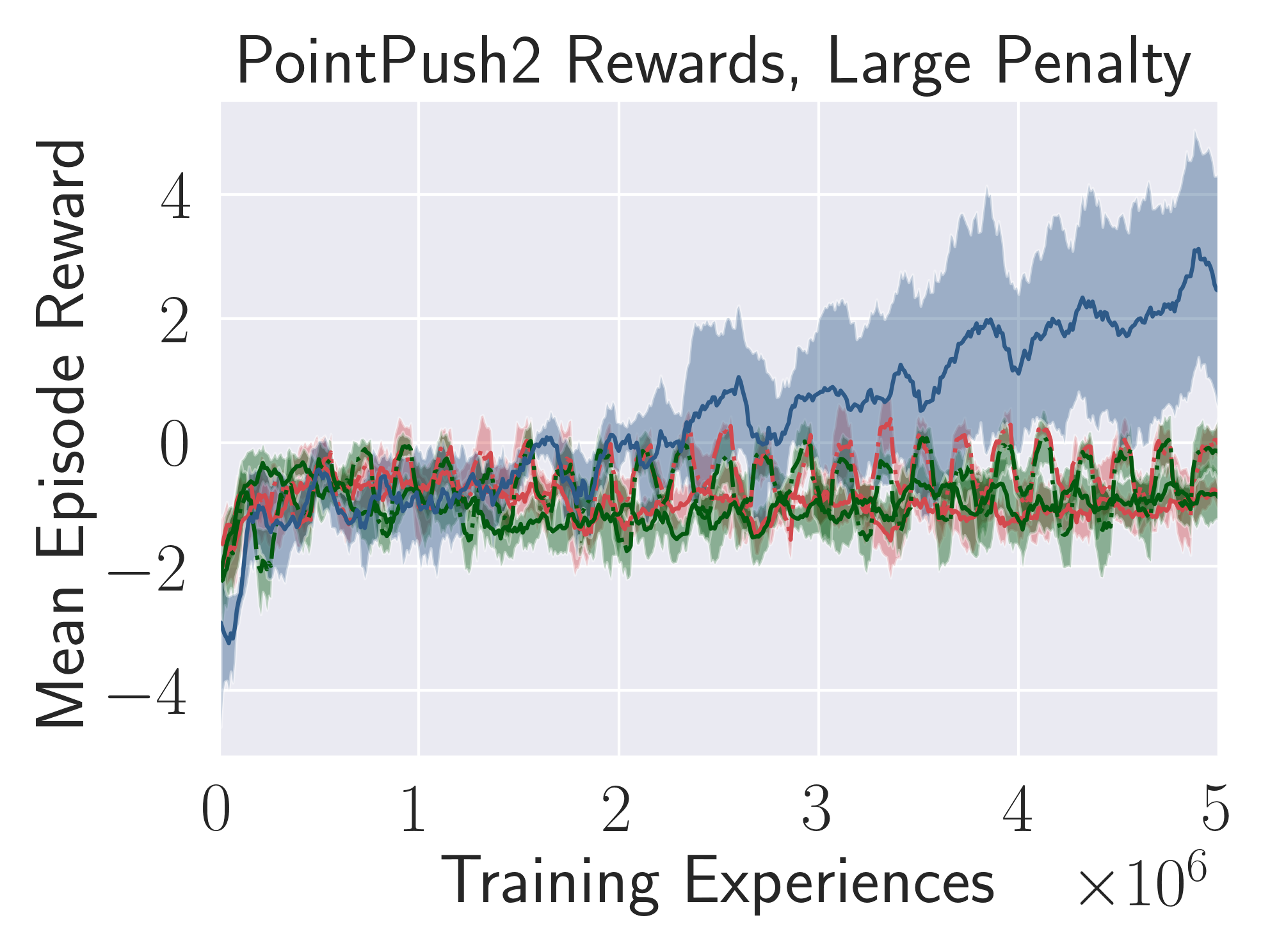}
    \includegraphics[width=0.9\textwidth]{figures/legend_2.png}
    \caption{Unconstrained Learning on Safety Gym.  SAC and TD3 struggle to learn without resetting. Even with periodic resets, they do not match the performance of OPAC$^2$.}
    \label{fig:rem_unc_rew}
\end{figure}

As mentioned in the main text, we observed negative $Q$ errors in all Point and Car environments, while the trend was more variable with Doggo.  While the magnitude of TD and Q errors varied per-environment, qualitative trends were consistent across environments.

\begin{figure}[H]
\centering
\includegraphics[width=0.325\textwidth]{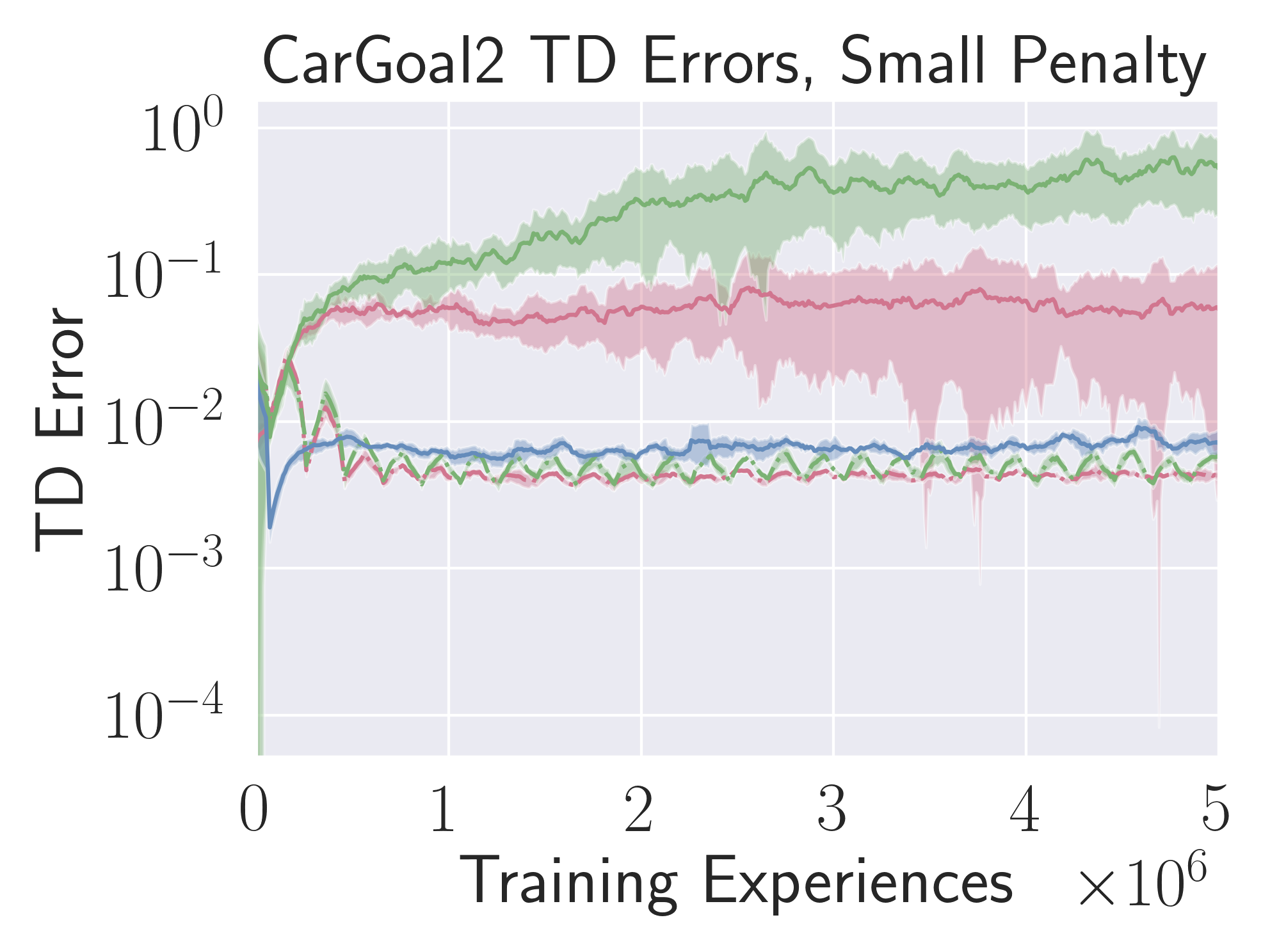}
\includegraphics[width=0.325\textwidth]{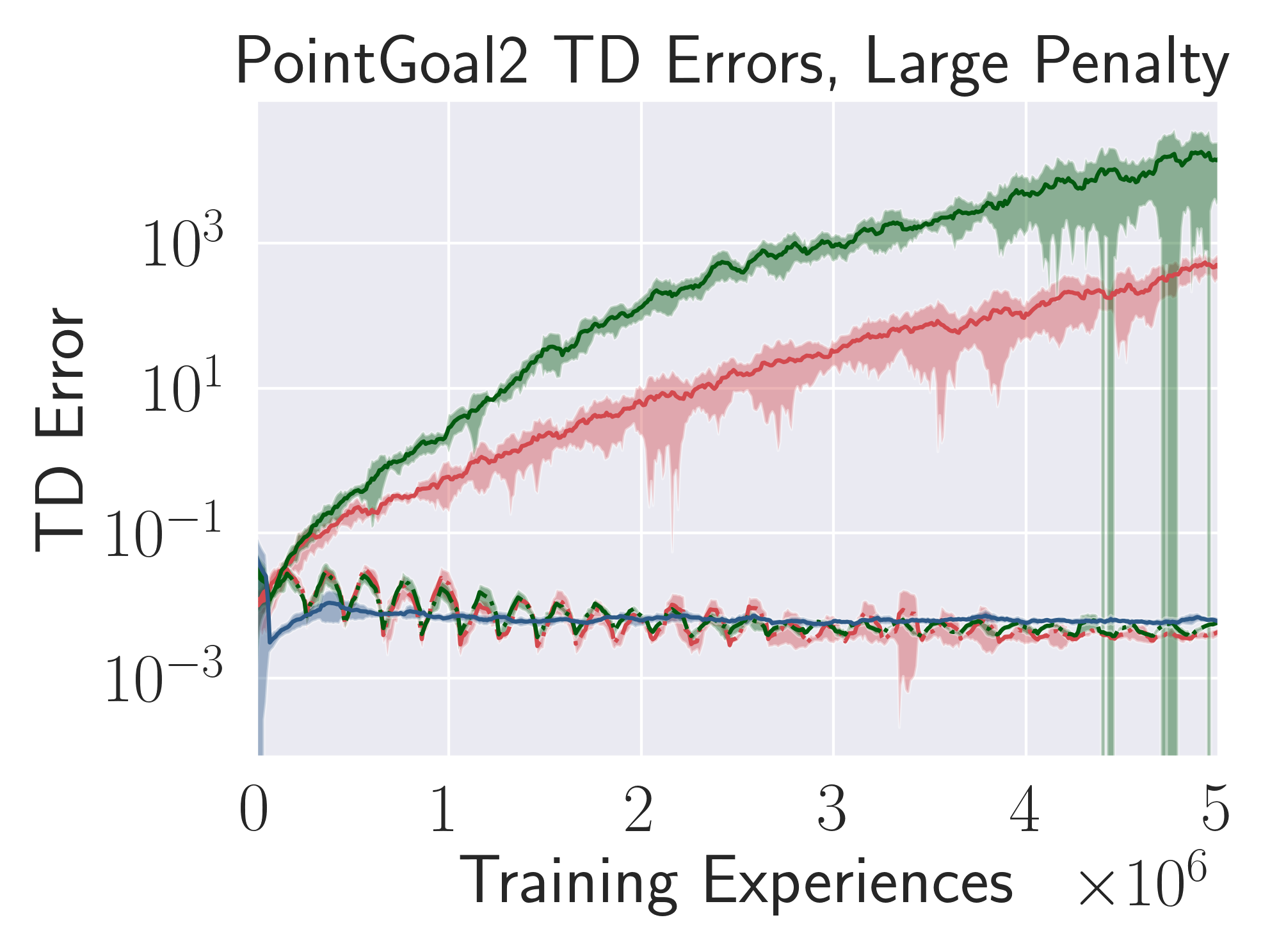}
\includegraphics[width=0.325\textwidth]{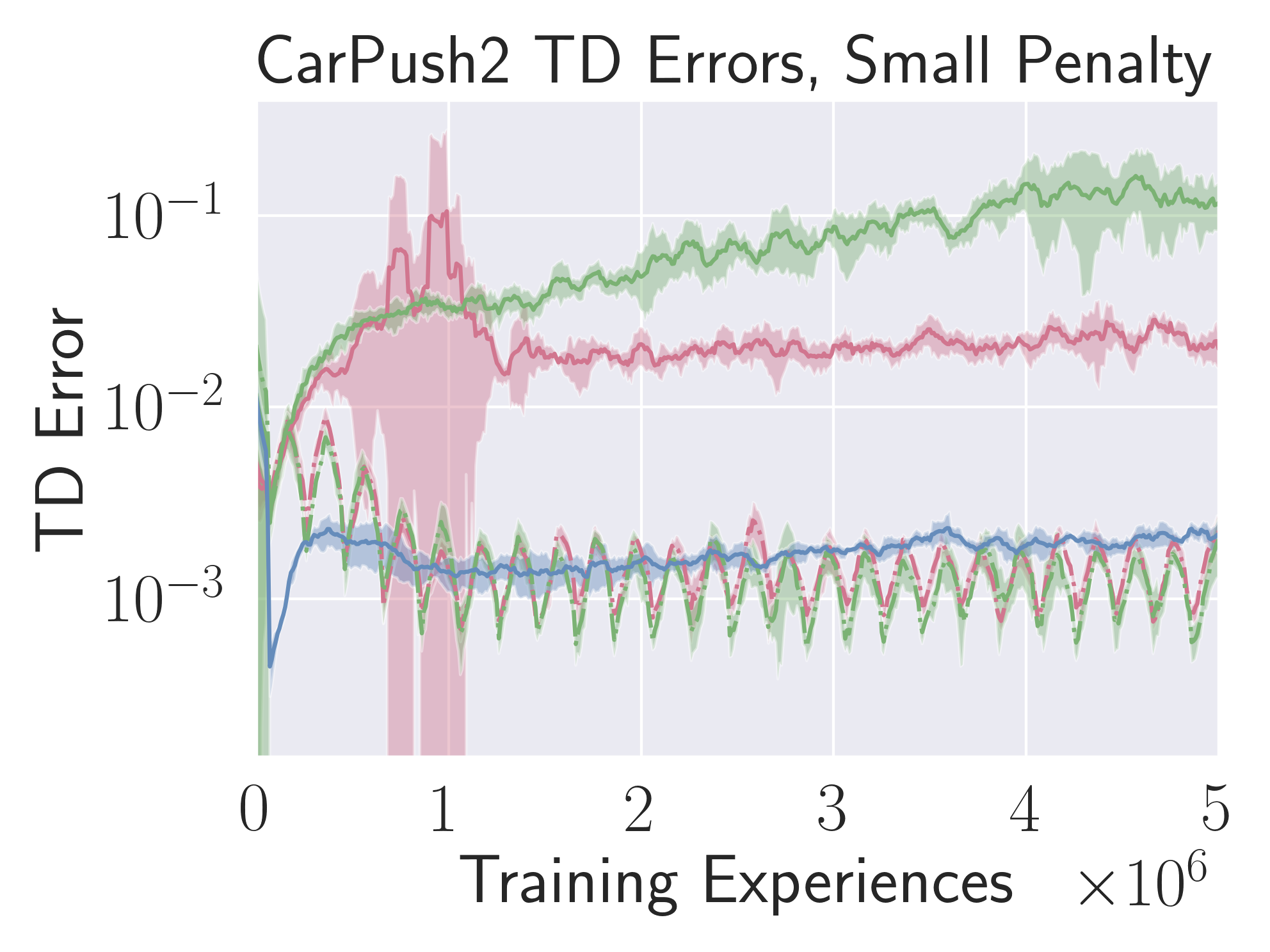}
\includegraphics[width=0.325\textwidth]{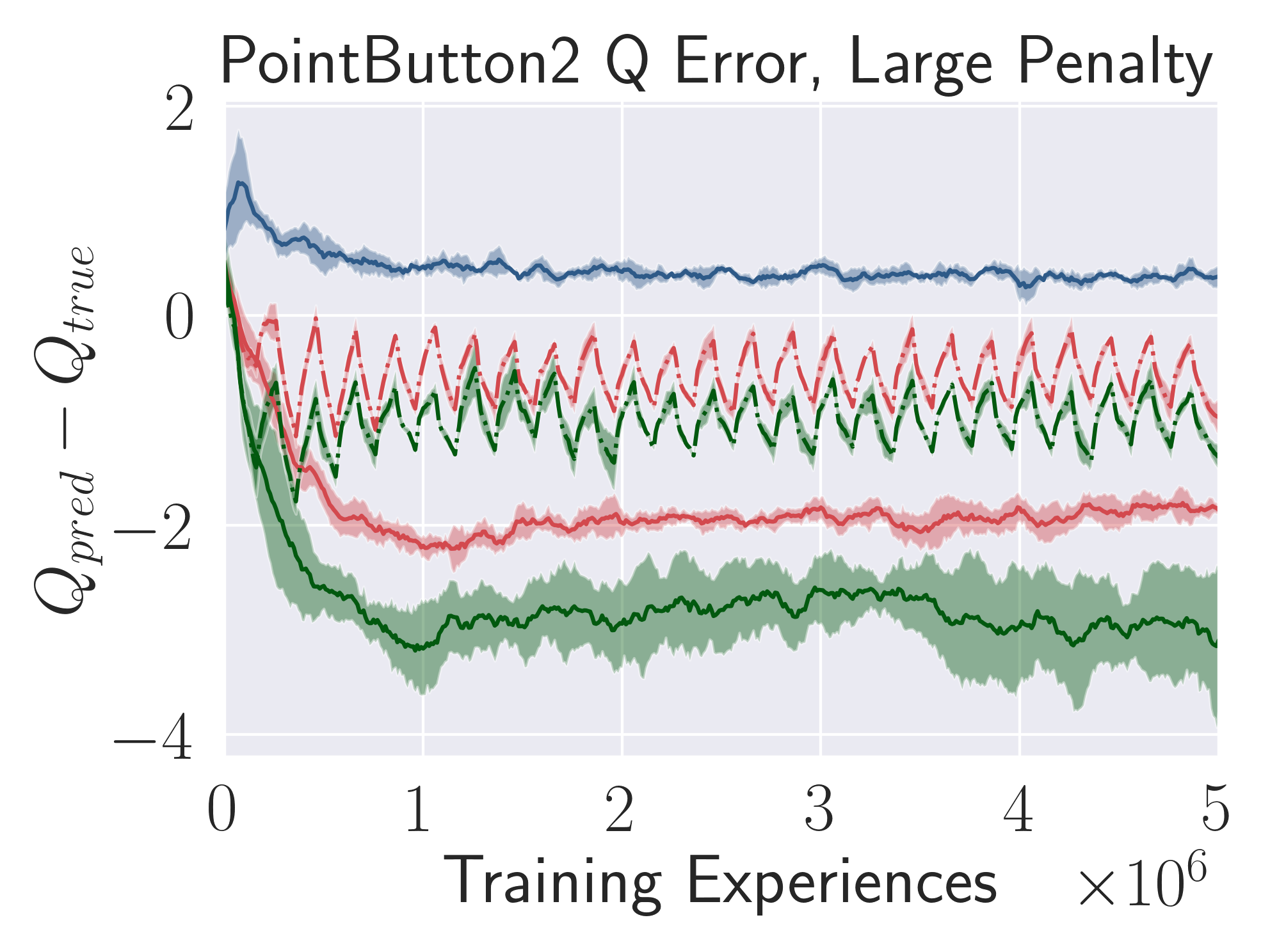}
\includegraphics[width=0.325\textwidth]{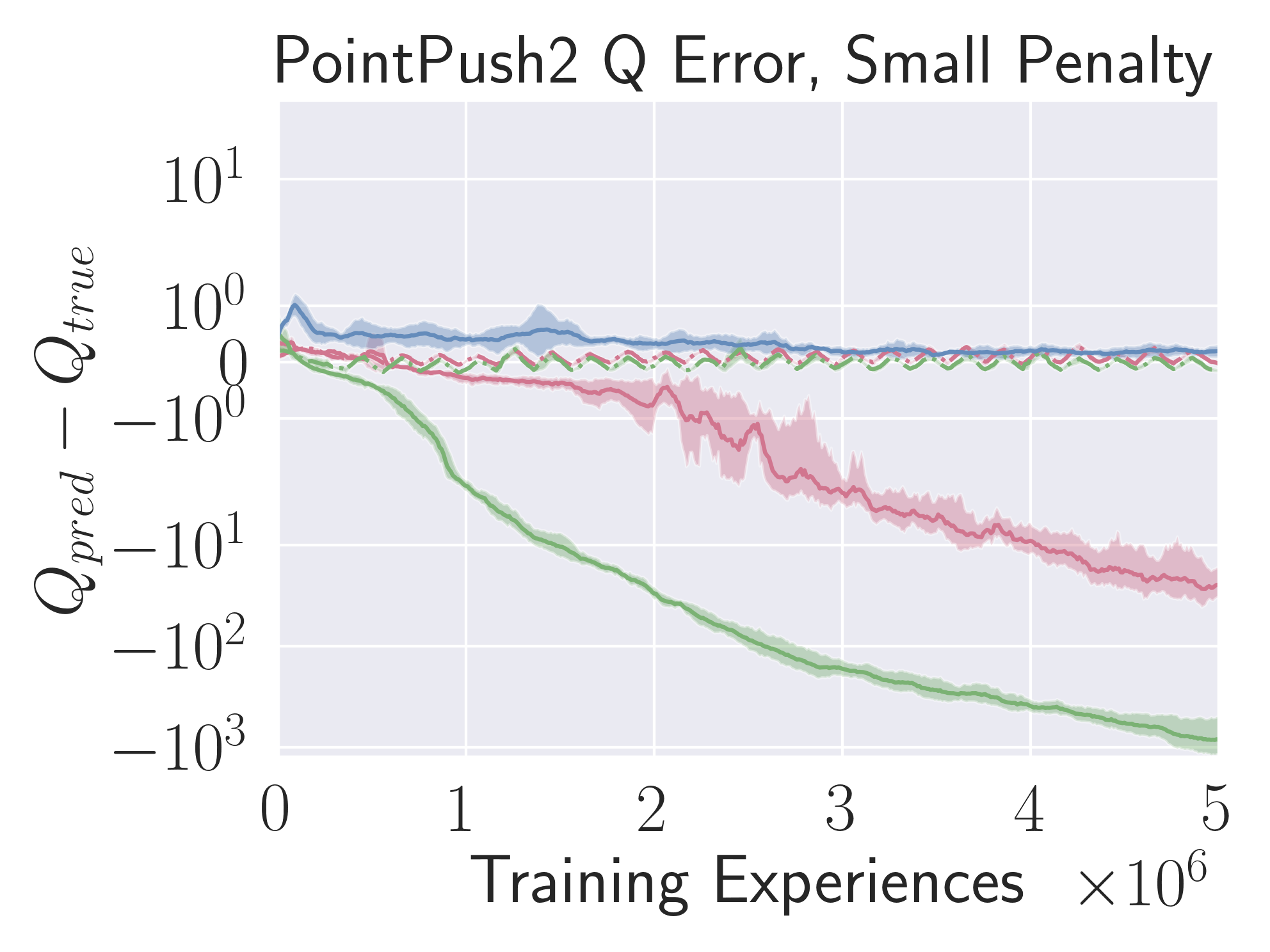}
\includegraphics[width=0.325\textwidth]{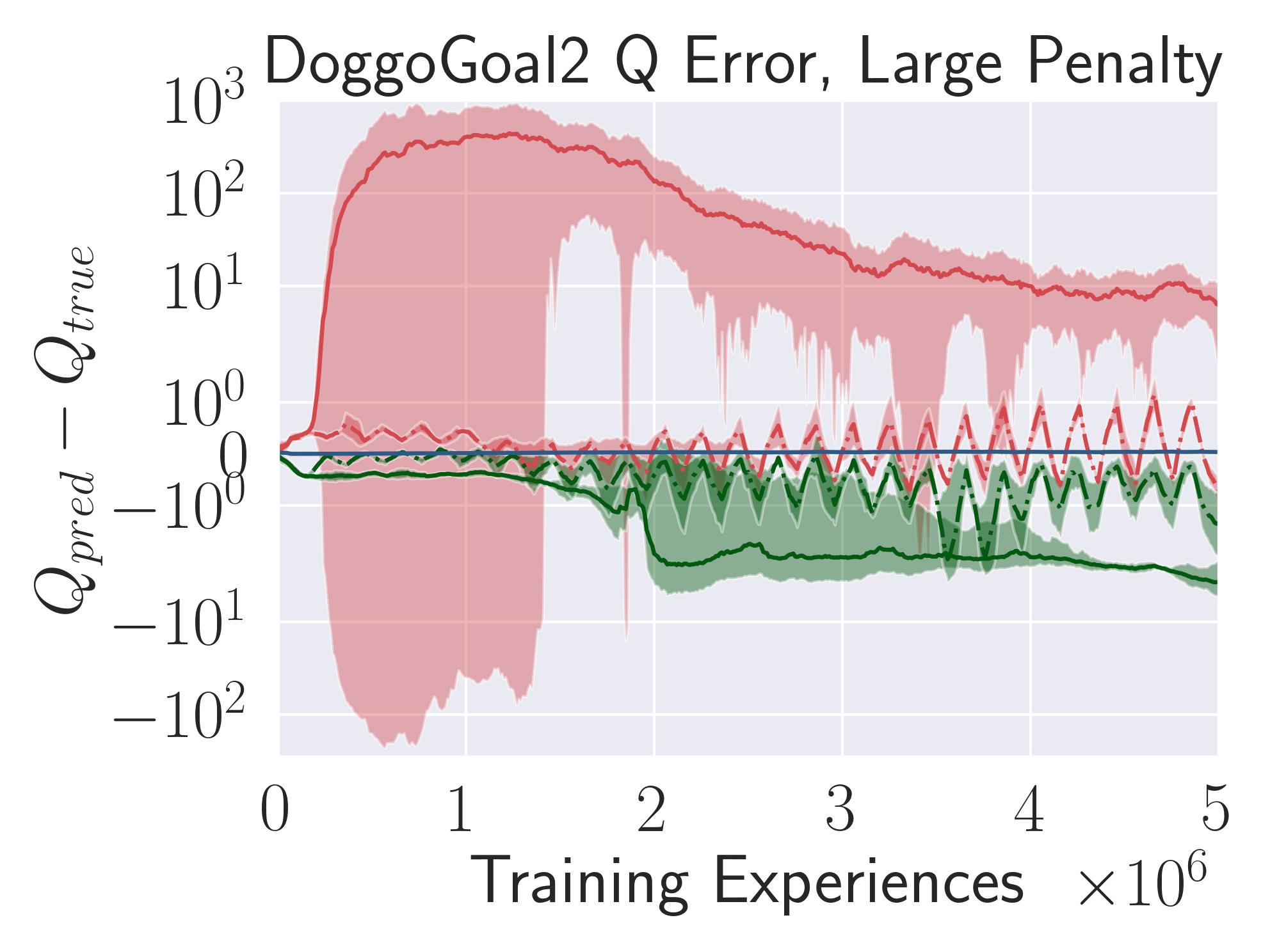}
\includegraphics[width=0.9\textwidth]{figures/legend_1.png}
\includegraphics[width=0.9\textwidth]{figures/legend_2.png}
\caption{Validation TD error, Q estimation error plots for sample environments.}
\label{fig:extra_diagnostics}
\end{figure}

\subsection{Resetting with the Off-Policy Actor-Critic}\label{reset_app}
We additionally explored the effect of resetting on OPAC$^2$.  We found resetting to not be beneficial, and in fact typically be detrimental, to the learning of OPAC$^2$ in Safety Gym. This makes sense, considering that OPAC$^2$ has fairly accurate value estimates to begin with. Any small gains in value estimation accuracy provided by resetting are not sufficient to overcome its disruptive effects on network training.

\begin{figure}[H]
    \centering
    \includegraphics[width=0.325\textwidth]{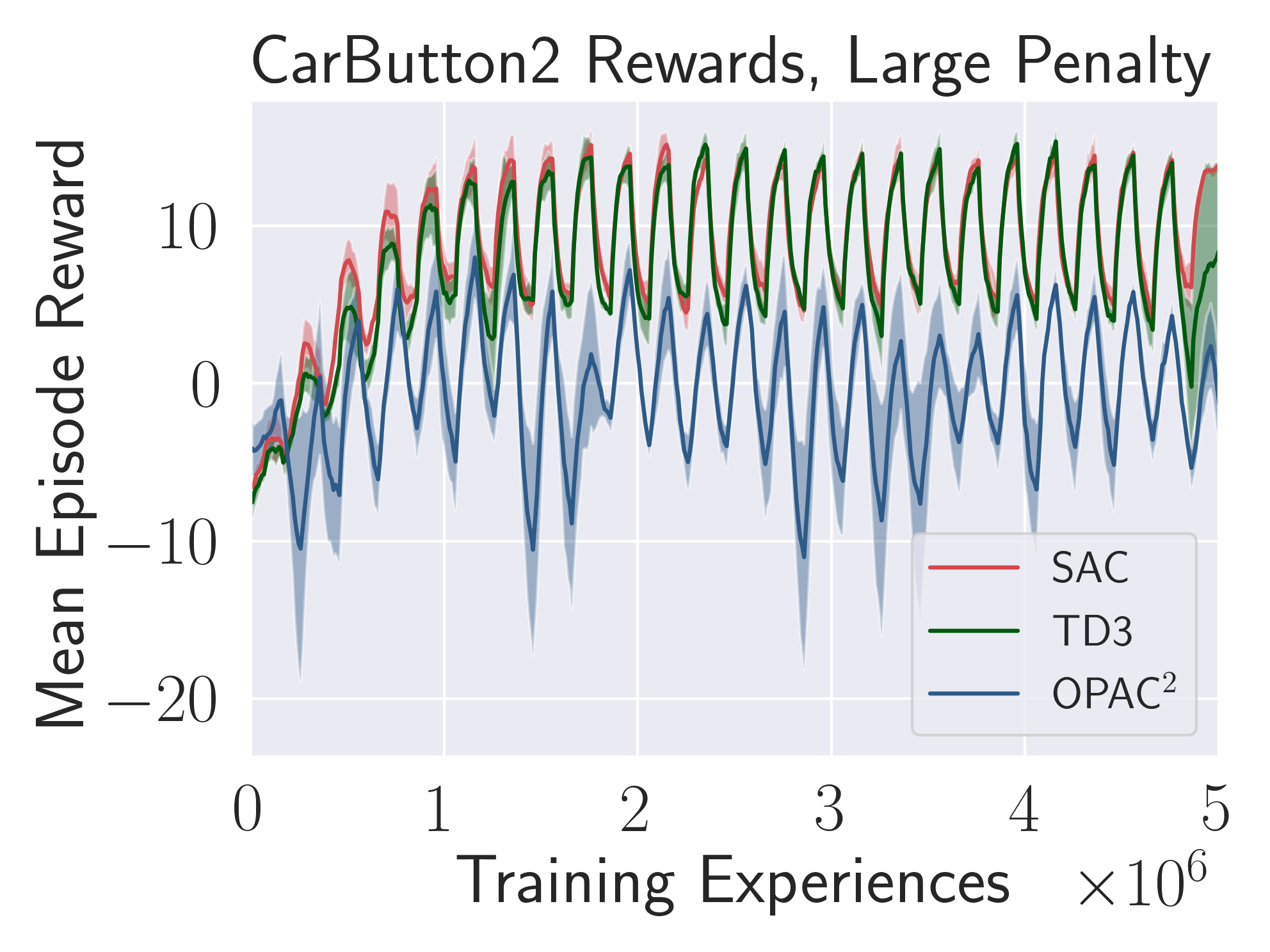}
    \includegraphics[width=0.325\textwidth]{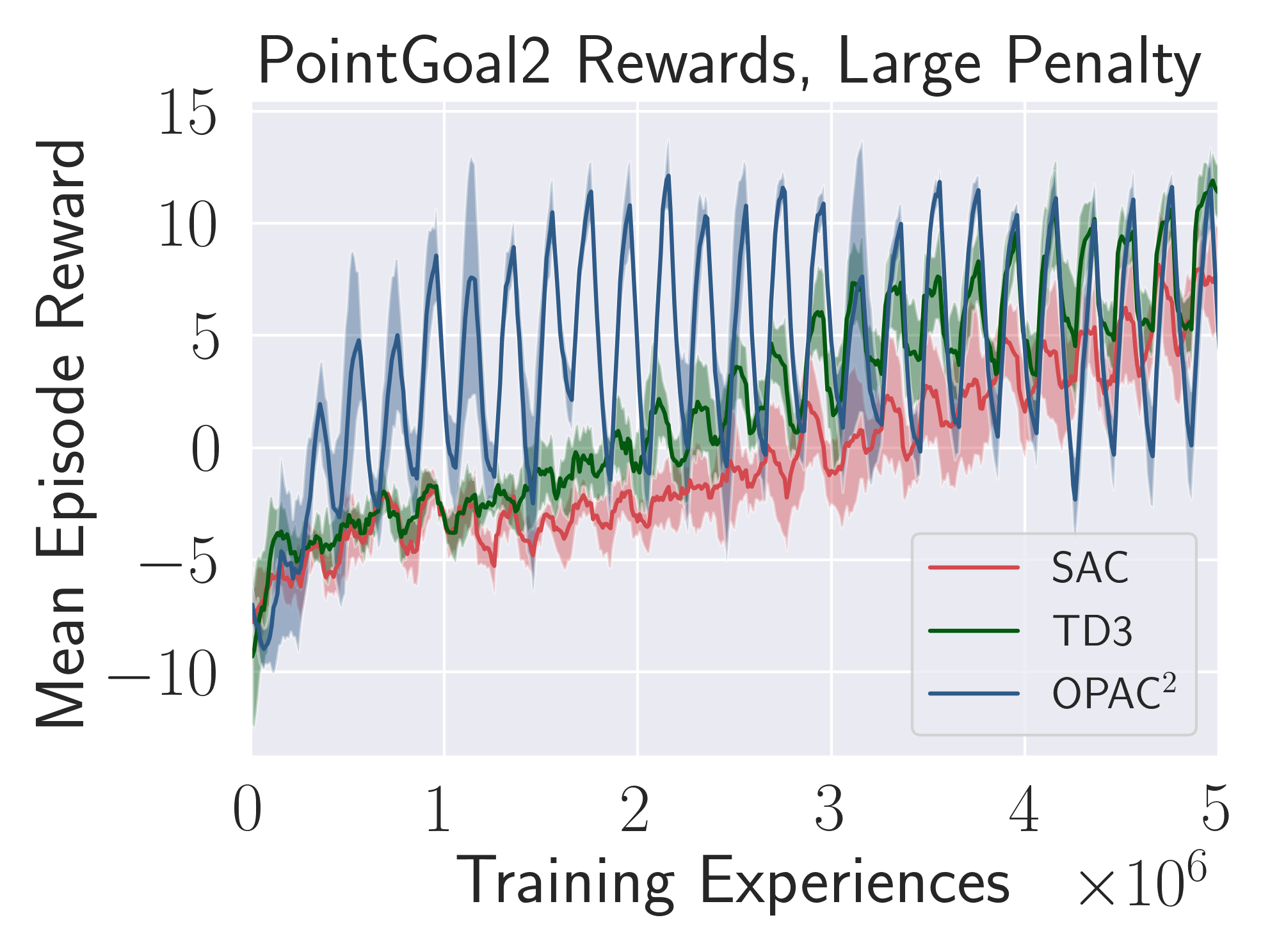}
     \includegraphics[width=0.325\textwidth]{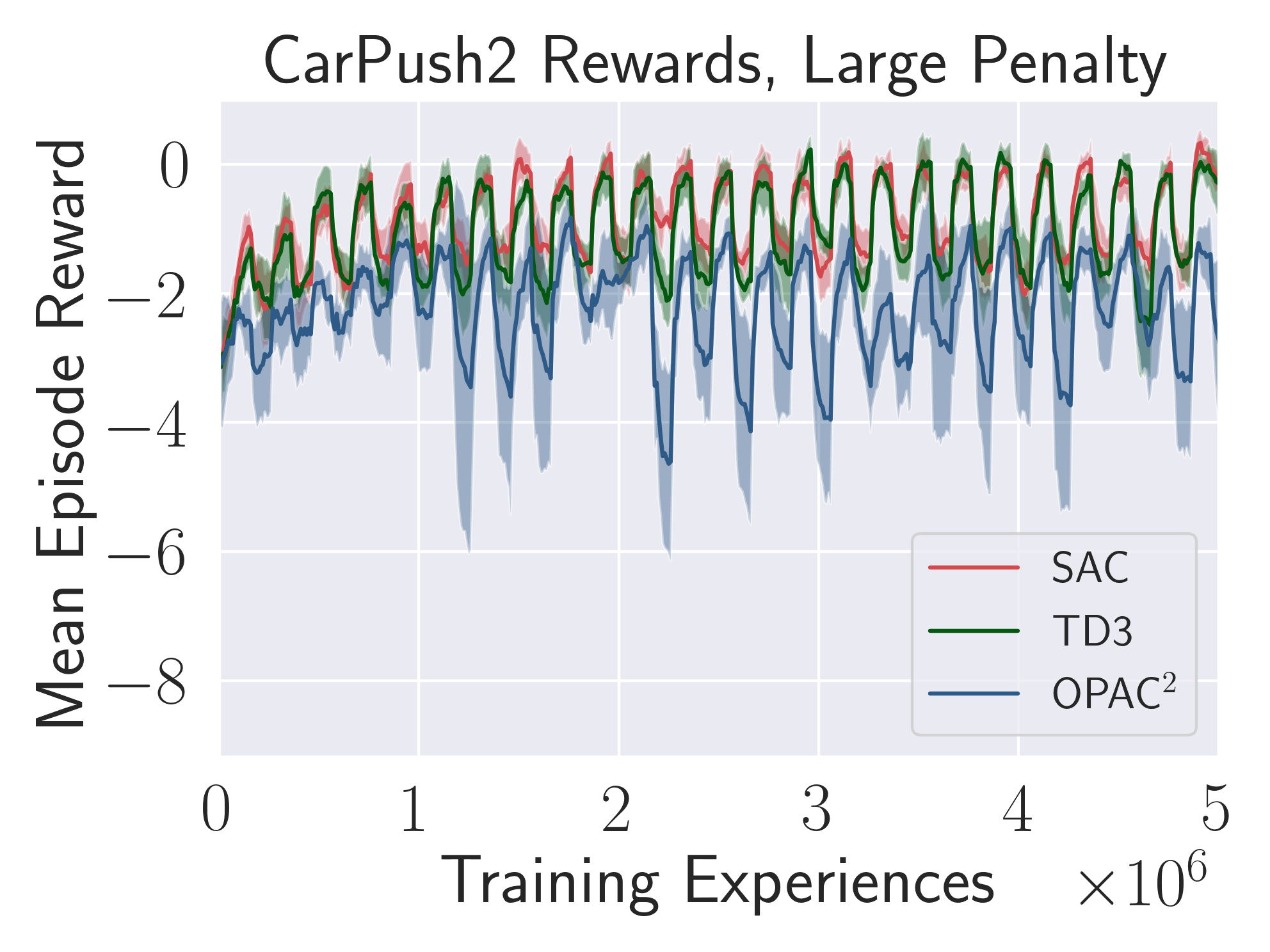}
\caption{Resetting is not found to be beneficial for OPAC$^2$, and in fact is typically detrimental.}
\end{figure}

\subsection{Entropy Regularization Strategies}\label{ent_app}
Below we plot the effect of different entropy regularization strategies in the DoggoButton2 and DoggoGoal2 environments. While not shown here, we found these trends to be consistent in the CarButton2 and PointButton2 environments.

\begin{figure}[H]
    \centering
    \includegraphics[width=0.325\textwidth]{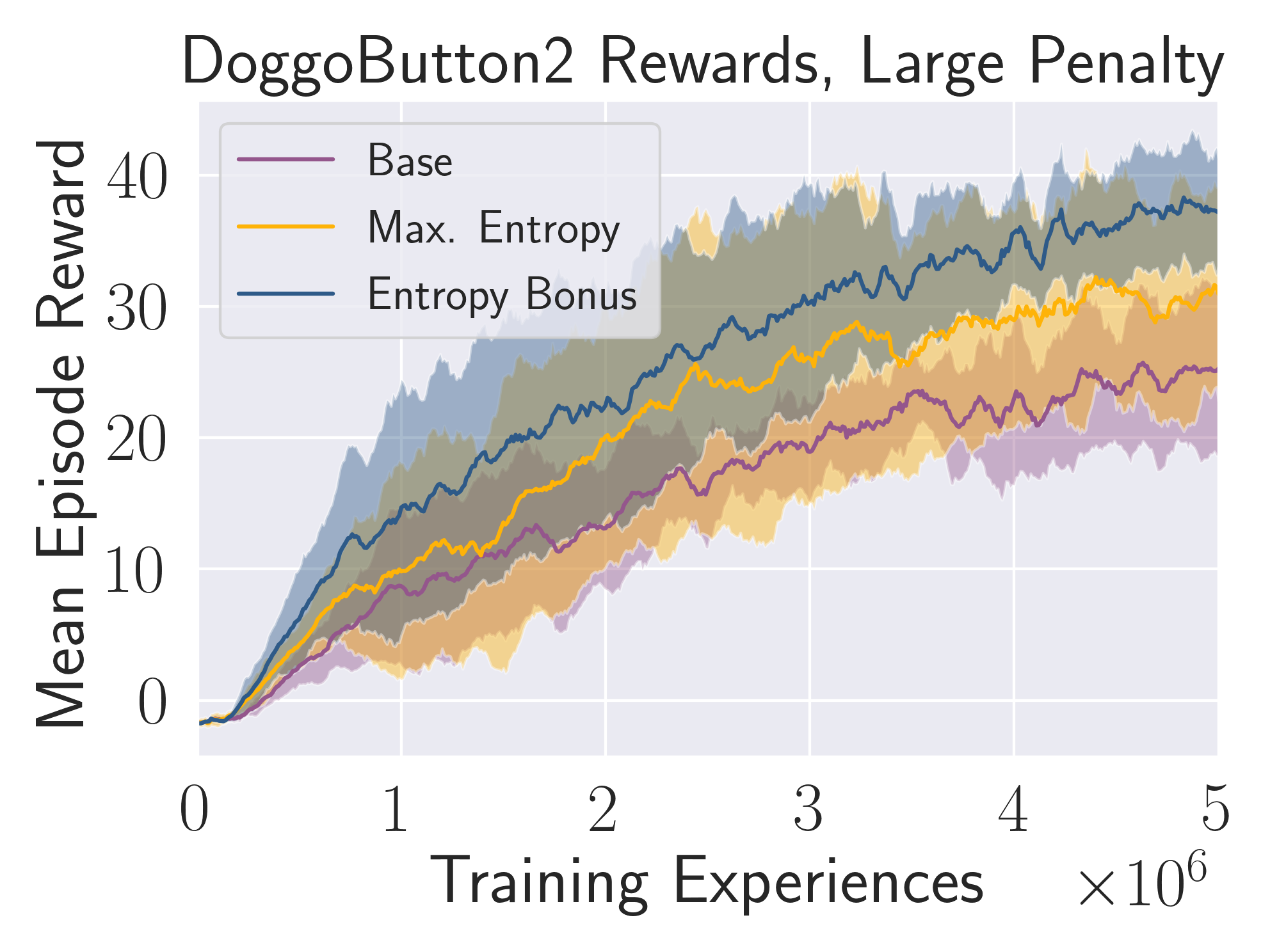}
    \includegraphics[width=0.325\textwidth]{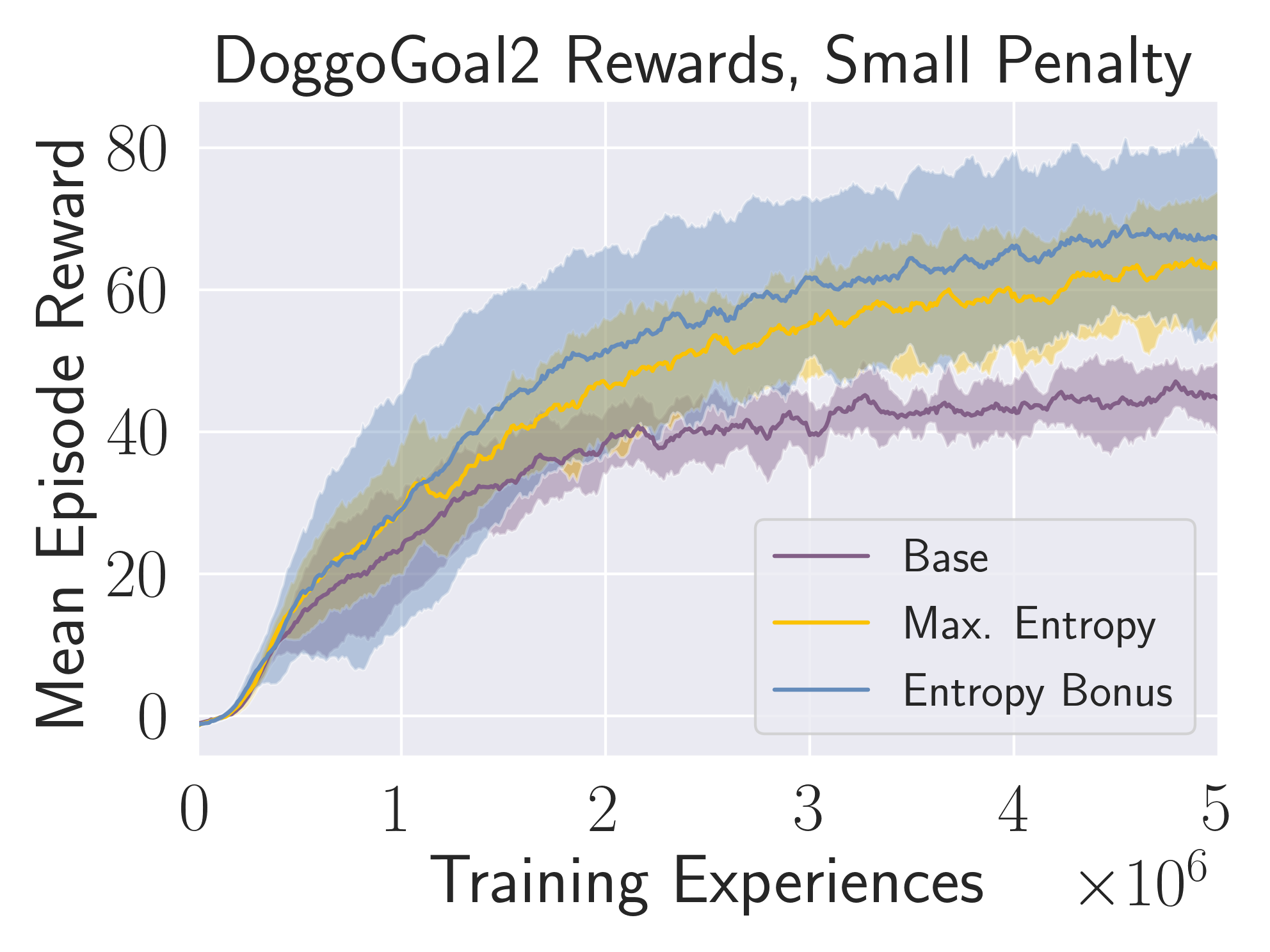}
    \includegraphics[width=0.325\textwidth]{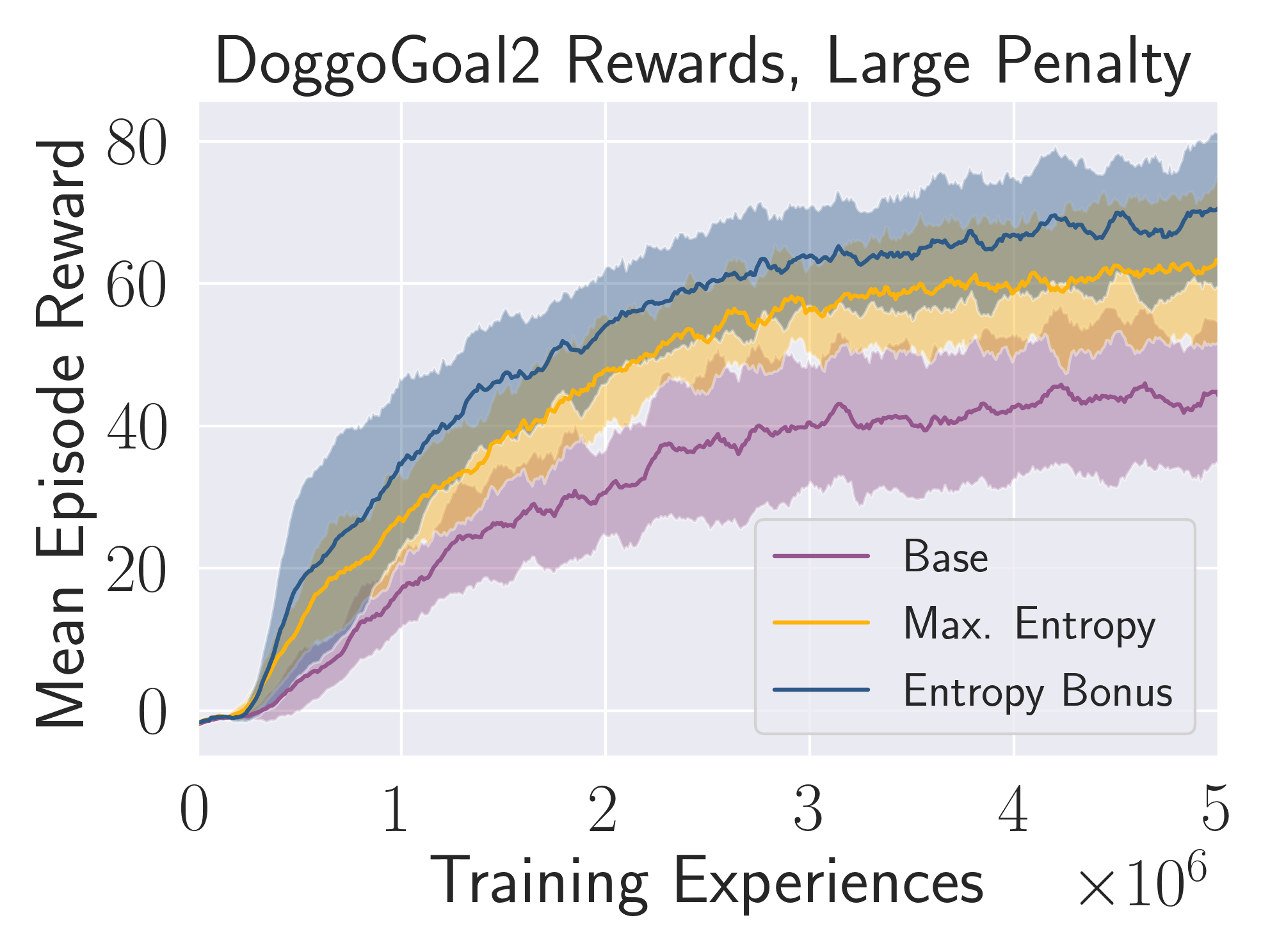}
    \caption{Different entropy regularization strategies on Doggo environments.  Our entropy bonus consistently, though sometimes narrowly, outperforms maximum entropy and no entropy regularization in these environments.}
    \label{fig:entropy_reg}
\end{figure}

\subsection{Justifying Two $Q_c$ Networks}\label{12qc_app}
We found 2 $Q_c$ networks to provide enhanced constrained learning for both SAC and TD3. We used this for the results in the main text in order to give the baselines the best chance at performing well. While the gap was larger in some environments than others, we always observed at least some improvement through the addition of an extra $Q_c$ network, for both algorithms.

\begin{figure}[H]
\centering
    \includegraphics[width=0.325\textwidth]{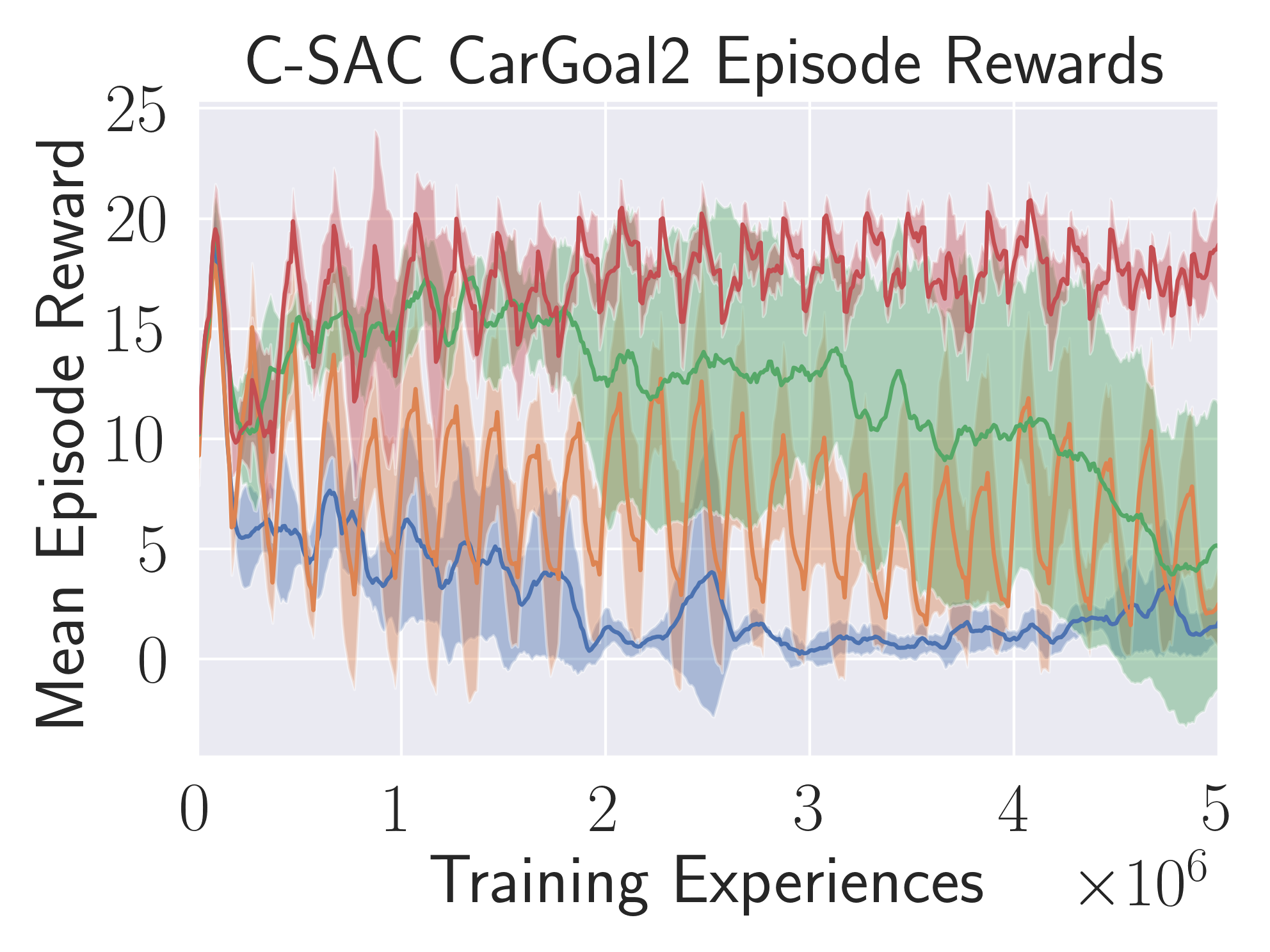}
    \includegraphics[width=0.325\textwidth]{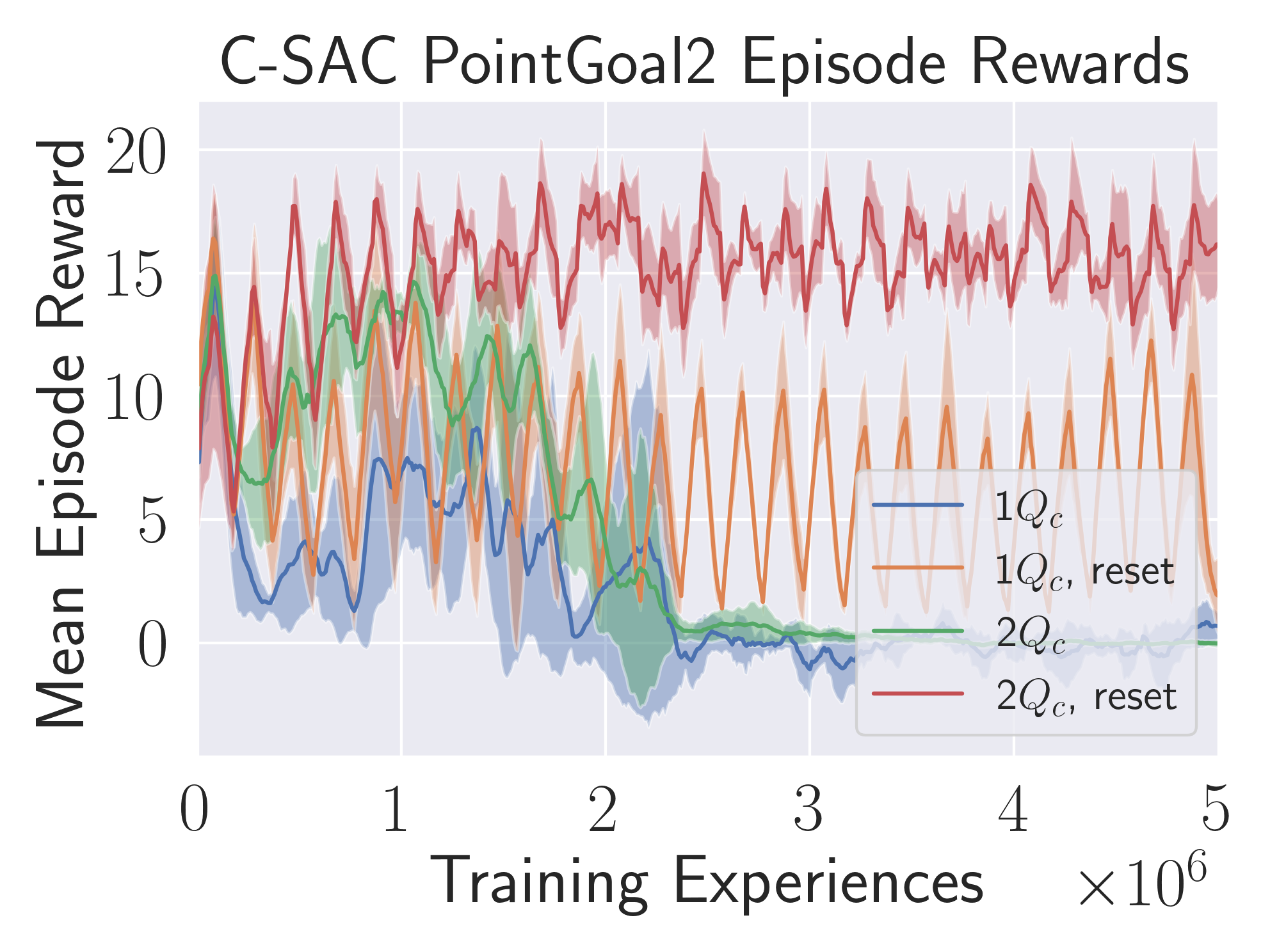}
    \includegraphics[width=0.325\textwidth]{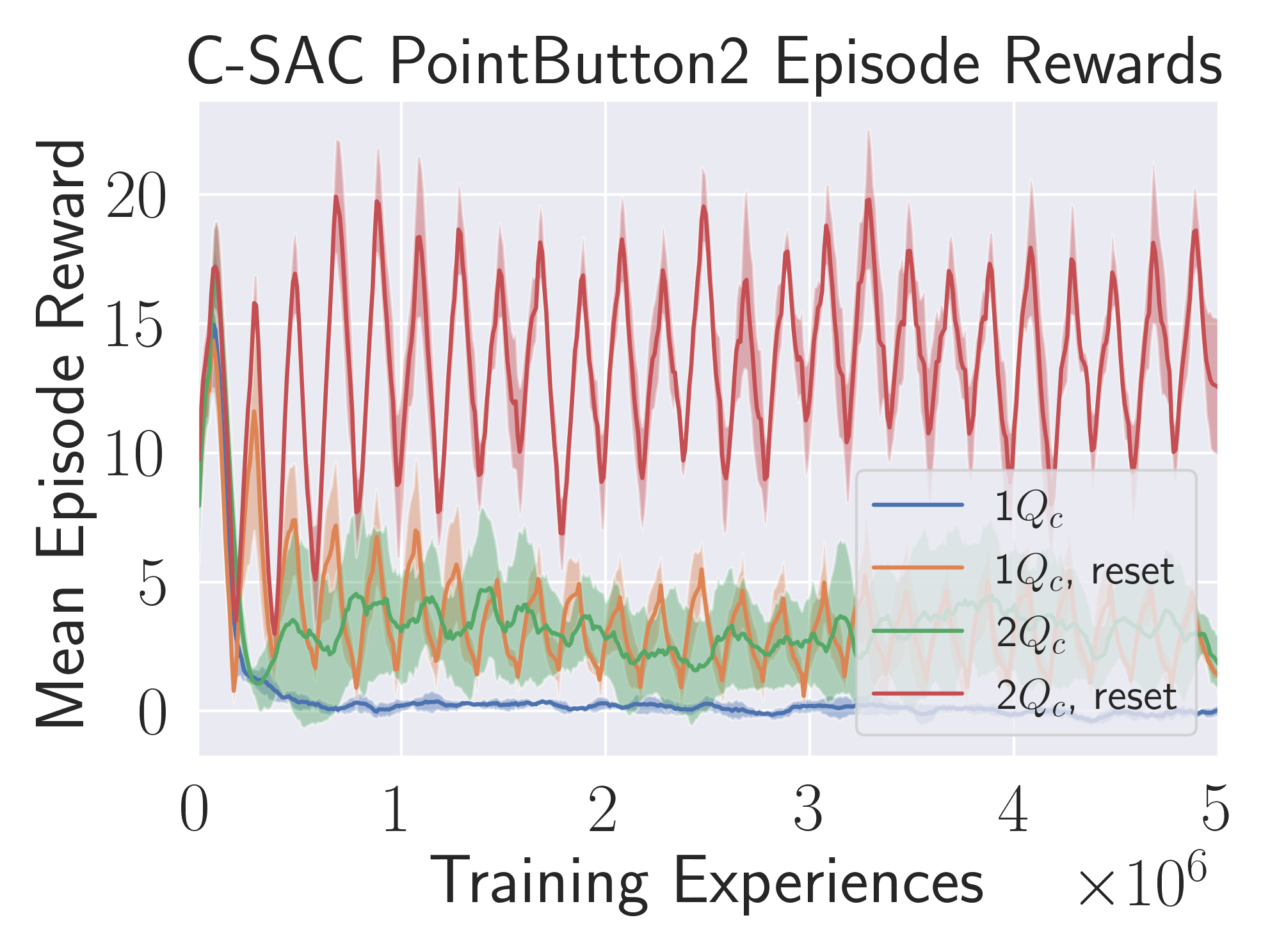}
    \includegraphics[width=0.325\textwidth]{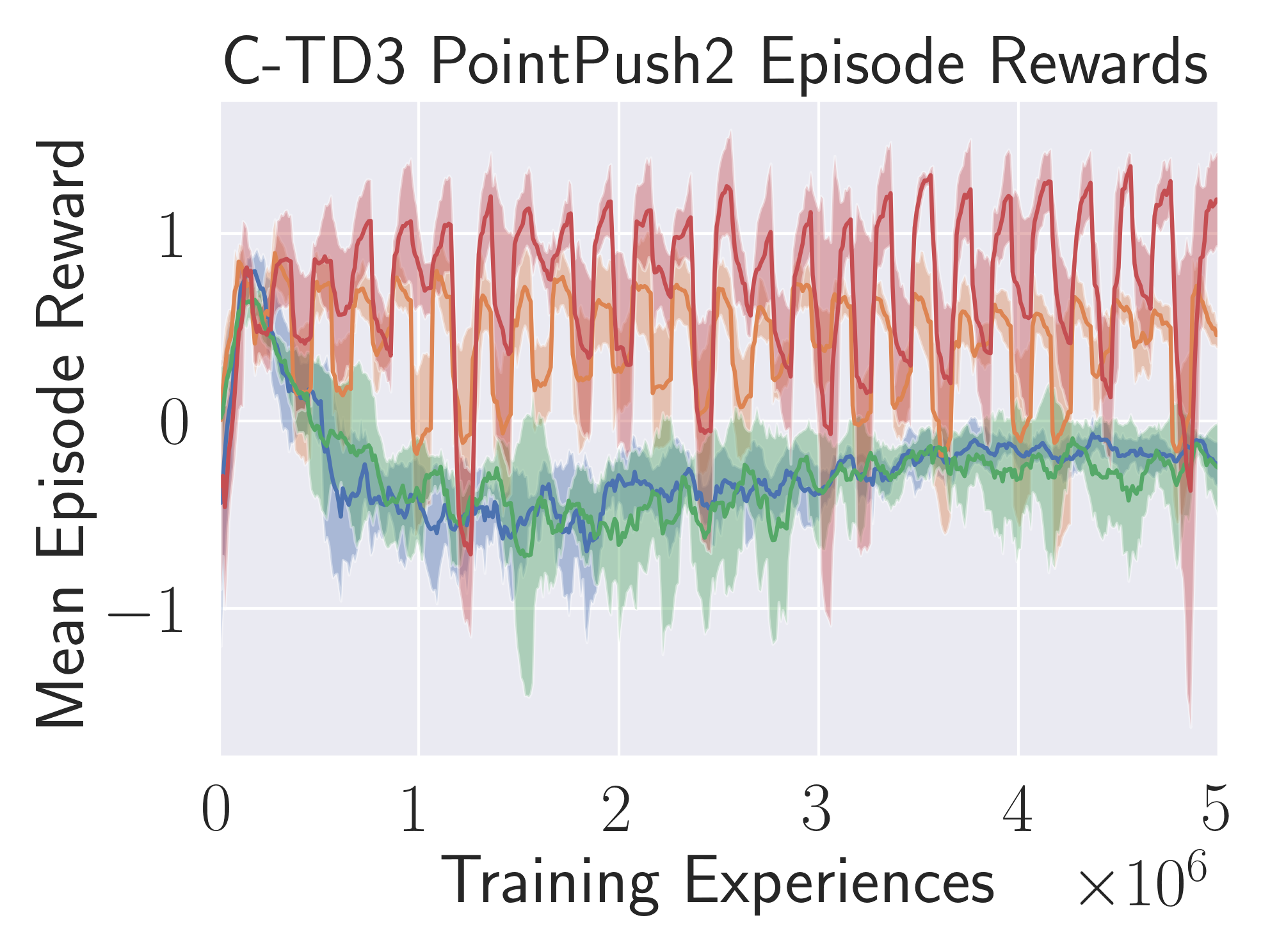}
    \includegraphics[width=0.325\textwidth]{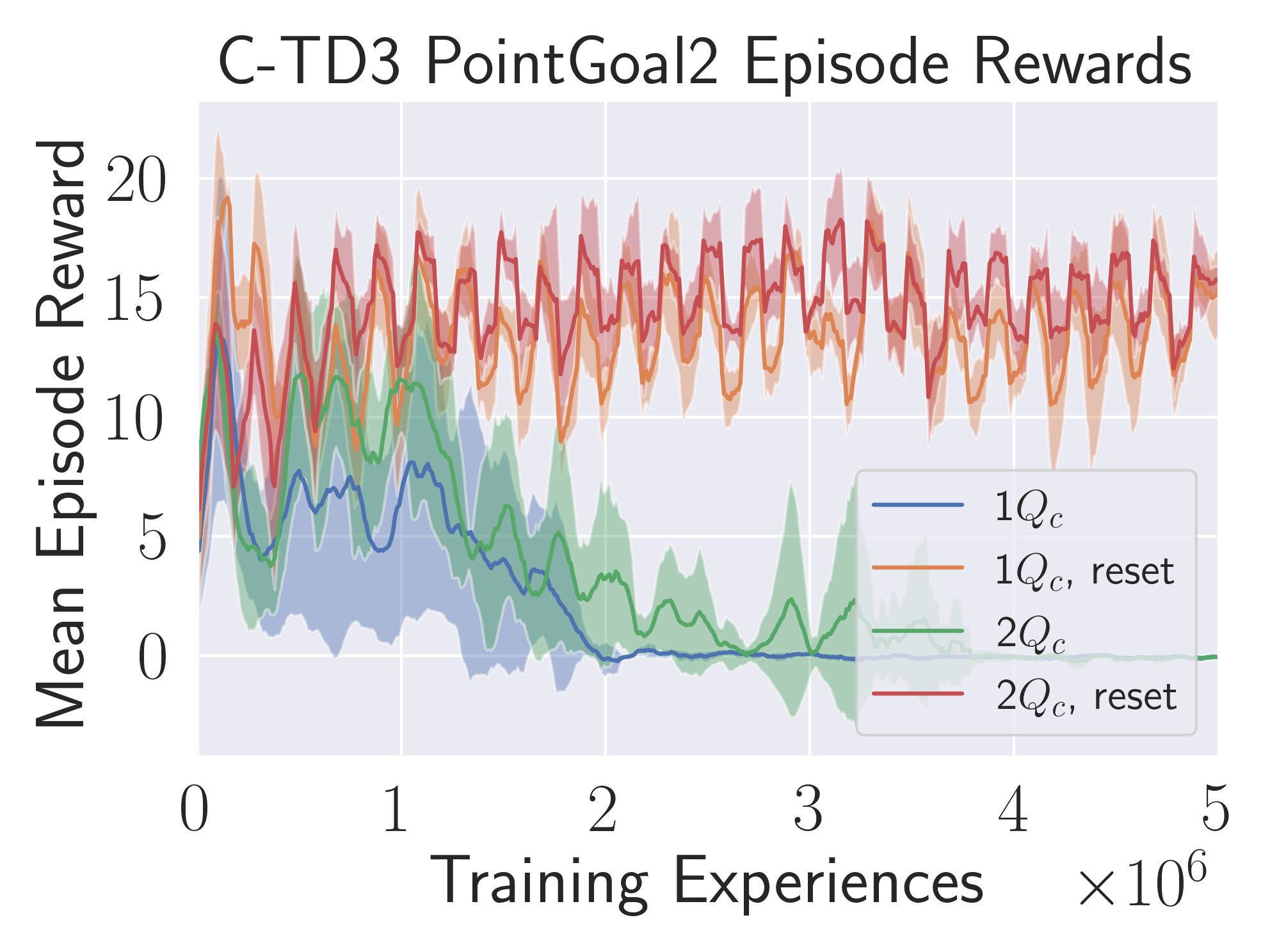}
\caption{Comparison of positive reward accumulation in constrained learning.  In all cases, cost plots were qualitatively similar.  \textbf{Top row}: comparison using SAC.  \textbf{Bottom row}: comparison using TD3.}

\end{figure}

\section{Additional Constrained Results: Safety Gym}\label{con_app}
Here we include additional reward and cost results from Safety Gym.  As mentioned in the main text, OPAC$^2$ was able to match specified cost levels in every environment tested, while providing higher positive rewards than other methods.

\begin{figure}[H]
\centering
    \includegraphics[width=0.325\textwidth]{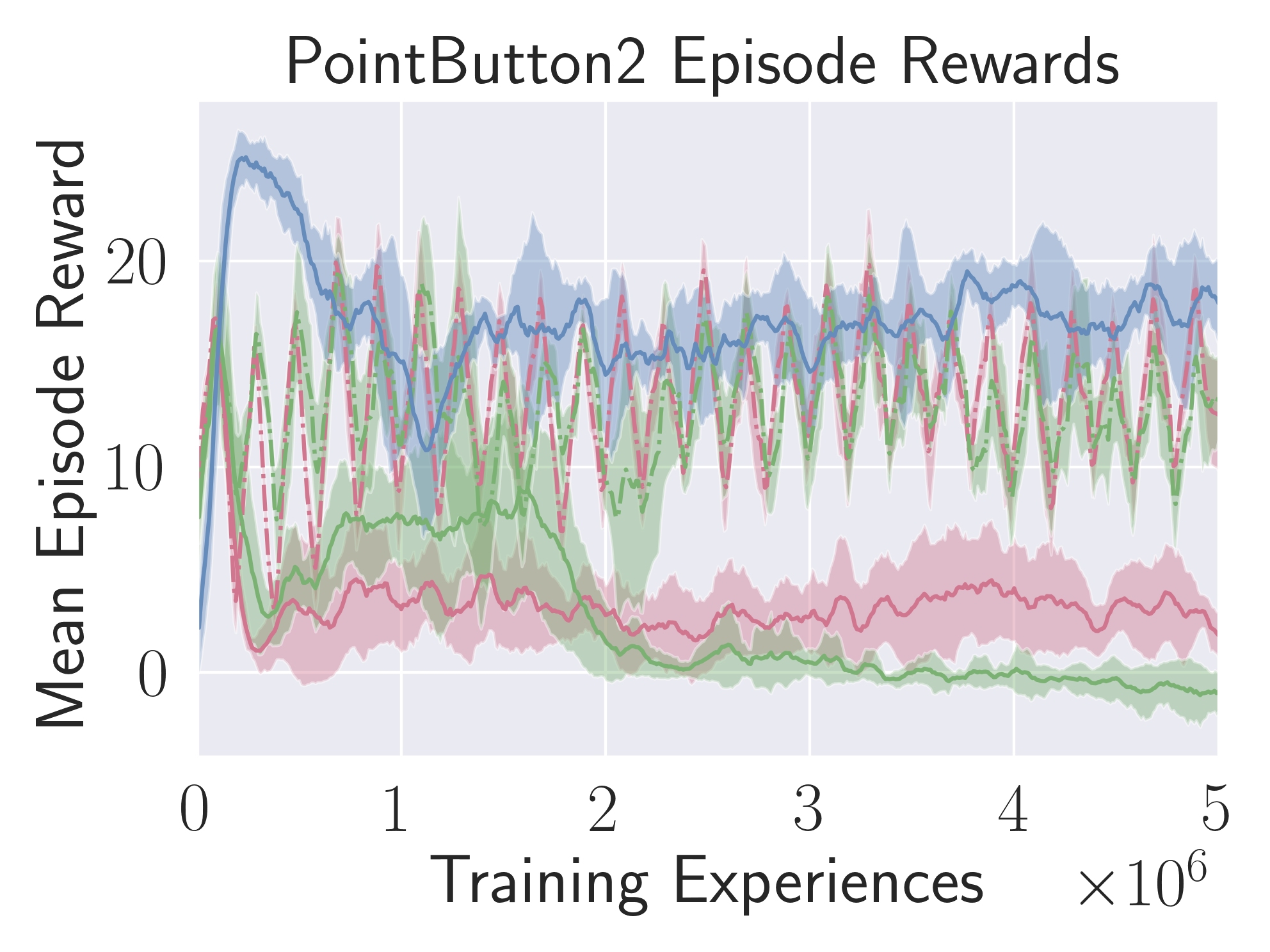}
    \includegraphics[width=0.325\textwidth]{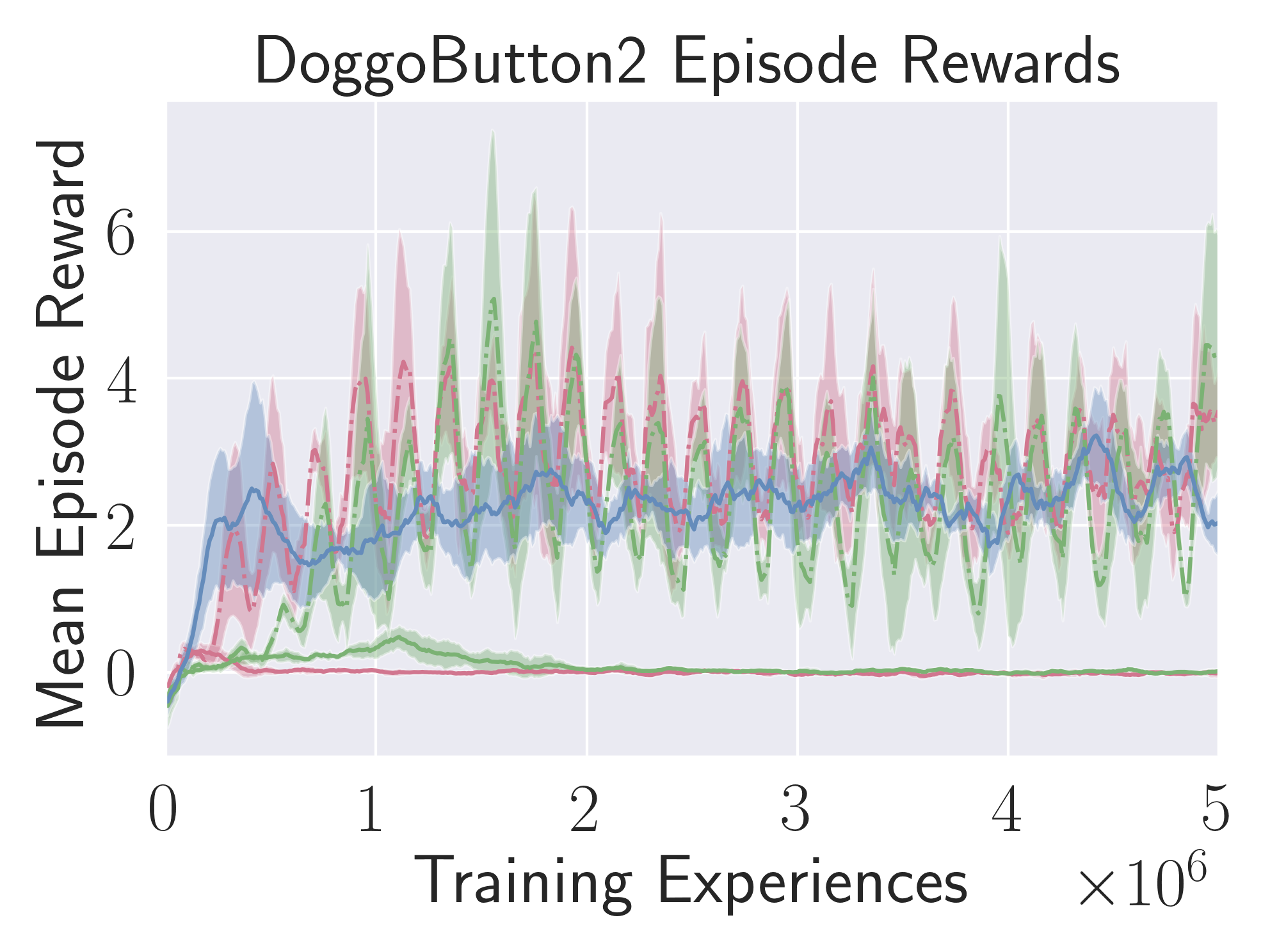}
    \includegraphics[width=0.325\textwidth]{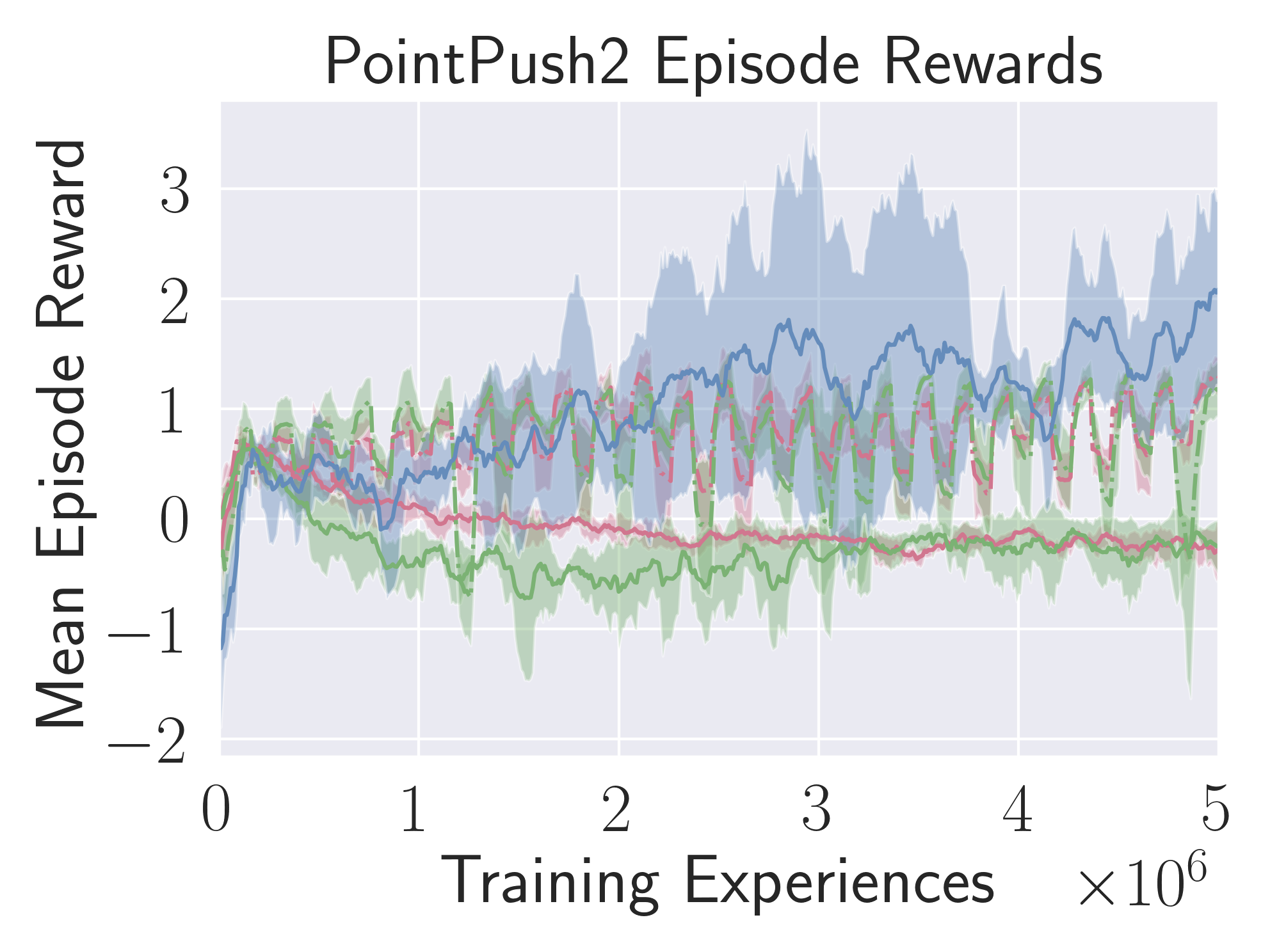}
    \includegraphics[width=0.325\textwidth]{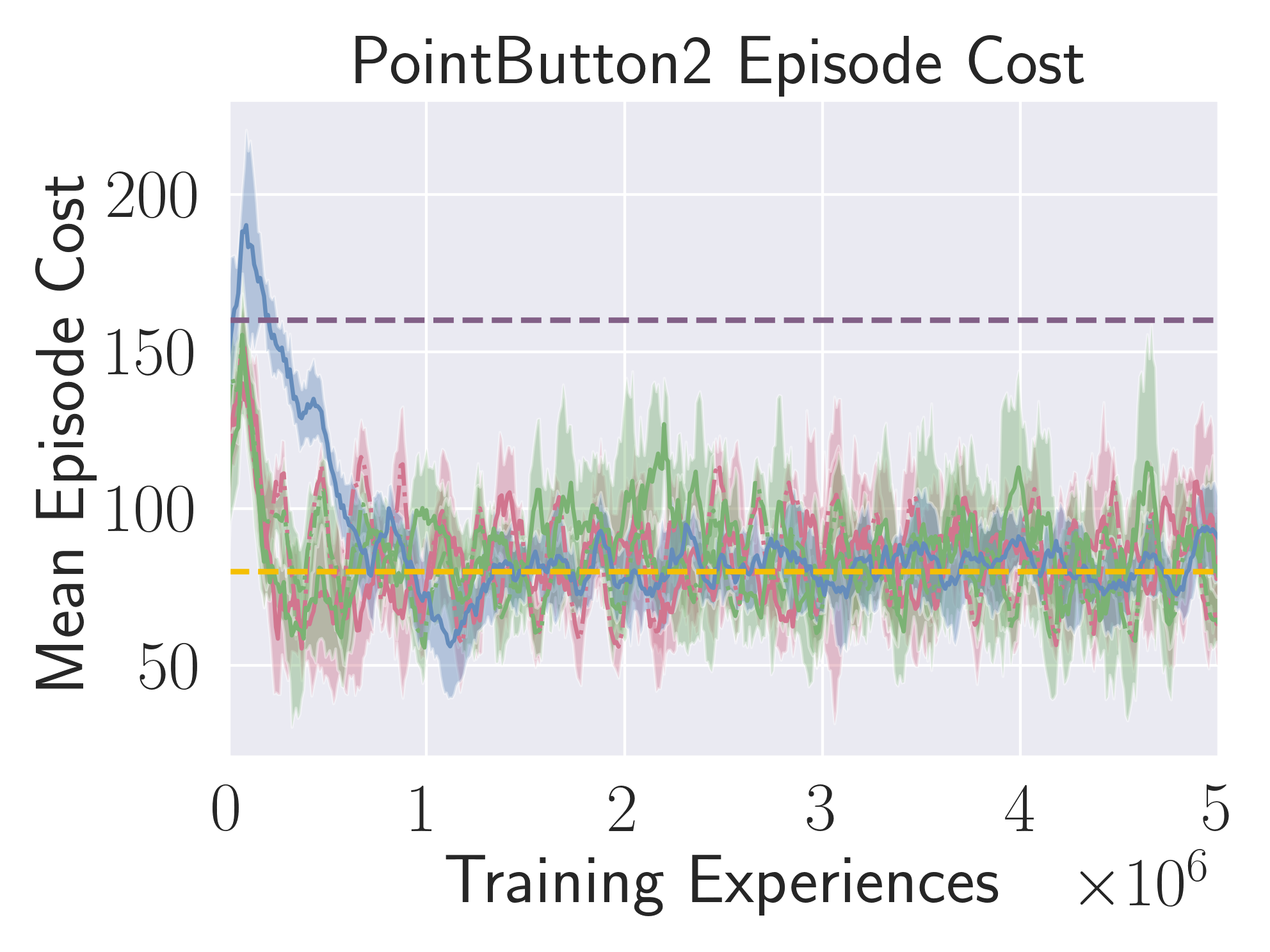}
    \includegraphics[width=0.325\textwidth]{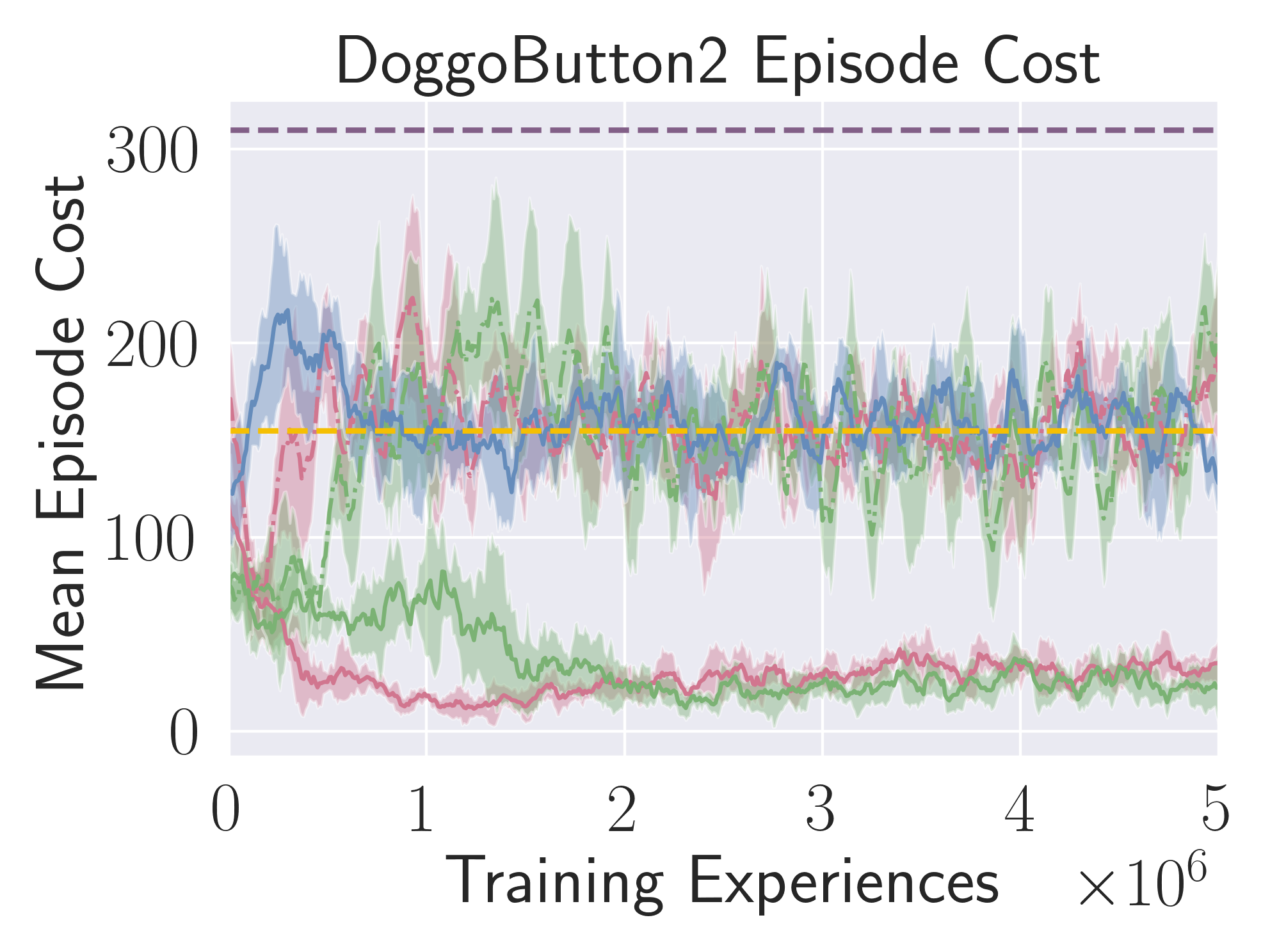}
    \includegraphics[width=0.325\textwidth]{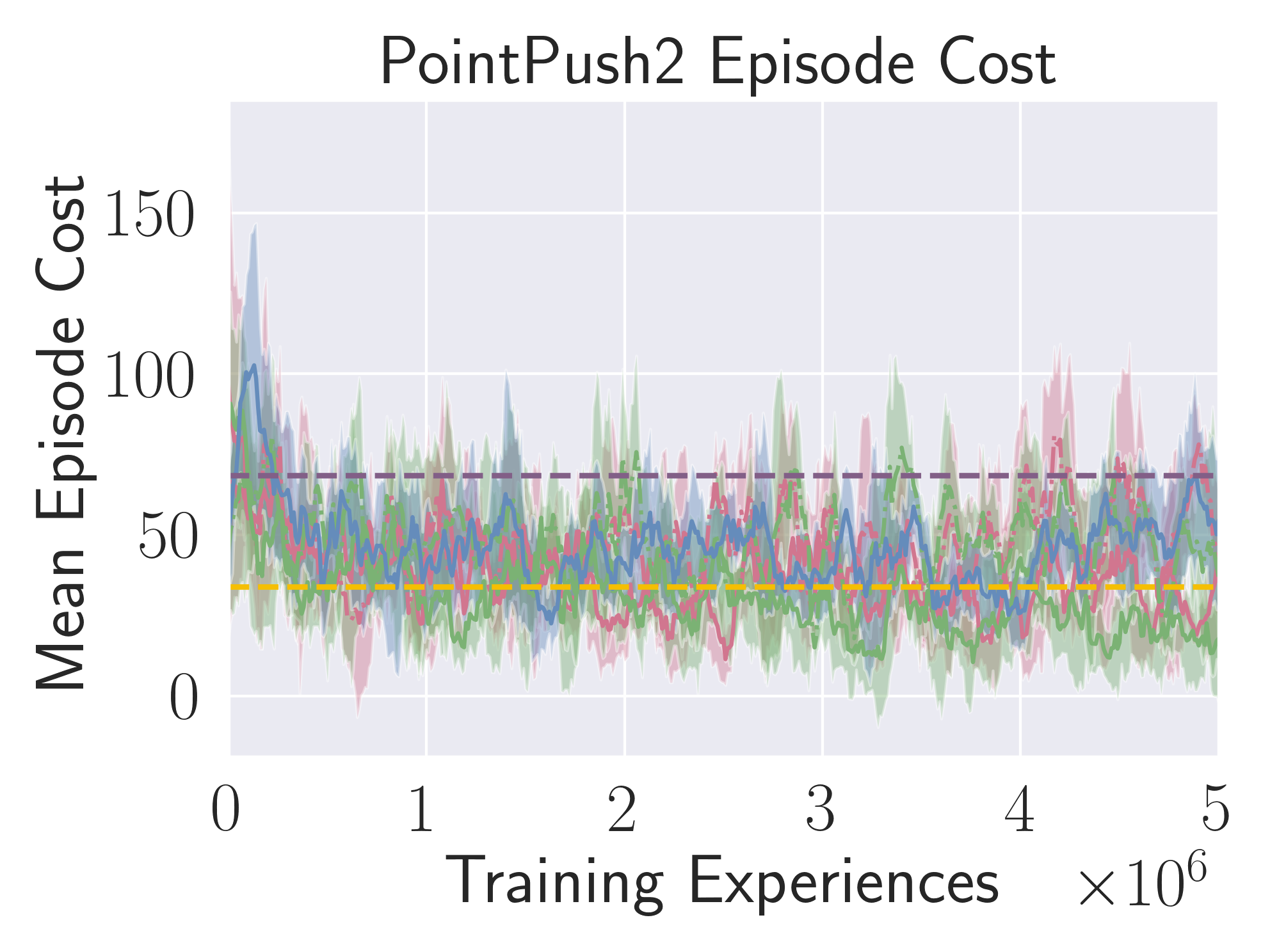}
    \includegraphics[width=0.9\textwidth]{figures/legend_3.png}
    \includegraphics[width=0.325\textwidth]{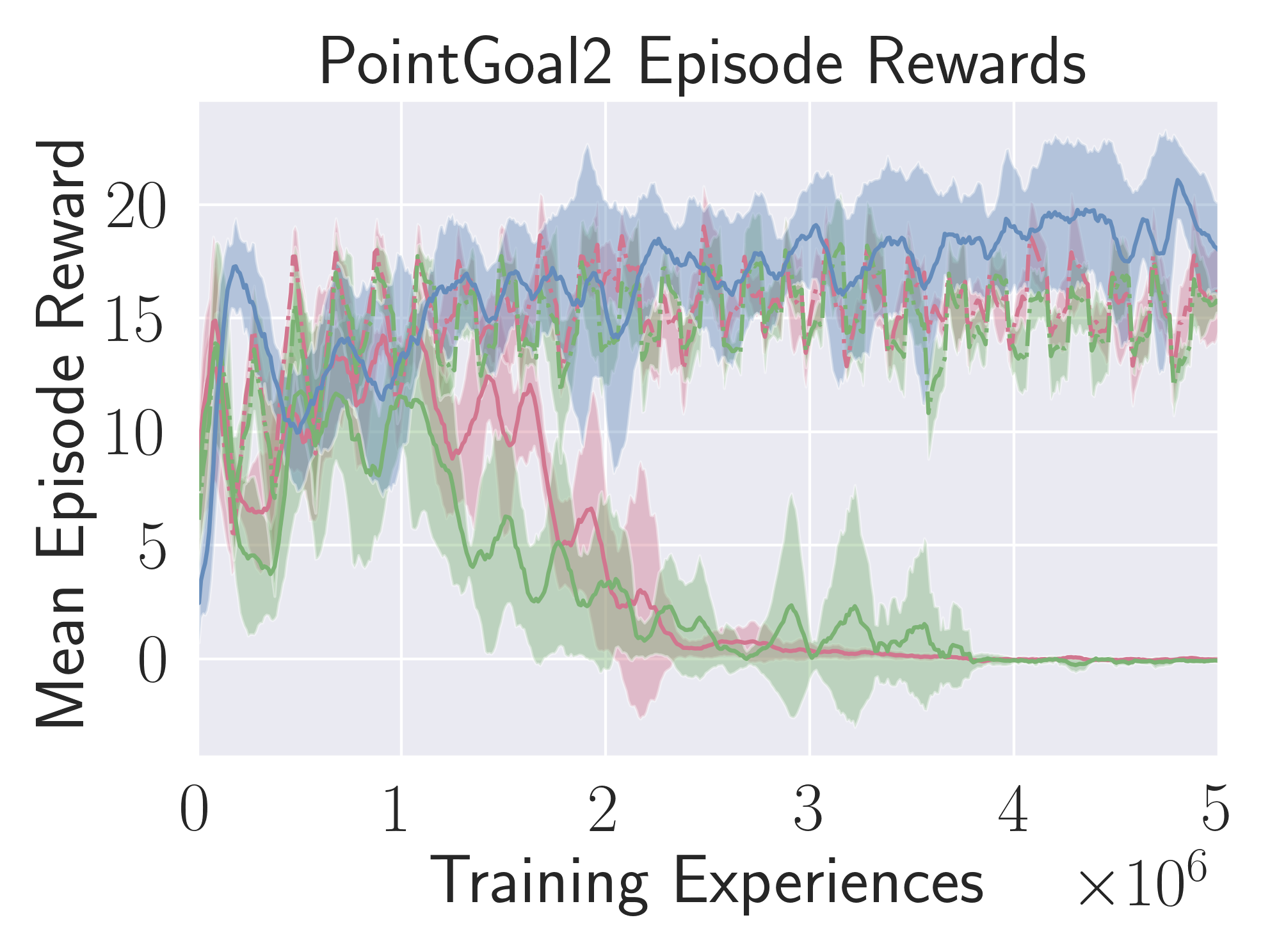}
    \includegraphics[width=0.325\textwidth]{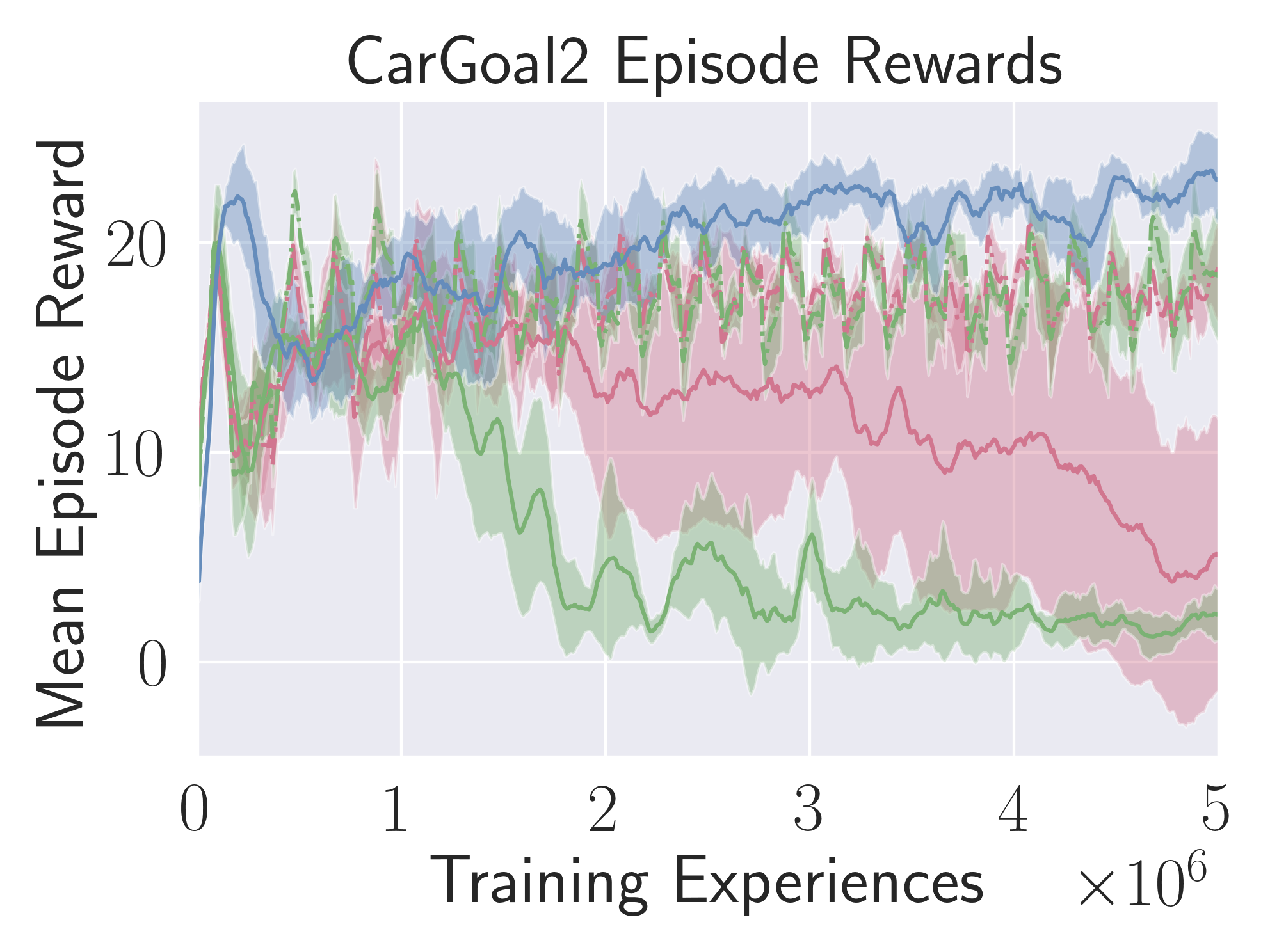}\\
    \includegraphics[width=0.325\textwidth]{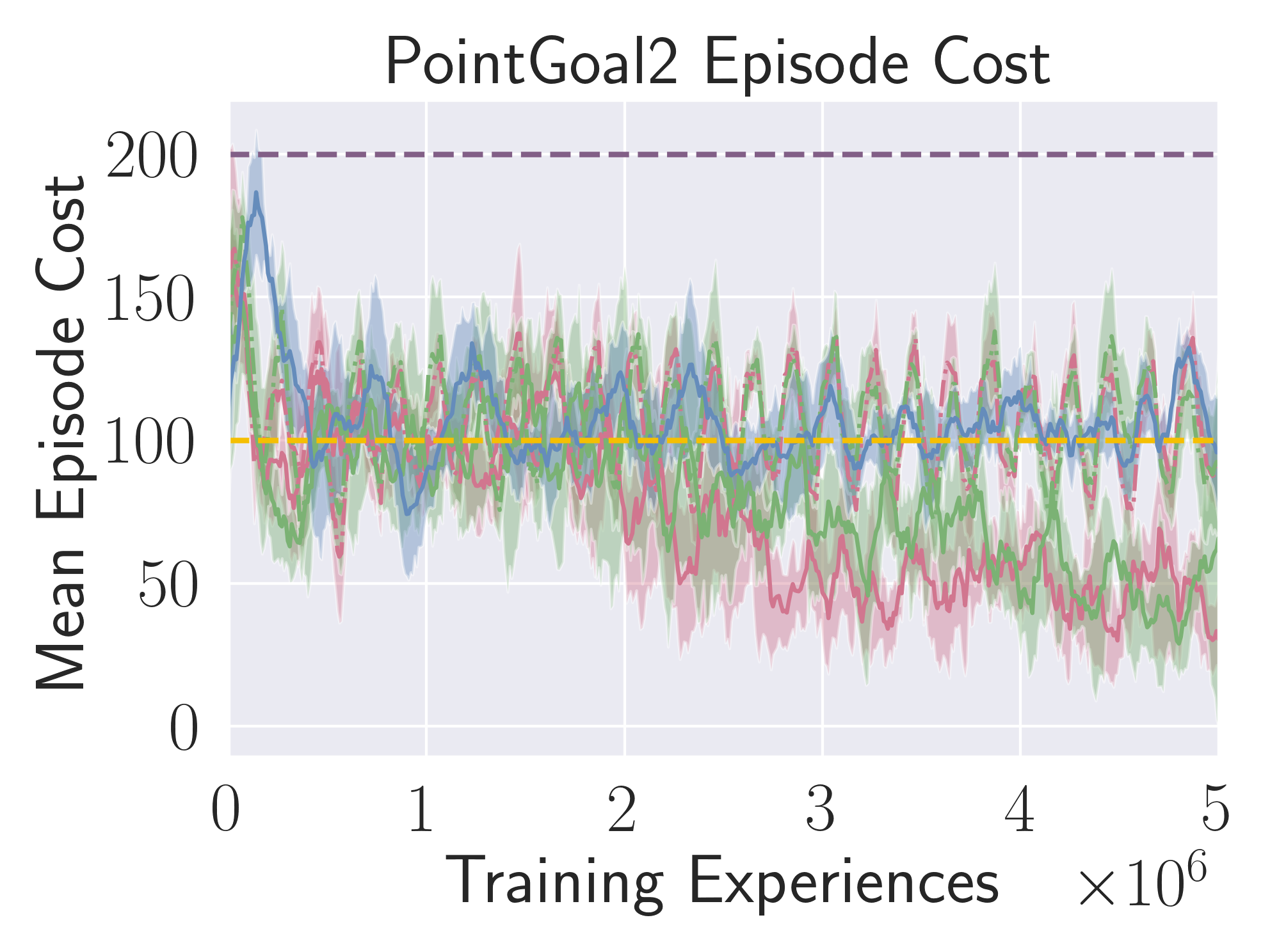}
    \includegraphics[width=0.325\textwidth]{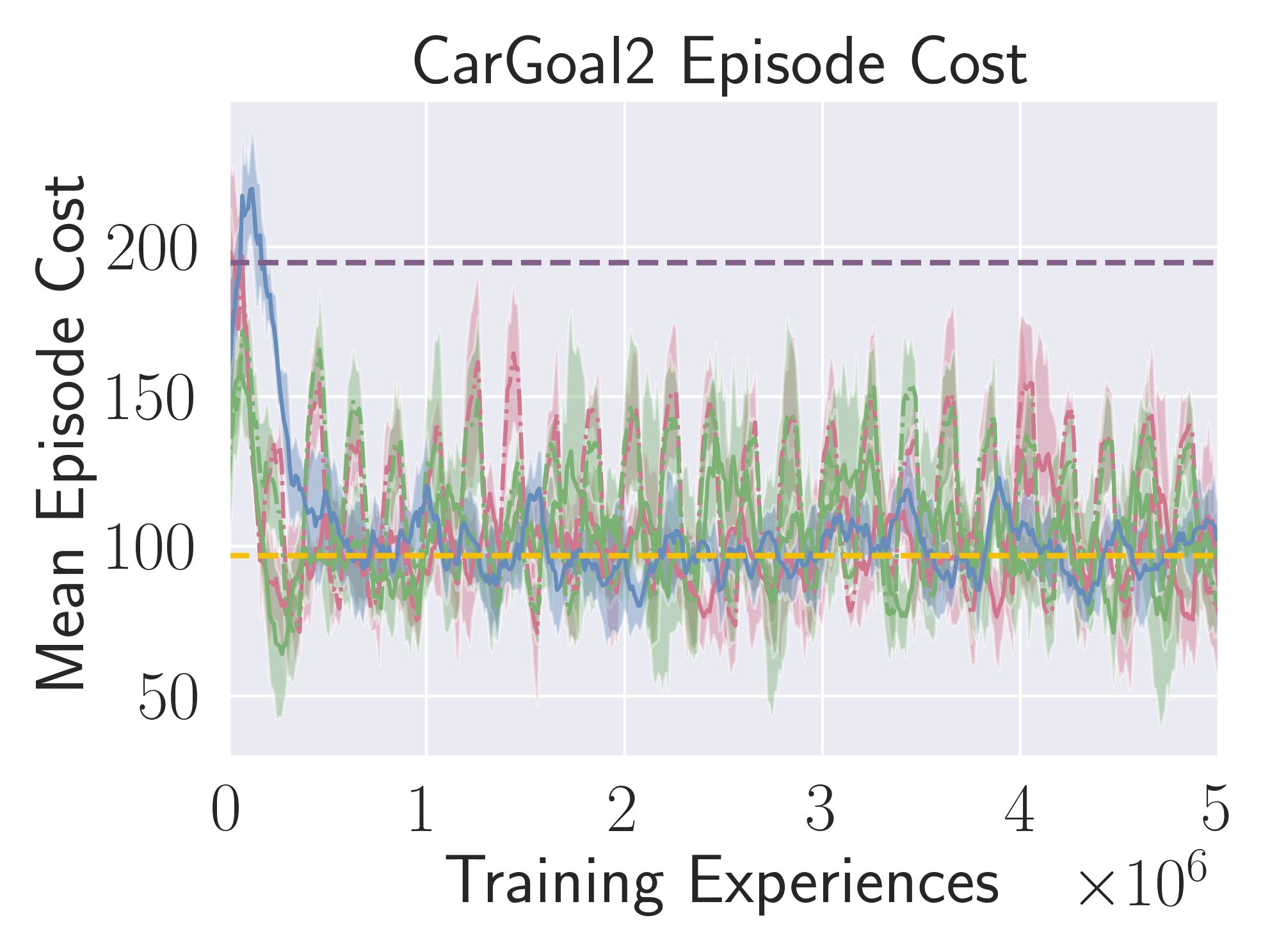}
    \includegraphics[width=0.9\textwidth]{figures/legend_3.png}
    \caption{Additional results from constrained learning experiments from Safety Gym. \textbf{Top rows}: reward accumulated by agents. \textbf{Bottom rows}: cost converges to the target level (yellow). Note that reward accumulation initially increases sharply while $\beta$, the penalty weighting, is low. It dips and then rises again as performance is optimized at the appropriate cost level.}
    \label{fig:con_app_fig}
\end{figure}

We additionally provide a few extra plots reflecting the superior ability of OPAC$^2$ to mitigate validation TD error for both cost and reward in constrained learning, compared to SAC and TD3.

\begin{figure}[H]
\centering
\includegraphics[width=0.325\textwidth]{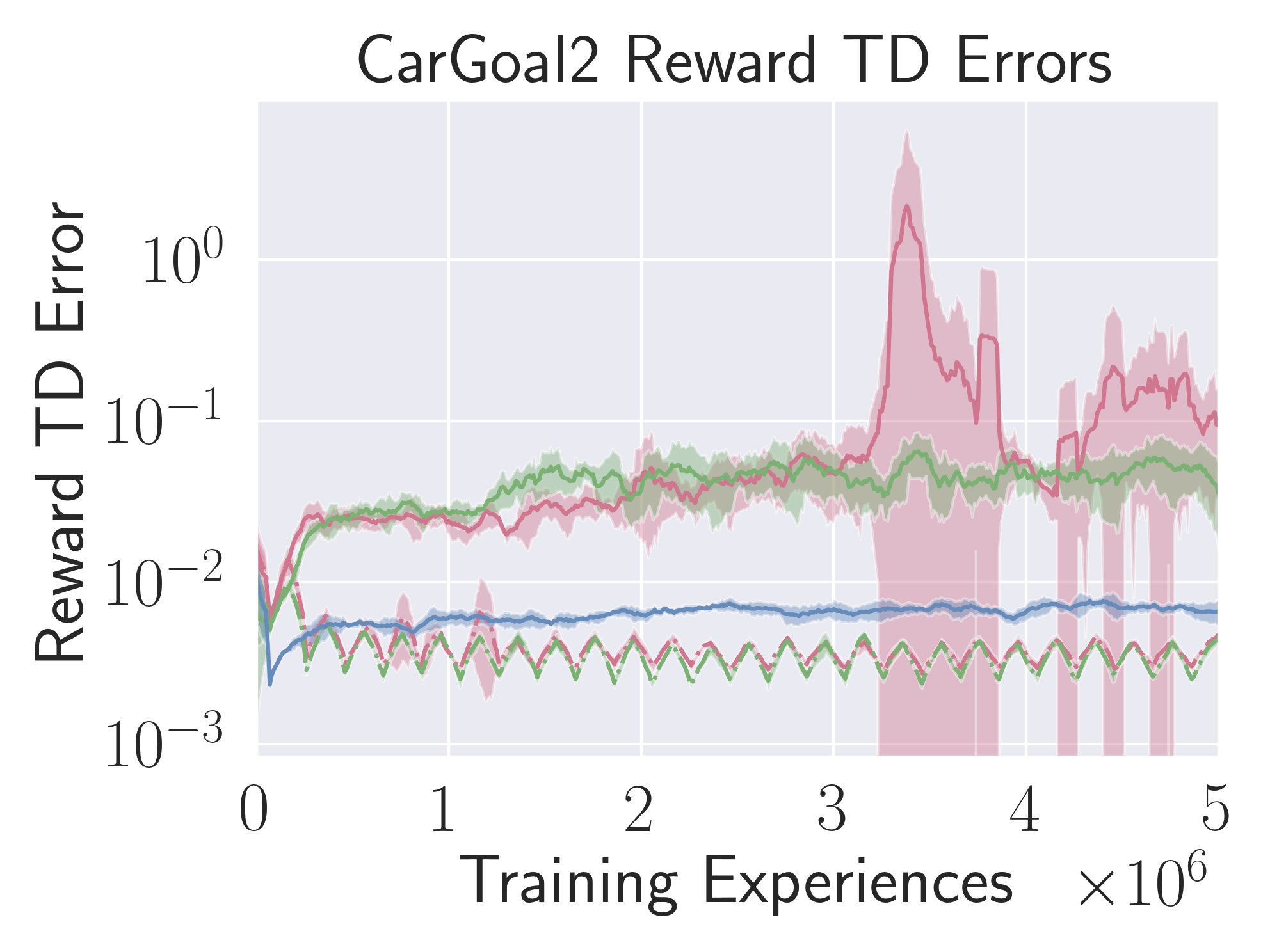}
\includegraphics[width=0.325\textwidth]{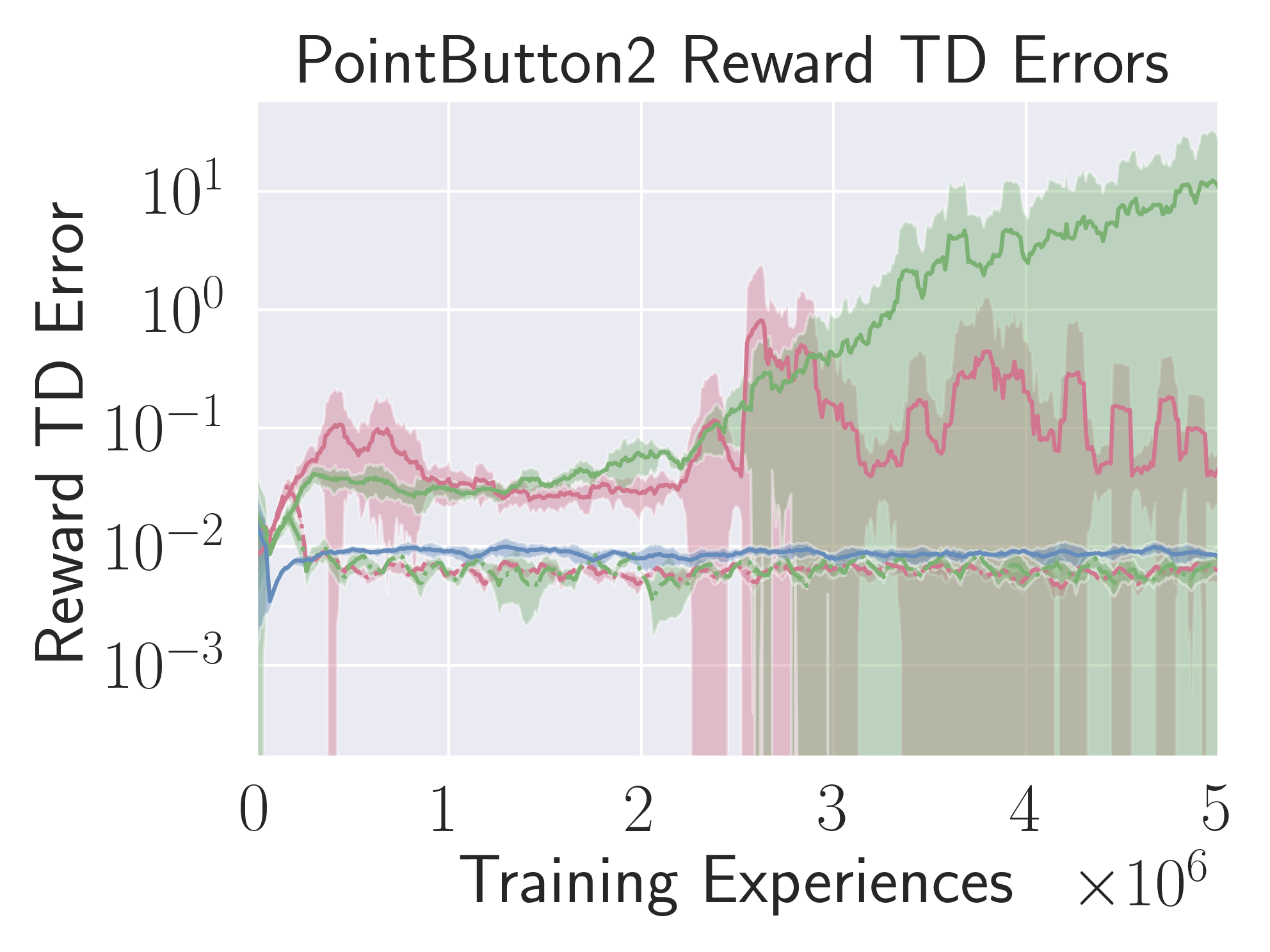}
\includegraphics[width=0.325\textwidth]{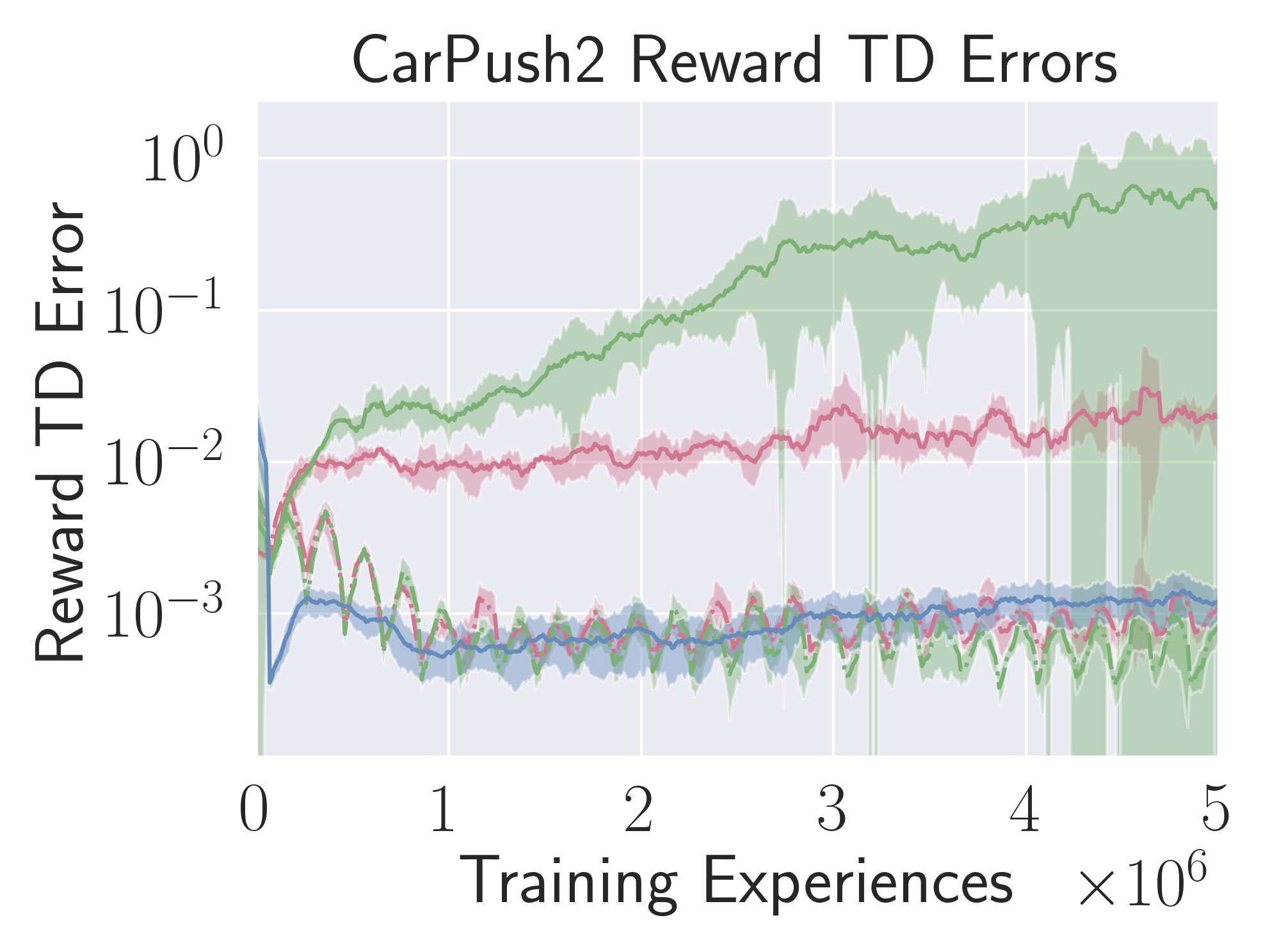}
\includegraphics[width=0.325\textwidth]{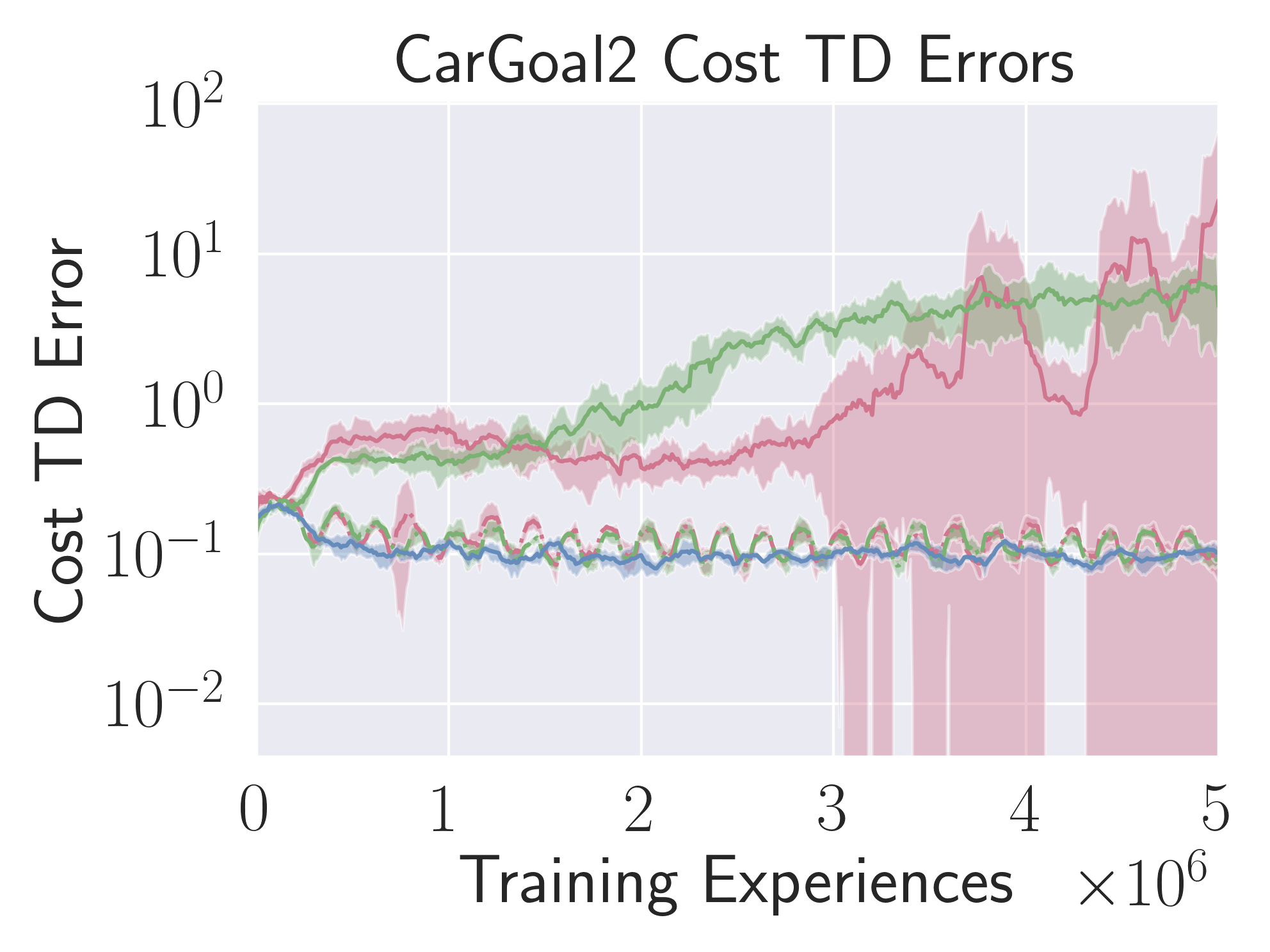}
\includegraphics[width=0.325\textwidth]{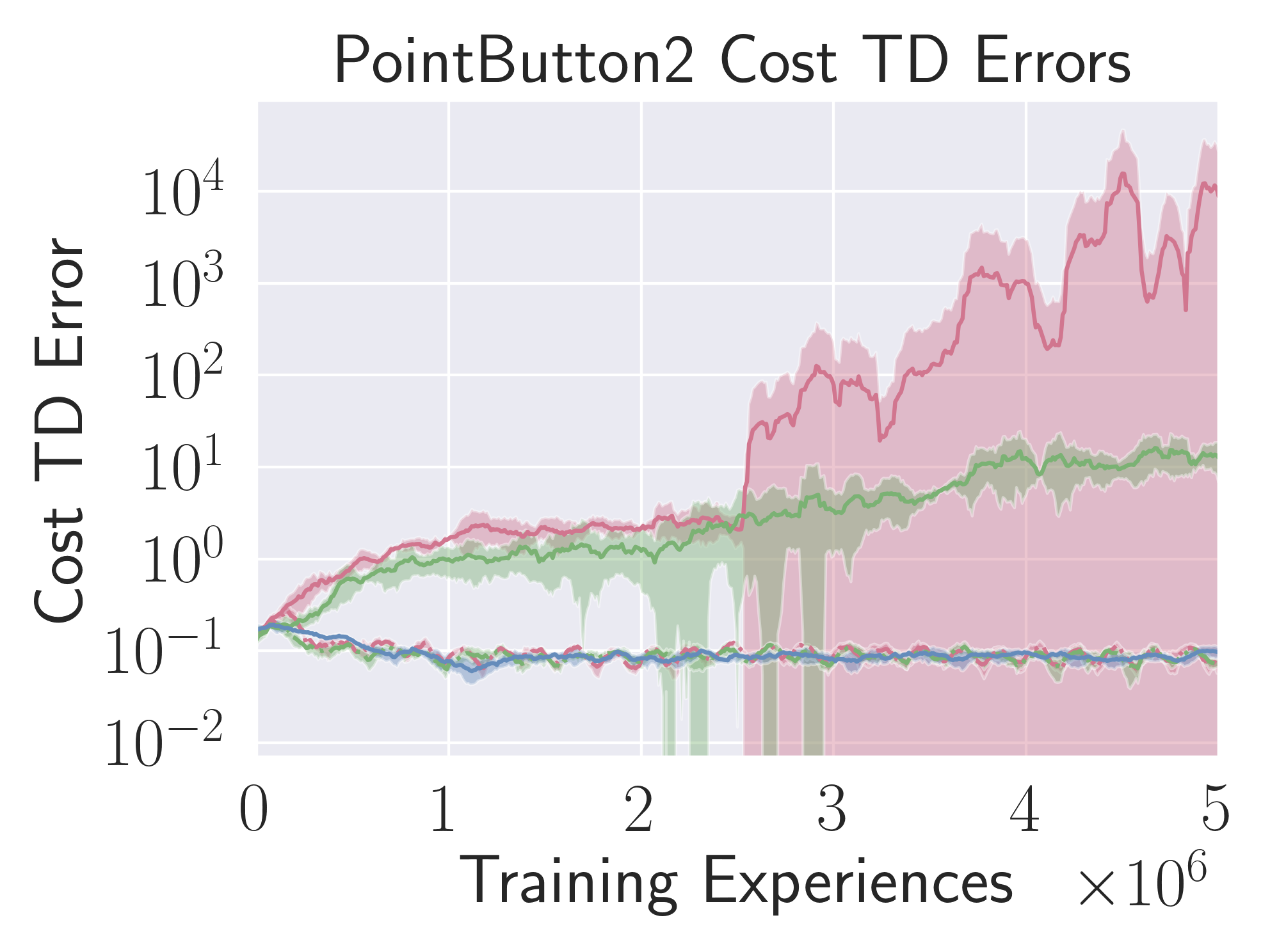}
\includegraphics[width=0.325\textwidth]{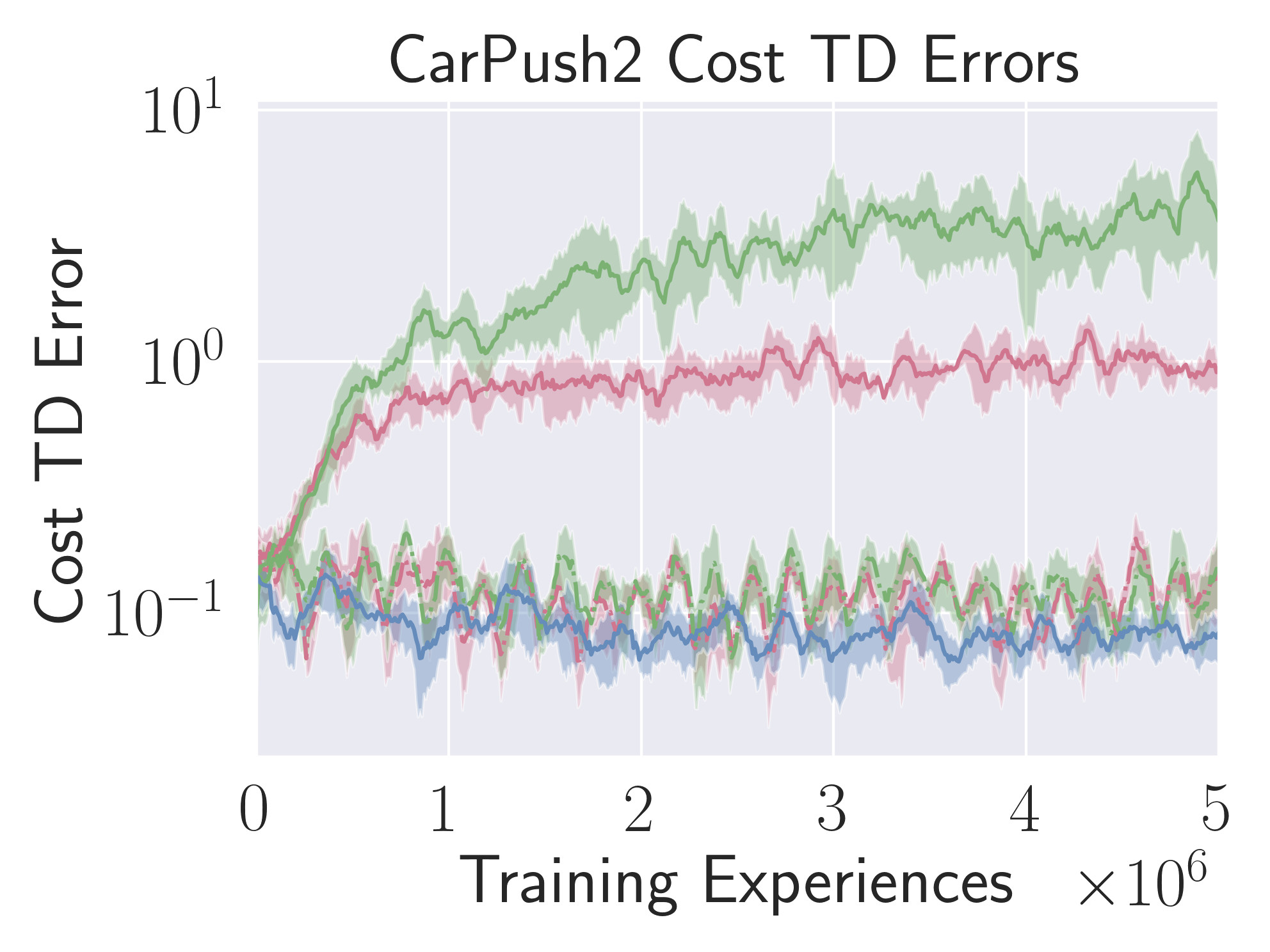}
\includegraphics[width=0.9\textwidth]{figures/legend_1.png}
\caption{\textbf{Top row}: Validation TD error for reward in constrained learning.  \textbf{Bottom row}: Validation TD error for cost in constrained learning.}
\end{figure}

\section{Additional Unconstrained Results: DeepMind Control}\label{app:dmc_full}
For completeness, we include the learning curves for our evaluation of OPAC$^2$, SAC, and TD3 in each of the 10 DeepMind Control Suite environments we tested.  We also provide an overall average plot, across the 10 environments.

\begin{figure}[H]
    \centering

    \begin{subfigure}{.24\textwidth}
    \centering
    \includegraphics[width=\textwidth]{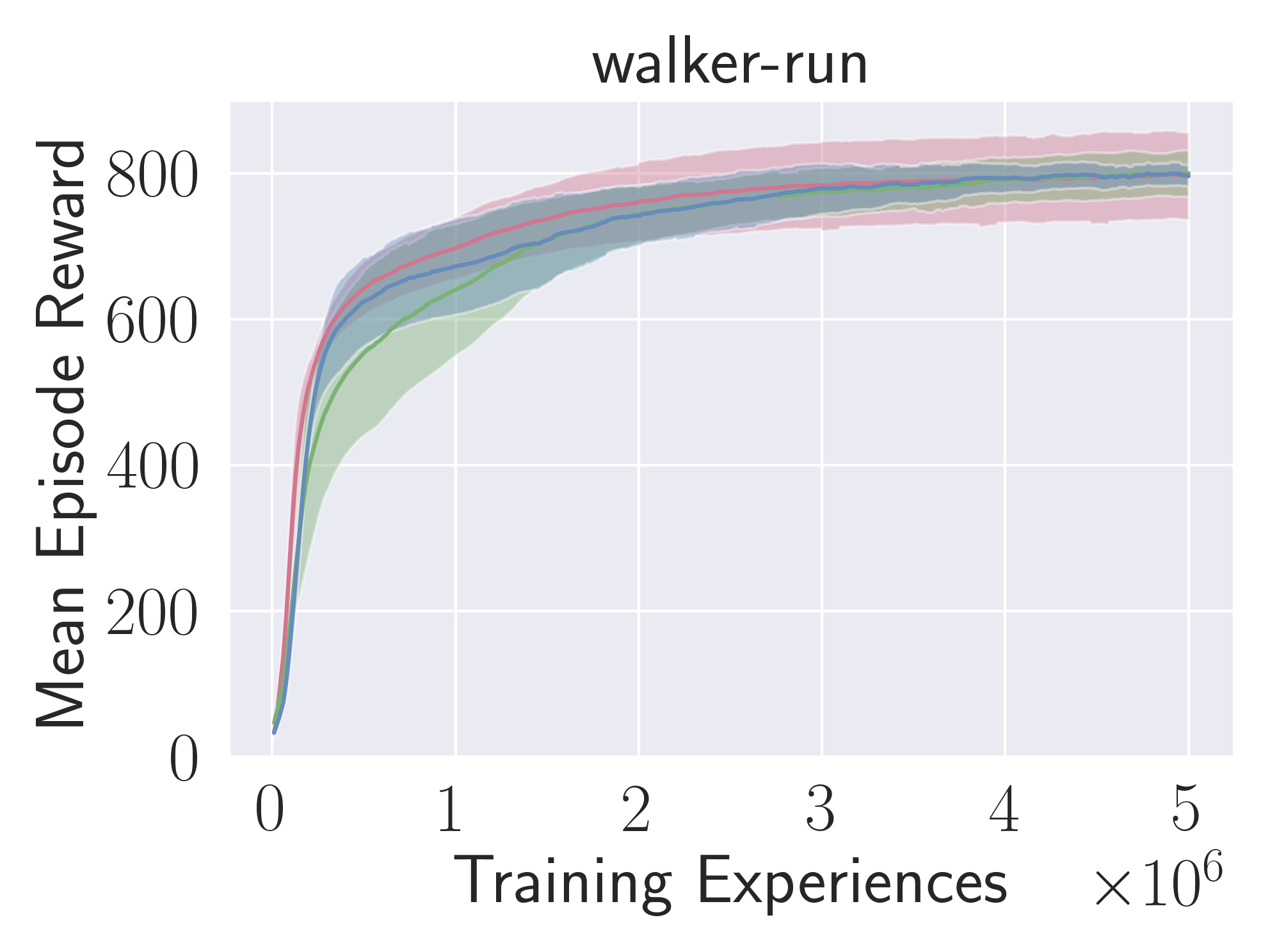}
    \end{subfigure}
    \begin{subfigure}{.24\textwidth}
    \centering
    \includegraphics[width=\textwidth]{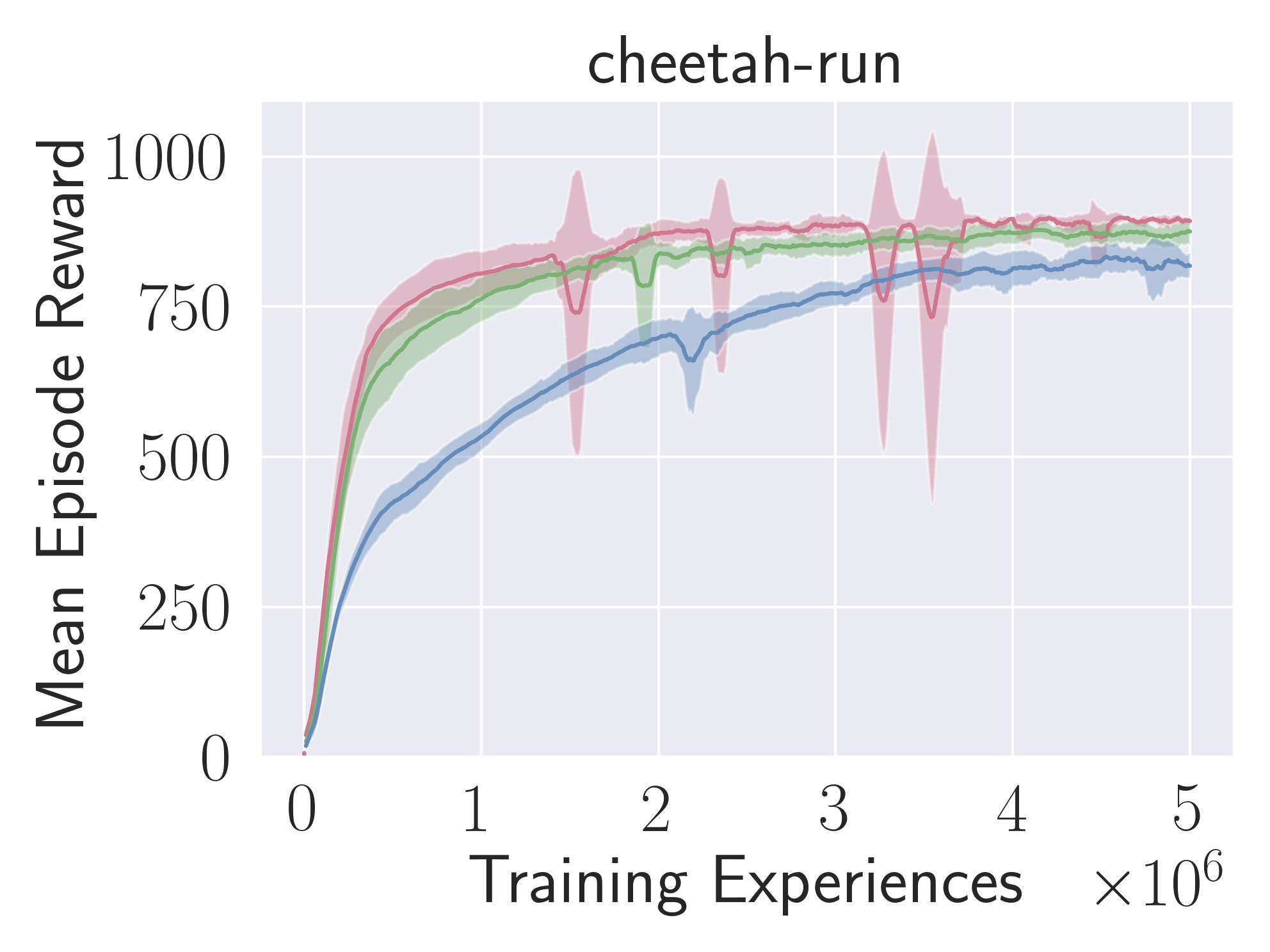}
    \end{subfigure}
    \begin{subfigure}{.24\textwidth}
    \centering
    \includegraphics[width=\textwidth]{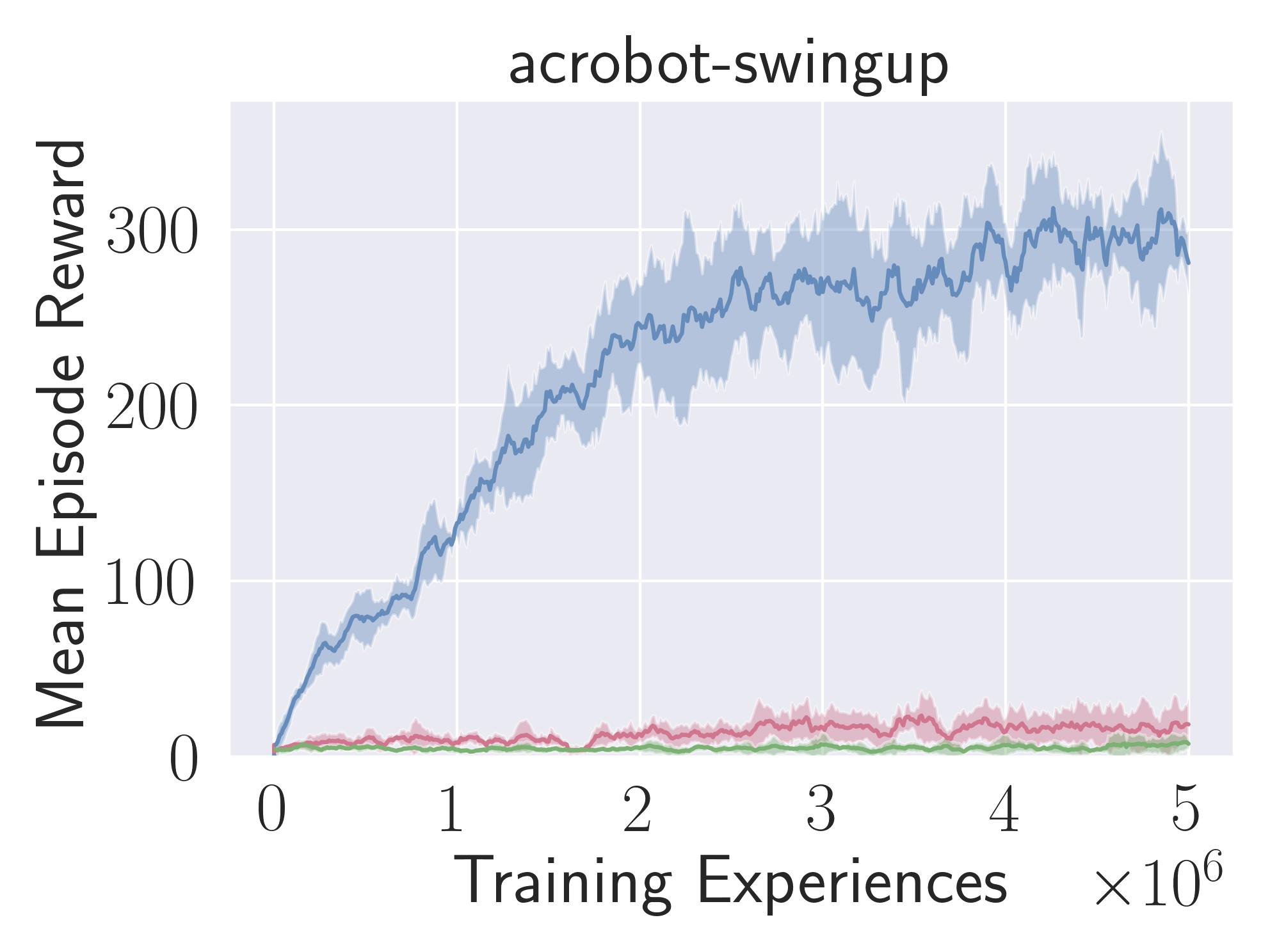}
    \end{subfigure}
    \begin{subfigure}{.24\textwidth}
    \centering
    \includegraphics[width=\textwidth]{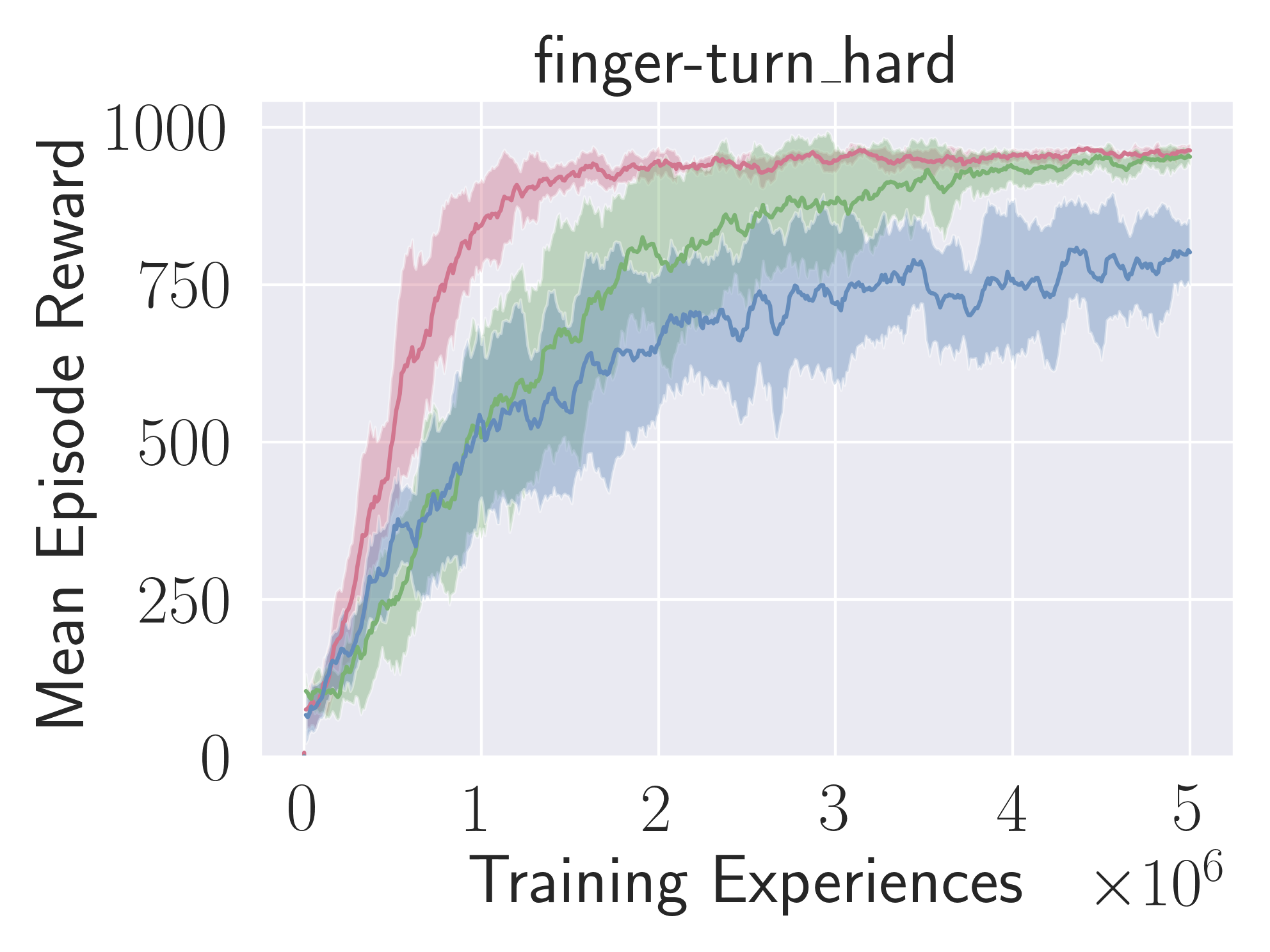}
    \end{subfigure}

    \begin{subfigure}{.24\textwidth}
    \centering
    \includegraphics[width=\textwidth]{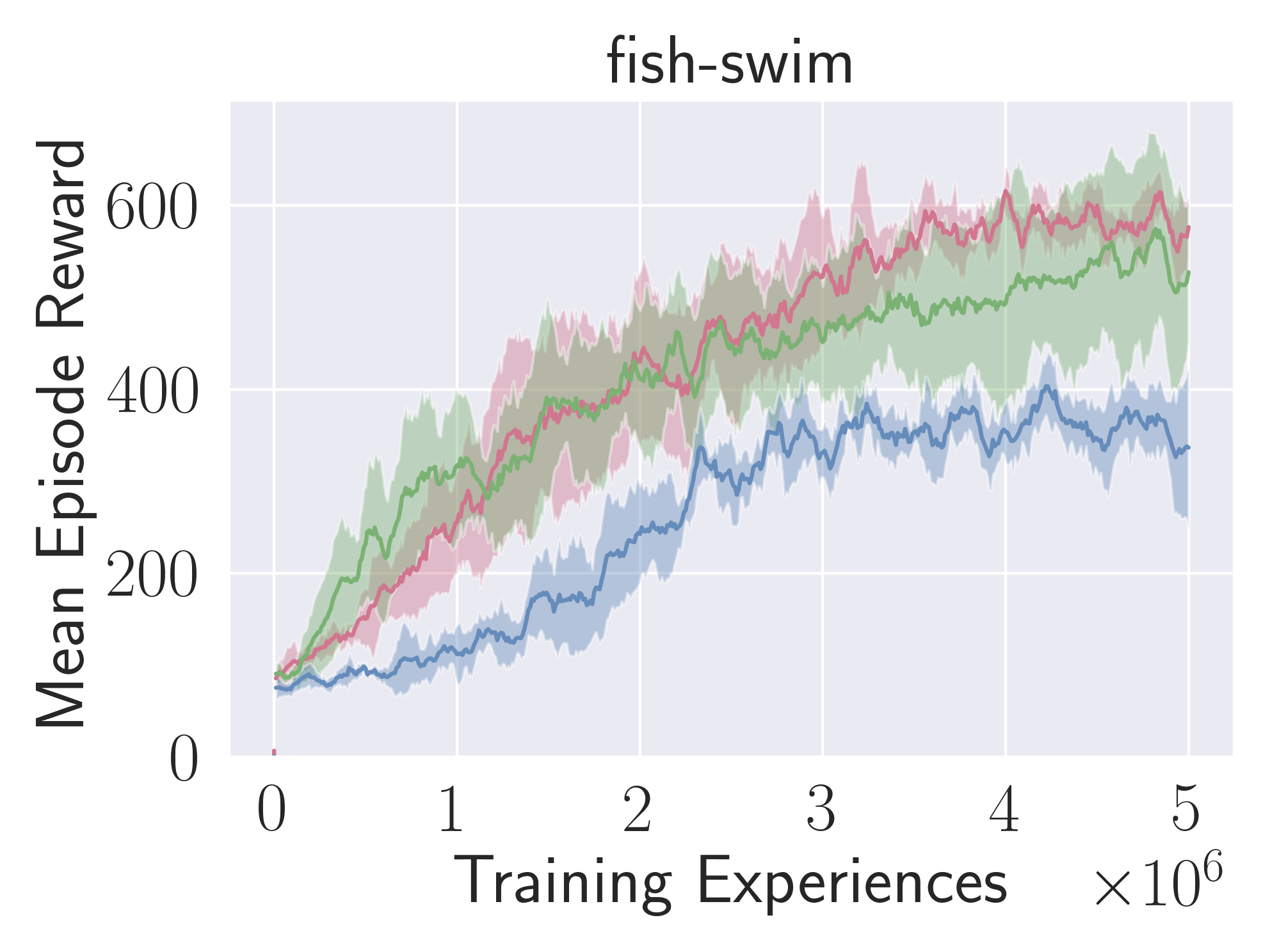}
    \end{subfigure}
    \begin{subfigure}{.24\textwidth}
    \centering
    \includegraphics[width=\textwidth]{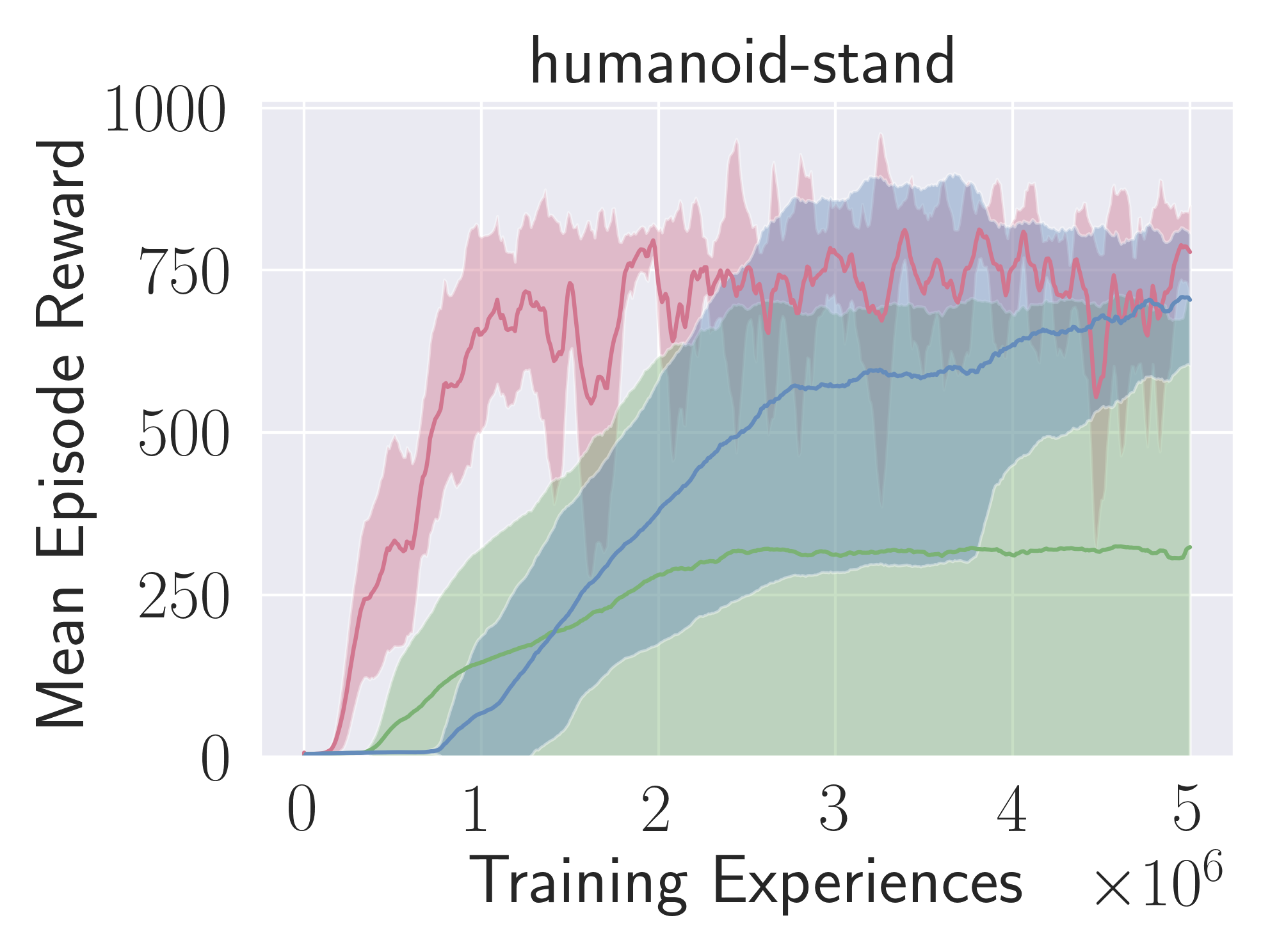}
    \end{subfigure}
    \begin{subfigure}{.24\textwidth}
    \centering
    \includegraphics[width=\textwidth]{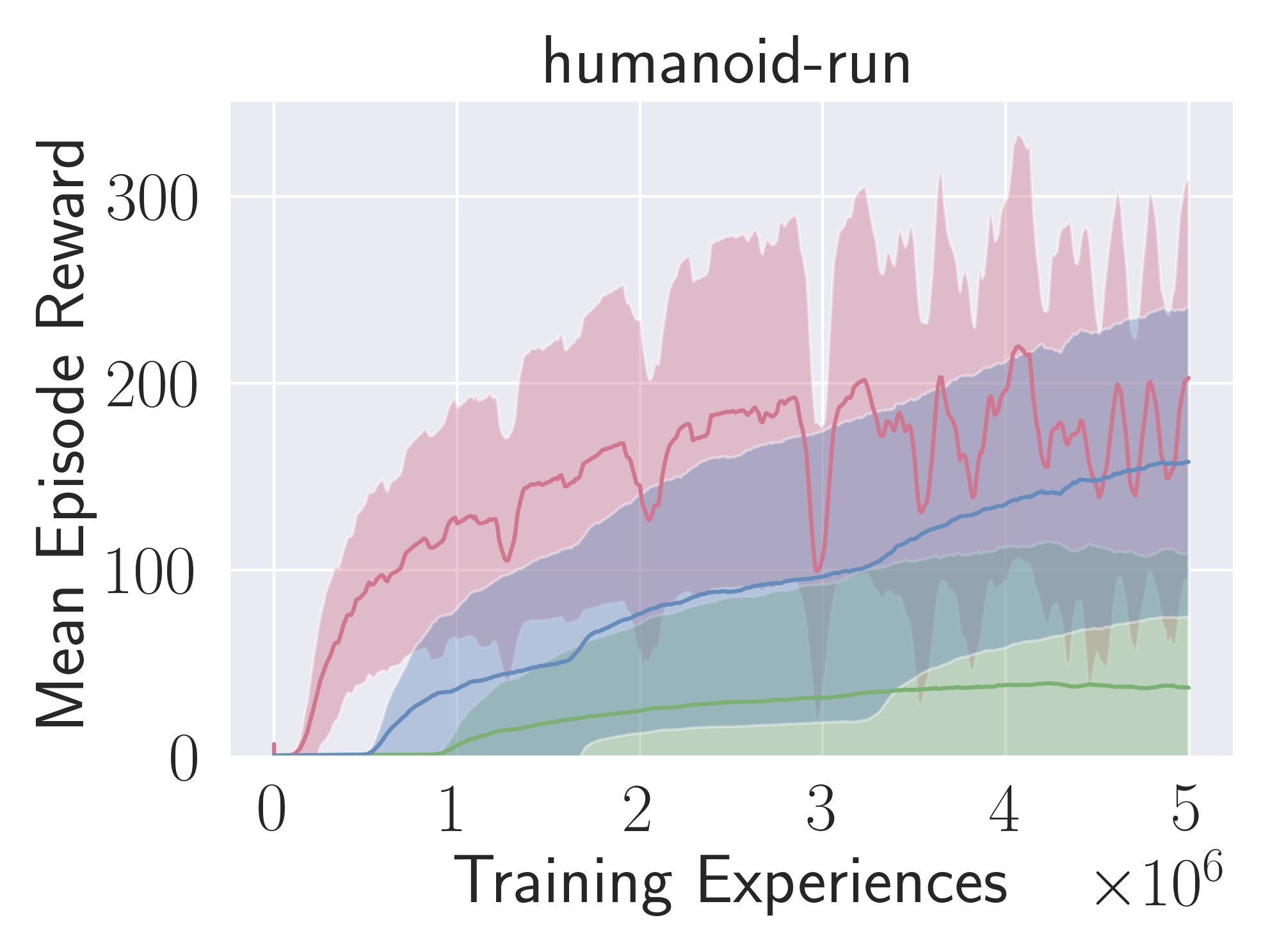}
    \end{subfigure}
    \begin{subfigure}{.24\textwidth}
    \centering
    \includegraphics[width=\textwidth]{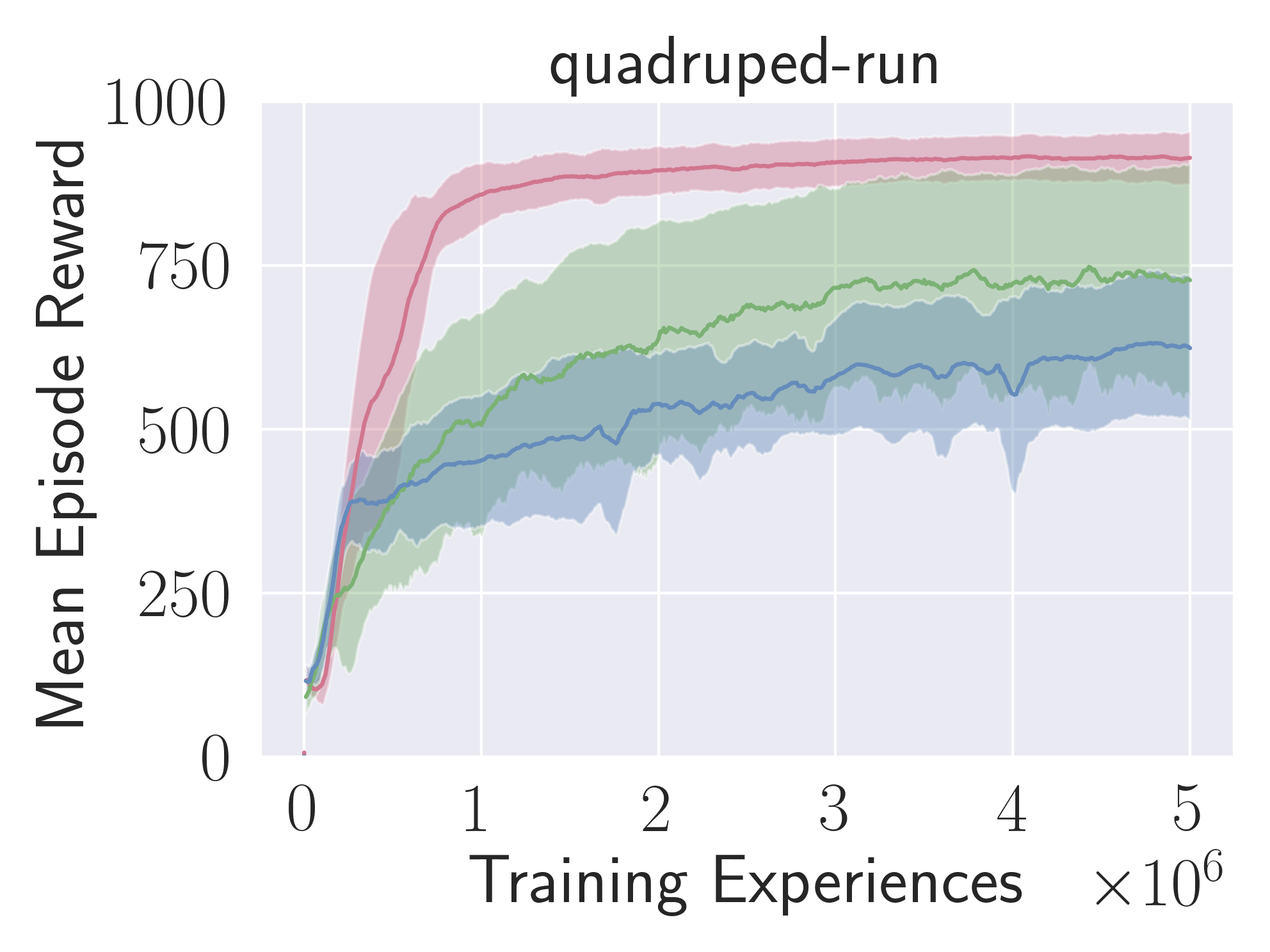}
    \end{subfigure}

    \begin{subfigure}{.24\textwidth}
    \centering
    \end{subfigure}
    \begin{subfigure}{.24\textwidth}
    \centering
    \includegraphics[width=\textwidth]{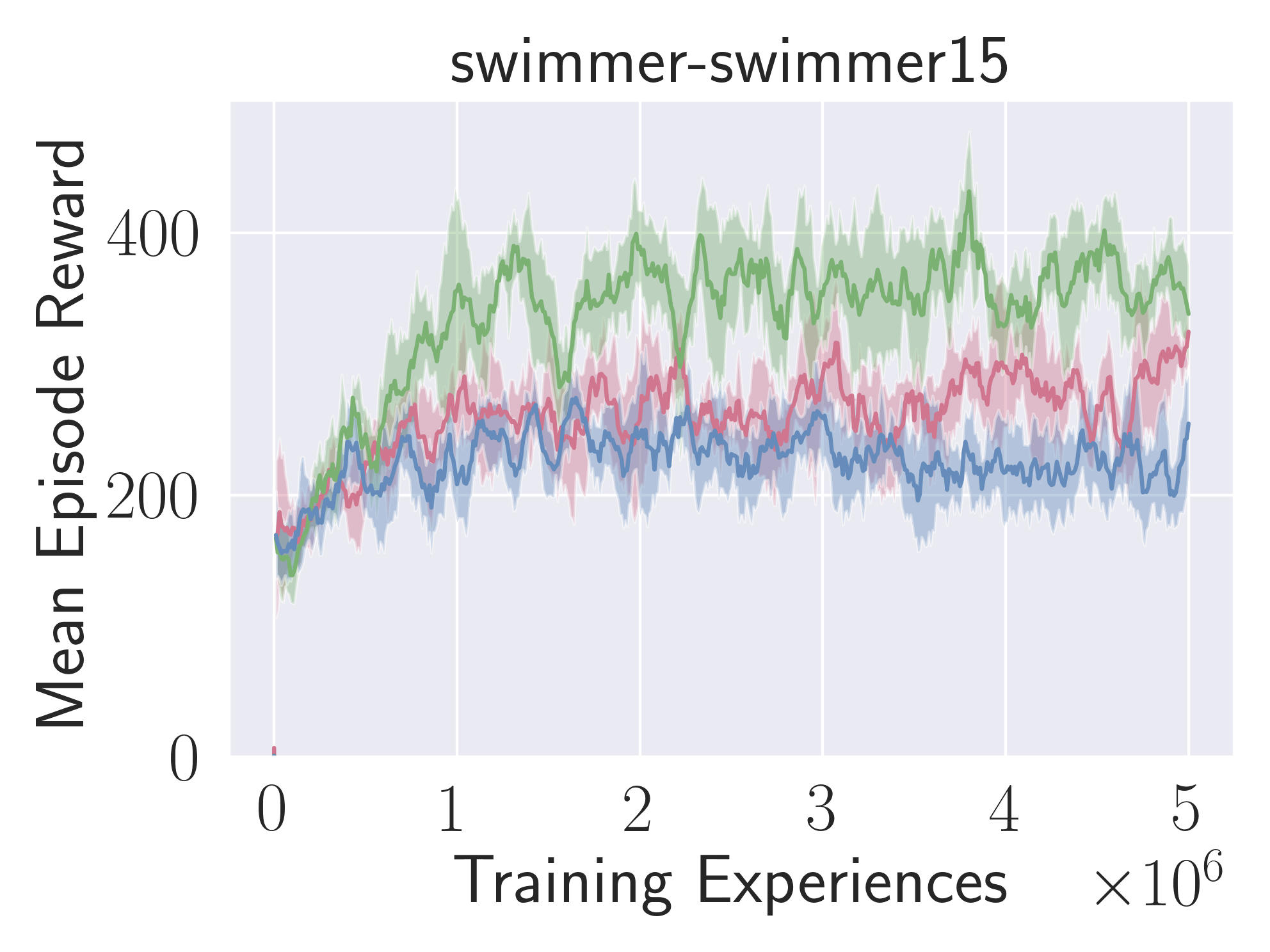}
    \end{subfigure}
    \begin{subfigure}{.24\textwidth}
    \centering
    \includegraphics[width=\textwidth]{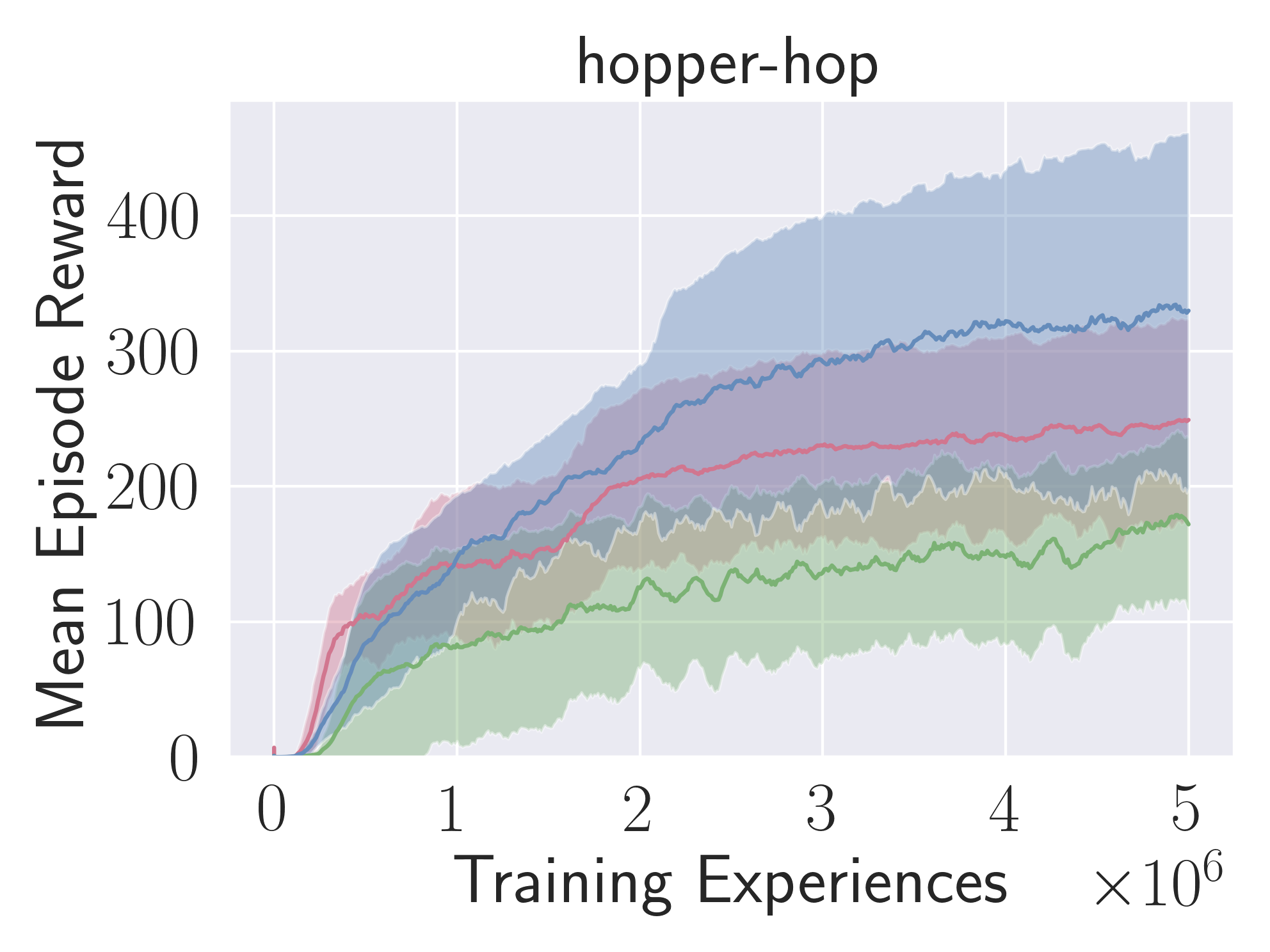}
    \end{subfigure}
    \begin{subfigure}{.24\textwidth}
    \centering
    \end{subfigure}
    \begin{subfigure}{\textwidth}
    \centering
    \includegraphics[width=0.5\textwidth]{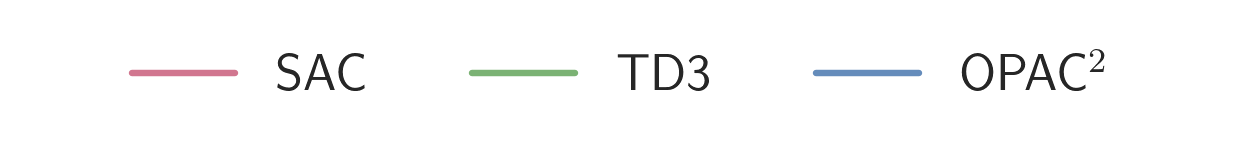}
    \end{subfigure}

    \caption{Per-environment results for 10 tasks from the DeepMind Control Suite.}
    \label{fig:full_dmc_rew}
\end{figure}

\begin{figure}[H]
\centering
\includegraphics[width=0.52\textwidth]{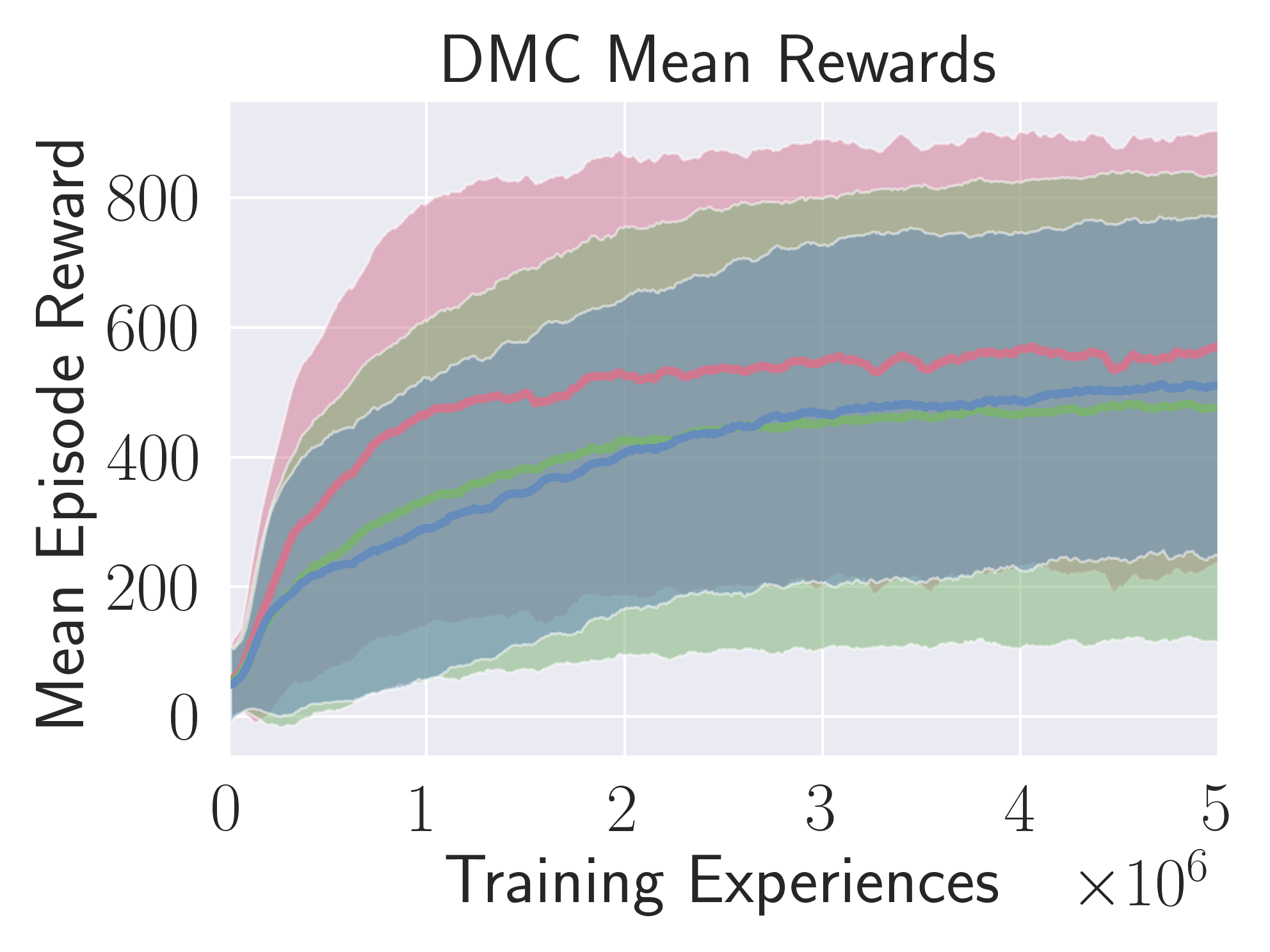}
\includegraphics[width=0.52\textwidth]{figures/dmc_new/legend.png}
\caption{Overall average performance of SAC, TD3, and OPAC$^2$ on 10 tasks from the DeepMind control suite.}
\end{figure}

\end{document}